%% file: main.tex
\documentclass[10pt,twocolumn,letterpaper]{article}

\usepackage{cvpr}              


\DeclareMathOperator*{\argmax}{argmax} 
\usepackage{multirow}
\usepackage{hhline}
\usepackage{tabularx}
\usepackage{subcaption}

\definecolor{cvprblue}{rgb}{0.21,0.49,0.74}
\usepackage[pagebackref,breaklinks,colorlinks,allcolors=cvprblue]{hyperref}

\def\paperID{X} 
\def\confName{CVPRW - OpenSUN3D}
\def\confYear{2025}

\title{Enforcing View-Consistency in Class-Agnostic 3D Segmentation Fields}

\author{
Corentin Dumery \quad
Aoxiang Fan \quad
Ren Li \quad
Nicolas Talabot \quad
Pascal Fua \\
EPFL CVLab \\
{\tt\small \{name.surname\}@epfl.ch}
}

\input{sec/defs}

\begin{document}
\maketitle

\input{sec/0_abstract}    
\input{sec/1_intro}
\input{sec/2_rw}
\input{sec/3_method}
\input{sec/4_experiments}
\input{sec/5_conclusion}
{
    \small
    \bibliographystyle{ieeenat_fullname}
    \bibliography{string,vision,graphics,biomed,optim,misc,main}
}

\input{sec/X_suppl}


\end{document}

%% file: sec/defs.tex
\newif\ifdraft
\draftfalse

\definecolor{orange}{rgb}{1,0.5,0}
\definecolor{green0}{rgb}{0.1,0.7,0.1}

\ifdraft
\newcommand{\PF}[1]{{\color{red}{\bf PF: #1}}}

\newcommand{\CD}[1]{{\color{blue}{\bf CD: #1}}}
\newcommand{\cd}[1]{{\color{blue} #1}}
\newcommand{\AF}[1]{{\color{green0}{\bf AF: #1}}}

\newcommand{\NT}[1]{{\color{violet}{\bf NT: #1}}}

\newcommand{\RL}[1]{{\color{orange}{\bf RL: #1}}}

\else
\newcommand{\PF}[1]{{\color{red}{}}}

\newcommand{\CD}[1]{{\color{blue}{}}}
\newcommand{\cd}[1]{ #1 }
\newcommand{\AF}[1]{{\color{green0}{}}}

\newcommand{\NT}[1]{{\color{violet}{}}}

\newcommand{\RL}[1]{{\color{orange}{}}}

\fi

\newcommand{\parag}[1]{\vspace{-2mm}\paragraph{#1}}

\newcommand{\bc}[0]{\mathbf{c}}
\newcommand{\bd}[0]{\mathbf{d}}

\newcommand{\bo}[0]{\mathbf{o}}

\newcommand{\br}[0]{\mathbf{r}}

\newcommand{\bx}[0]{\mathbf{x}}

%% file: sec/0_abstract.tex
\begin{abstract}
Radiance Fields have become a powerful tool for modeling 3D scenes from multiple images. However, they remain difficult to segment into semantically meaningful regions. 
Some methods work well using 2D semantic masks, but they generalize poorly to class-agnostic segmentations. More recent methods circumvent this issue by using contrastive learning to optimize a high-dimensional 3D feature field instead. However, recovering a segmentation then requires clustering and fine-tuning the associated hyperparameters.
In contrast, we aim to identify the necessary changes in segmentation field methods to directly learn a segmentation field while being robust to inconsistent class-agnostic masks, successfully decomposing the scene into a set of objects of any class.

By introducing an additional spatial regularization term and restricting the field to a limited number of competing object slots against which masks are matched, a meaningful object representation emerges that best explains the 2D supervision.
Our experiments demonstrate the ability of our method to generate 3D panoptic segmentations on complex scenes, and extract high-quality 3D assets from radiance fields that can then be used in virtual 3D environments.
\end{abstract}

%% file: sec/1_intro.tex

\input{fig/statement}

\section{Introduction}

Neural Radiance Fields (NeRFs)~\cite{Mildenhall20} and 3D Gaussian Splatting~\cite{Kerbl23} have been proposed for novel view synthesis from multiple images and have now become powerful 3D scene reconstruction tools. One limitation, however, is that they represent the scene geometry in terms of a single 3D density field in which all objects are blended. This makes downstream tasks such as editing or object identification particularly challenging. While there has been work on assigning semantic labels to parts of scenes described in this manner~\cite{Vora22,Zhi21,Liu23d}, the majority of early works are limited to classes seen during training. Such classes usually include easily identifiable household items in interior settings and  vehicles in street scenes for which labeled data is abundant. Unfortunately, this approach does not scale well to in-the-wild scenes with previously unseen objects. Panoptic Lifting~\cite{Siddiqui23} and Instance-NeRF~\cite{Liu23d} consistently miss objects that are not in a predefined set of classes. When given class-agnostic masks as input, we observed that these methods still fail, as they are unable to deal with the inconsistencies that arise from zero-shot segmentation, as illustrated by \cref{fig:statement}(a). More recently, some methods have attempted to overcome this limitation by directly learning 3D features instead of labels. For example,  DFF~\cite{Kobayashi22} and LERF~\cite{Kerr23} achieve this by distilling 2D features from Dino~\cite{Caron21} and CLIP~\cite{Radford21} respectively. However, the resulting segmentations tend to be fuzzy and are unable to separate different instances of the same object type. In another line of works, Garfield~\cite{Kim24} and Contrastive Lift~\cite{Bhalgat23} aim to learn a continuous 3D feature field from 2D masks, which must then be clustered as a post-processing step in order to produce a segmentation field. This additional step tends to produce imprecise segmentations unless additional hyperparameter are carefully set, such as the \textit{desired scale} of Garfield~\cite{Kim24} or the clustering distance threshold in Contrastive Lift~\cite{Bhalgat23}.

\cd{Finally, SA3D~\cite{Cen23a} and SANeRF-HQ~\cite{Liu23g} rely on user-interaction to segment a single object of interest in a scene. They are unable to produce panoptic segmentations over the complete 3D scene, as they simply learn a binary field around the user input. This proves to be a significantly simpler task, as real scenes can incorporate hundreds of objects and matching inconsistent masks across views is highly challenging.
}

In this work, we propose \textit{DiscoNeRF}, an approach to discover 3D objects in an open-world environment. DiscoNeRF segments 3D radiance fields without a preset number of classes or user-interaction while still producing sharp segmentations, as illustrated by \cref{fig:statement}(b). We observe that foundation models such as  SAM~\cite{Kirillov23} or Mask-DINO~\cite{Li23b} can assign labels to pixels in scenes containing an arbitrary number of objects at the cost of these labels being potentially inconsistent across views of the same scene. 
While operations such as mask dilation or disconnected component removal can sometimes help reduce the signal-to-noise ratio in these 2D masks, they are insufficient for significant artifacts and view-inconsistent representations.
Our key insight is that we can efficiently constrain the training of an object field to produce a view-consistent 3D segmentation despite the noisiness of the supervisory signal. This enables us to produce sharp segmentations that are consistent across views, unlike those of the original segmentation algorithm, and for objects belonging to arbitrary classes. These images can then be compared to the 2D segmentation masks returned by the foundation model~\cite{Kirillov23}.

\cref{fig:pipeline} depicts our architecture. In addition to the density and color that a standard radiance field predicts at each point in a volume, ours predicts a vector of probabilities of belonging to a particular object. These vectors are then used to render images whose pixels are the probabilities to belong to each class. Note that these object classes are not assigned a semantic meaning, they simply correspond to a channel in the probability vector. To enforce consistency across views between probability images and 2D segmentation masks, we introduce a robust loss function that generalizes the traditional Intersection over Union (IoU) measure to floating point values and relies on the Hungarian algorithm to establish correspondences between regions in the probability images and 2D masks in the segmentations. Minimizing this loss function using the appropriate regularization constraints yields the desired result.

Our contribution is therefore an approach to reconstructing 3D scenes in such a way that objects of {\it any} class can be properly segmented. We achieve this by matching mask with object channels and introducing an additional spatial-consistency regularizer. 
This allows a wide range of additional applications of reconstructed scenes, and the identified objects can be used for scene editing or provide spatial understanding to an autonomous system.

%% file: fig/statement.tex

\begin{figure*}[t]
	\centering
	\begin{subfigure}{0.49\linewidth}
		\includegraphics[width=\linewidth]{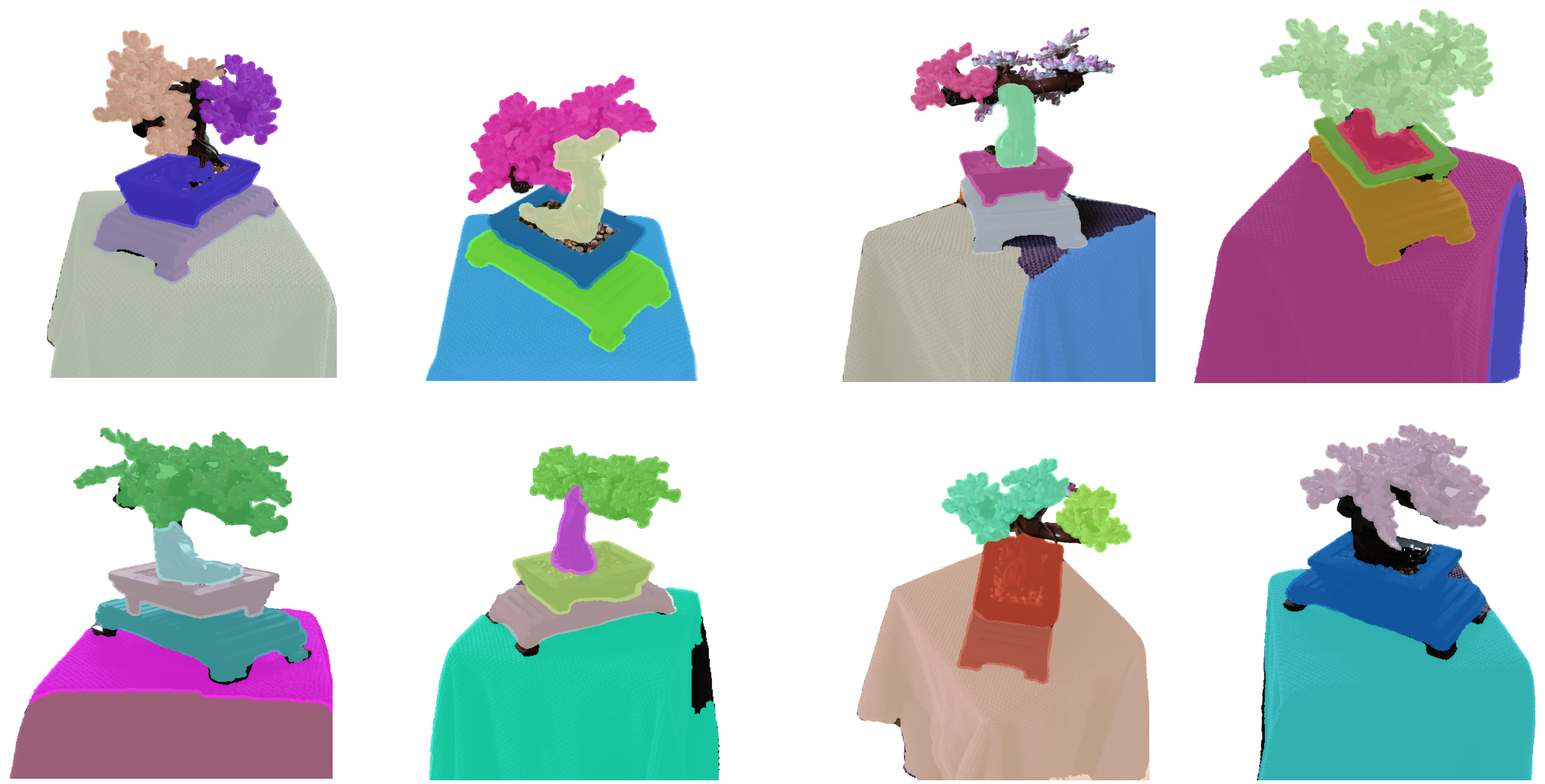}
		\caption{}
	\end{subfigure}
	\hfill
	\begin{subfigure}{0.49\linewidth}
		\includegraphics[width=\linewidth]{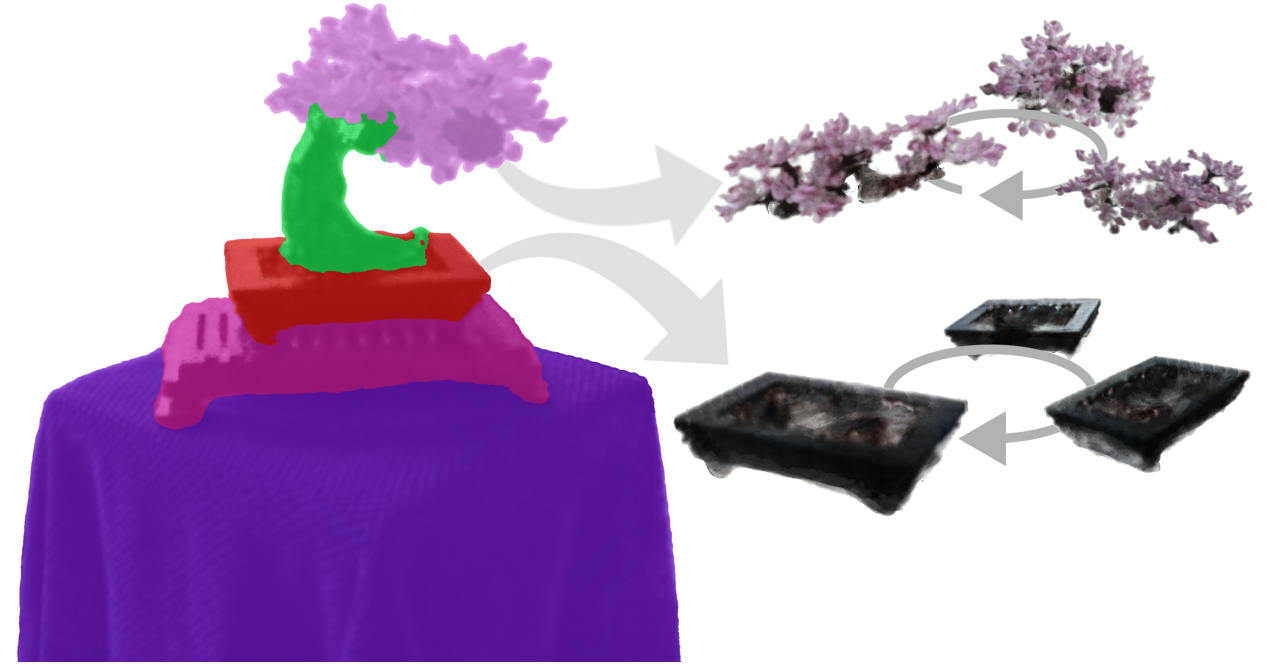}
		\caption{}
	\end{subfigure}
	\caption{\small {\bf Problem statement.} (a) Given as input a set of class-agnostic 2D masks \cite{Kirillov23} with little consistency across views, we aim to learn (b) a meaningful 3D object field that segments the different instances in the scene. The discovered objects can then be extracted and rendered independently.}
	\label{fig:statement}
\end{figure*}

%% file: sec/2_rw.tex

\input{fig/pipeline}

\section{Related Work}

In this section, we first briefly introduce the radiance field formulation~\cite{Mildenhall20} and then discuss the various approaches to semantic segmentation of such a representation.

\subsection{Radiance Fields}
\label{sec:nerf}

Given a set of \textit{posed} images of a scene, radiance fields~\cite{Mildenhall20} aim to generate new images of that scene from novel poses. This is done by learning a 3D scene representation which can be used to render the original images. More recently, 3D Gaussian Splatting~\cite{Kerbl23} has successfully proposed an explicit radiance field representation. In both cases, a radiance field model assigns a density $\sigma(\bx)$ and a view-dependent color $\bc(\bx,\bd)$ to each 3D point $\bx$, where $\bd$ is the direction from which $\bx$ is seen. The density can be interpreted as the differential probability of a ray terminating at an infinitesimal particle at location $\bx$.  Given a pixel in an image and the line of sight 
$\br(t)= \bo + t {\bf d}$  for $t$ between the near and far bounds  $t_n$ and $t_f$ going through it, the expected color of that pixel is  
\begin{equation}
	\begin{split}
	\hat{C}({\bf r}) =\int_{t_n}^{t_f} T(t)\, \sigma(\br(t))\, {\bf c}({\bf r}(t),\bd))\, dt \\
	\mbox{ where } T(t) = \exp \left( - \int_{t_n}^t \sigma (\br(s))  ds \right) \; . 
	\label{eq:nerf}
	\end{split}
\end{equation}
The function $T$ denotes the accumulated transmittance along the ray from $t_n$ to $t$, that is the probability that the ray travels from $t_n$  to $t$ without hitting any other particle.  In practice, one uses a multi-layer perceptron with weights $\Theta$ to define a function $f_{\Theta}$ that assigns to each $\bx$ and $\bd$ a pair $f_{\Theta}(\bx,\bd)=(\sigma(\bx),\bc(\bx,\bd)$). Note that $f$ is formulated such that the density $\sigma$ does not depend on the view direction $\bd$.

The weights $\Theta$ are adjusted so that the images generated by this model are as similar as possible to the input images of the scene. This is done by minimizing a Mean Squared Error loss of the form
\begin{equation}
	    \mathcal{L}_{RGB} = \sum \sum \| \hat{C} - C \|_2^2 \; ,
	    \label{eq:lossRGB}
\end{equation}
where $\hat{C}$ is the rendered color of \cref{eq:nerf}, $C$ is the actual color in the images, and the double summation is over all the pixels in all the input images. In practice, the loss is computed over batches of rays randomly sampled from all views.  

While the original NeRF implementation suffered from drawbacks in speed and accuracy, recent works have significantly mitigated these issues. 
A promising line of works improved the training speed of NeRFs by replacing the neural scene representation. 
TensoRF~\cite{Chen22f} proposes a reprensentation based on tensor-decomposition to compactly encode scene properties. More recently, InstantNGP~\cite{Mueller22} has introduced multi-resolution hash encoding to accomplish both efficient and high-fidelity scene modeling, providing a flexible and versatile tool for downstream applications. MipNeRF~\cite{Barron21} further improves the view synthesis quality of NeRF by representing the scene at a continuously-valued scale and efficiently rendering anti-aliased conical frustums instead of rays. MipNeRF360~\cite{Barron22} and NeRF++\cite{Zhang20f} propose to use different re-parametrization techniques to better model general and unbounded scenes. Finally, explicit methods such as SVRaster~\cite{sun2024sparse}, Radiant Foam~\cite{govindarajan2025radiant}, or 3D Gaussian Splatting~\cite{Kerbl23} have proposed explicit representations that greatly speed up rendering speed whilst preserving state-of-the-art visual quality. These recent developments have made it possible to reconstruct complex scenes with high visual accuracy. However, the reconstructions remain fully fused in a single reconstruction and thus unusable for many downstream applications, such as 3D asset generation or editing.

\subsection{Segmentation fields}

After training, the weights of the network that implements $f_{\Theta}$ jointly capture all the scene information, including lighting and color. However, they do not separate the different 3D objects in the scene and doing so is far from trivial.  Recently, it has been proposed to interactively extract objects given user-defined prompt points identifying a target in some of the training images~\cite{Cen23a} or directly in 3D~\cite{Chen23b}. The manually provided inputs eliminate much of the ambiguity arising from zero-shot segmentation, resulting in a significantly simpler task. In addition to requiring human interaction, this approach can only identify  one object at a time. In contrast, we propose an automated method that decomposes the whole scene into a set of objects without any additional human supervision.

In a separate line of work, some methods use contrastive learning to optimize a continuous field of 3D features that allows clustering of different objects in the scene~\cite{Kim24,Bhalgat23,Engelmann24,Liu23h,He24}. These representations learn a 3D \textit{feature field} and necessitate additional clustering to produce a \textit{segmentation field}, requiring carefully set hyperparameters to produce the desired results. Furthermore, they do not define a specific number of objects and is intrinsically continuous, making it a poor fit for segmenting NeRFs into discrete components. 
Similarly, \textit{feature distillation}~\cite{Kobayashi22}  can be used to learn a 3D field that matches the 2D features from various foundation models. This has been used by recent work to perform  text-queries~\cite{Kerr23} or interactive segmentation~\cite{Chen23b}.

Some previous work demonstrated the potential of radiance fields for 3D semantic modeling~\cite{Fu22,Yu22c,Siddiqui23,Liu23d,Vora22,Cheng23c}. Some of these algorithms~\cite{Vora22,Liu23d} query the NeRF model on a sparse grid, and then apply a point-cloud segmentation method based on a 3D U-Net~\cite{Cicek16}.
The generality of these approaches is inherently limited by the pre-defined classes they are trained to identify and they generalize poorly to zero-shot classes. 
In a different but related line of work, some methods~\cite{Qin24,Dou24,Lyu24,Ye23,Nguyen24} have been proposed to segment 3D gaussians~\cite{Kerbl23} by assigning labels to individual splats. These approaches do not define a 3D field, unlike NeRF segmentation methods, and due to their vastly different scene representation it is not straightforward to adapt them to segment scenes reconstructed with NeRFs. 

%% file: fig/pipeline.tex

\begin{figure*}[th]
\centering
\includegraphics[width=0.99\textwidth]{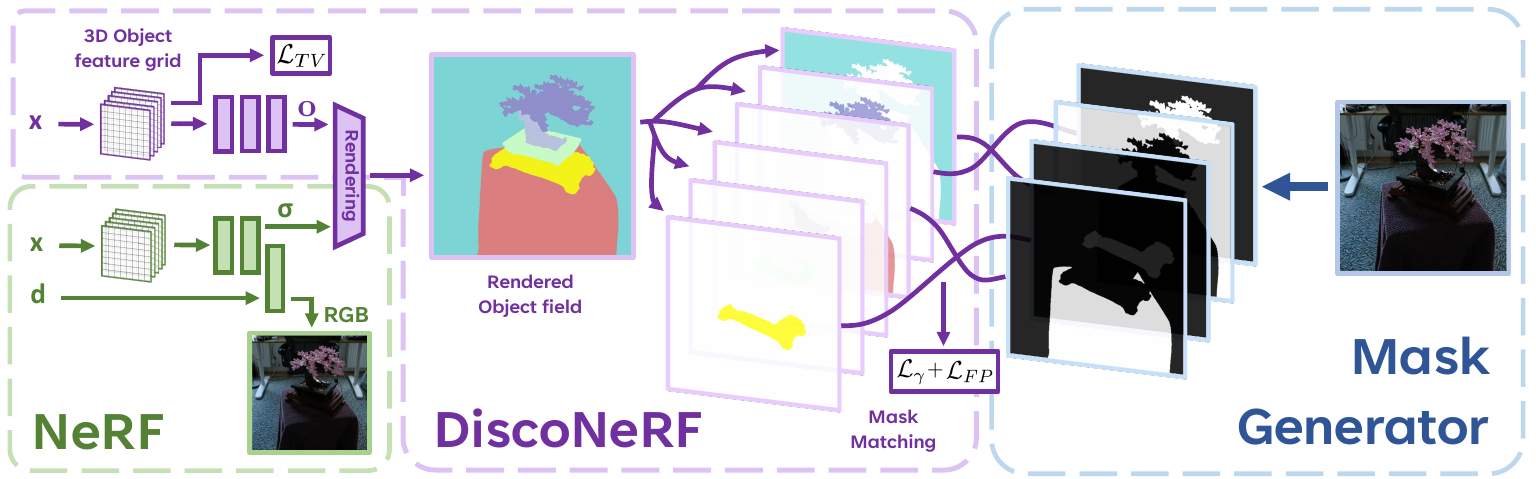}
\caption{\small {\bf DiscoNeRF pipeline.} We add to a radiance field (green) an object network (violet) that predicts probabilities of belonging to each object. These are used to render 2D probability images, which are compared to the segmentation masks (blue) produced by a foundation model~\cite{Kirillov23}. We introduce three losses $\mathcal{L}_{TV}, \mathcal{L}_{\gamma}$ and $\mathcal{L}_{FP}$, designed to be robust to the inconsistency in the supervision signal.
}
\label{fig:pipeline}
\end{figure*}

%% file: sec/3_method.tex

\input{fig/assign}

\section{Approach}
\label{sec:method}

In radiance fields, each point in space is assigned a color and a density, as discussed in \cref{sec:nerf}. Our goal is to also assign a label, which should be the same for all points belonging to the same object. In earlier work~\cite{Siddiqui23,Liu23d}, this has been tried by making the function $f_{\Theta}$ introduced in \cref{sec:nerf} to output not only a density and a color but also an object label, thus implicitly defining a \textit{segmentation field}. This additional output can then be supervised using 2D semantic masks extracted from the ground truth images. However, masks of in-the-wild objects from previously unseen classes are often unavailable or inconsistent across views, making the training of the extended $f_{\Theta}$ difficult. 

In this work we expand on the idea of~\cite{Siddiqui23,Liu23d} to make the training more robust to these inconsistencies. We write the output of our function $f_{\Theta}$ as
\begin{equation}
    f(\bx, \bd) = (\sigma(\bx), \mathbf{c}(\bx,\bd), \bo(\bx)) \; , \label{eq:network}
\end{equation}
where $\bo$ is an $N$-dimensional vector, which we refer to as the {\it object probability vector}, and $N$ is the maximum number of objects that can be detected. 
The $i$-th coordinate of $\bo$ is interpreted as the probability of point $\mathbf{x}$ being part of the $i$-th object. 
Given $\bx$ and $\bd$, it computes $\sigma$ and $\bc$ in the standard NeRF manner. $\bo$ is predicted by a separate {\it object network}, consisting of a small MLP and a \textit{softmax} layer. Our approach is depicted by \cref{fig:pipeline} and we describe its individual components below.

%

First, the input $\bx$ is encoded by interpolating features from a learned 3D hash grid~\cite{Mueller22}. Then the encoded $\bx$ is fed to the object decoder. This produces a 3D object field $\bo(\bx)$, which can then be rendered to 2D using $\sigma(\bx)$ and \cref{eq:nerf}. Note that, by design, $\bo$ takes $\bx$ as its sole input because a semantic label should be independent of the viewpoint.

\subsection{Loss Function}

To train our object network, we need to design a loss function that is minimized when the binary masks generated from probabilities in the vectors $\bo$ are consistent with the 2D segmentation masks generated by the foundation model. The main difficulty being the noisiness in the supervision signal, and the fact that there is {\it a priori} no correspondence between object labels and segmentation masks in individual images. 

To this end, let us consider one specific image in which the foundation model generates $K$ binary masks $(M_m)_{1 \leq m \leq K}$. In our experiments, these masks are generated by the \textit{Segment Anything Model}~\cite{Kirillov23}. Recall that each one of the $N$ channels of $\bo$ represents the probability of belonging to a specific object. We will refer to them as {\it object slots}.  By using the volume rendering scheme of \cref{eq:nerf}, we can generate a probability image $O$, in which $O_{n, i, j}$ denotes the probability of pixel $(i, j)$ belonging to object $n$. Since no correspondence  between the $m$ and $n$ indices is given, we need to establish them by comparing the $M_m$ and $O_n$ masks. To this end, we first compute the affinity $\alpha$ between $M_m$ and $O_n$  as

\begin{equation}
\alpha(O_n, M_m) = \frac{\sum\limits_{i,j} \mathbf{min}(O_{n,i,j}, M_{m,i,j})}{\sum\limits_{i,j} \mathbf{max}(O_{n,i,j}, M_{m,i,j})}   \; . \label{eq:affinity}
\end{equation}

Given that $M_{m,i,j}$ is either 0 or 1 and $0 \leq O_{n,i,j} \leq 1$, this definition can be understood as generalizing the Intersection over Union (IoU) score to floating point values, as illustrated by \cref{fig:assignment}. Finally, we compute a matching between masks and object slots that maximizes affinity using the Hungarian algorithm~\cite{Kuhn55}. We set the maximum number of object slots $N$ to be at least as large as the number of masks $K$ from any given view, such that all masks get assigned an object slot while some slots can be left unassigned. Note that the computation of the affinity $\alpha$ and the matching of masks with object slots are performed with a stop-gradient to prevent this association from interfering with the learning of the object field. 

Let us denote the output of the Hungarian matching as $\gamma$. We define the two losses
\begin{align}
    \mathcal{L}_{\gamma} & = \frac{1}{K} \sum\limits_{m = 1}^{K} \|M_m - O_{\gamma(m)}\|_2^2    \; , \label{eq:objLoss} \\
    \mathcal{L}_{FP} & =  \frac{1}{K} \sum\limits_{m=1}^{K}\sum\limits_{n \neq \gamma(m)} \|M_m * O_{n}\|_2^2    \; , 
\end{align}
where $*$ denotes the pixel-wise product.  Minimizing the {\it Matching} loss $\mathcal{L}_{\gamma}$ encourages the assigned channel to match its corresponding mask, while minimizing the {\it False Positive} loss $\mathcal{L}_{FP}$ penalizes all other object slots that are not zero in the masked area. We compute a matching $\gamma$ for each image, and average $\mathcal{L}_{\gamma}$ and $\mathcal{L}_{FP}$ over all images in a given training batch.

Intuitively, the matching $\gamma$ identifies the most-likely object slot for each input mask. These masks sometimes suffer from \textit{over-segmentation} arising from zero-shot segmentation. This effect is mitigated by our object field $\bo$, which computes probabilities with a \textit{softmax} layer such that different slots are competing for each pixel and two overlapping representations cannot coexist. Consequently, over-segmented masks will tend to be ignored as masks that are present more often in the supervision signal will spontaneously be preferred during optimization. 

While a comparable mask matching is present in~\cite{Siddiqui23}, it is used on masks with limited classes and fails on class-agnostic segmentations. As demonstrated in our experiments, our proposed $\alpha$, $\mathcal{L}_{\gamma}$ and $\mathcal{L}_{FP}$ are necessary to robustly deal with this inconsistent 2D signal.
To further improve the consistency of our segmentation, we introduce an additional regularization term in the following section.

\subsection{Field Regularization}


As demonstrated in our experiments, minimizing $\mathcal{L}_{\gamma}$ and $\mathcal{L}_{FP}$ alone can still produce noisy segmentation fields. To regularize learning, we implement the object field $\bo$ as a 3D hash grid of learnable features~\cite{Mueller22}. To query $\bo(\bx)$, the 3D point $\bx$ is first encoded by interpolating the hash grid features, then these features are decoded by a fully-connected network. Unlike in earlier approaches~\cite{Liu23d,Siddiqui23}, this modeling allows us to regularize the object field by smoothing the learned features. Specifically, we regularize $\bo(\bx)$ by introducing an additional total variation loss $\mathcal{L}_{TV}$ taken to be

\begin{equation}
\mathcal{L}_{TV} = \sum\limits_{i,j \in \mathcal{N}} \|h_i - h_j\|_2^2
\end{equation}
where $h_i$ is the learned features for the $i$-th voxel in the grid, and $\mathcal{N}$ defines the neighborhood of adjacent voxels in the grid. We found this additional regularization to be essential to handle the ambiguities of zero-shot 2D segmentations, as we show below in an ablation study. In the presence of conflicting 2D masks such as the ones in \cref{fig:statement}, a naive implementation will attempt to fit all masks in the training set, resulting in noisy 3D segmentations. By introducing an additional TV regularizer, our method encourages the emergence of a consistent 3D segmentation. Even if the introduction of this additional loss term can in rare cases lead to overly smooth segmentations, we observed that it overall improved results both quantitatively and qualitatively.

\subsection{Training}

Given a set of posed images as input, we first train a neural radiance field to reconstruct the scene by sampling random rays across different views. We use recent techniques that help make radiance field reconstruction magnitudes faster and more accurate. We train a proposal network~\cite{Barron22} to improve the ray sampling strategy, and employ hash features~\cite{Mueller22} to speed up training. We use a coarse-to-fine activation of hash grids~\cite{Li23c}, improving the robustness of reconstruction at a minimal cost in training speed. 

Then, we freeze the weights of the model and train the object network. Unlike in the previous step, this optimization is performed by sampling random views and considering all the masks associated to these views jointly. We then minimize	$\mathcal{L}_{\gamma} + \lambda_{FP} \mathcal{L}_{FP} + \lambda_{TV} \mathcal{L}_{TV}$, where $ \lambda_{FP}$ and $ \lambda_{TV}$ are taken to be 0.01 unless explicitly stated otherwise.

%% file: fig/assign.tex

\begin{figure*}[t]
    \centering
    \begin{subfigure}{0.31\textwidth}
      \centering
      \includegraphics[width=\linewidth]{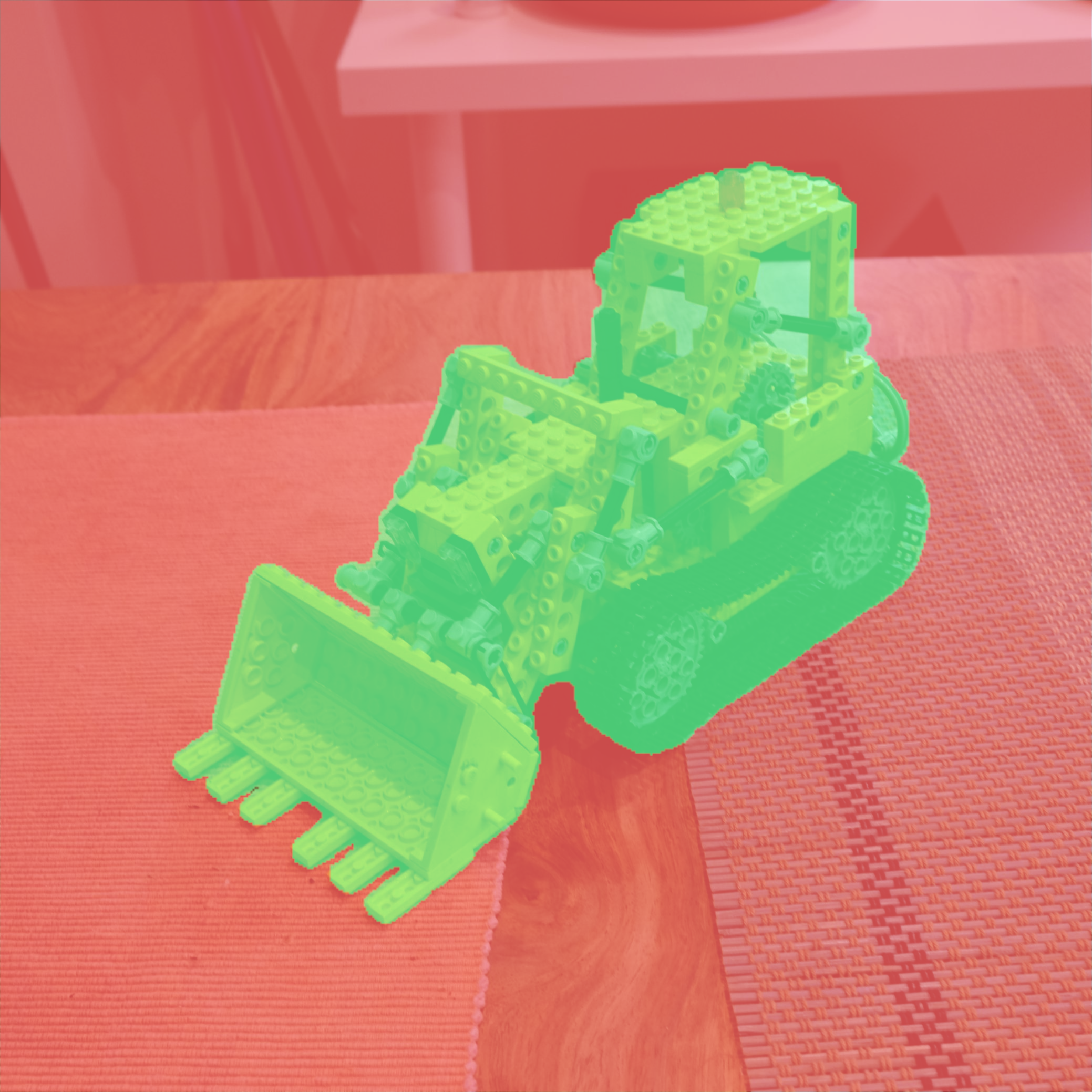}
      \caption{}
      \label{subfig:a}
    \end{subfigure}%
    \begin{subfigure}{0.6\textwidth}
      \centering
      \begin{tabular}{ccccc}
        \begin{subfigure}{0.19\linewidth}
          \centering
          \caption*{$O_1$}
          \includegraphics[width=\linewidth]{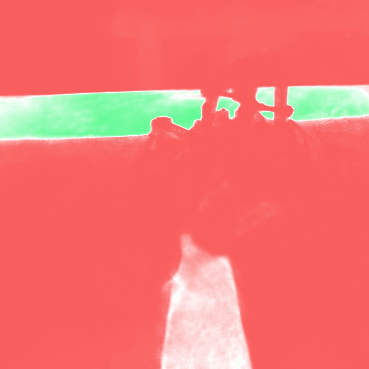}
        \end{subfigure} \hspace{-1em} &
        \begin{subfigure}{0.19\linewidth}
          \centering
          \caption*{$O_2$}
          \includegraphics[width=\linewidth]{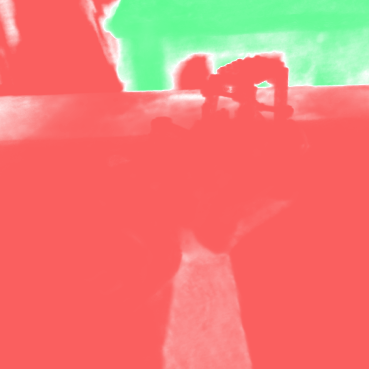}
        \end{subfigure} \hspace{-1em} &
        \begin{subfigure}{0.19\linewidth}
          \centering
          \caption*{$O_3$}
          \includegraphics[width=\linewidth]{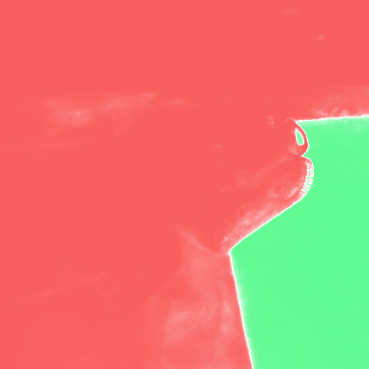}
        \end{subfigure} \hspace{-1em} &
        \begin{subfigure}{0.19\linewidth}
          \centering
          \caption*{$O_4$}
          \includegraphics[width=\linewidth]{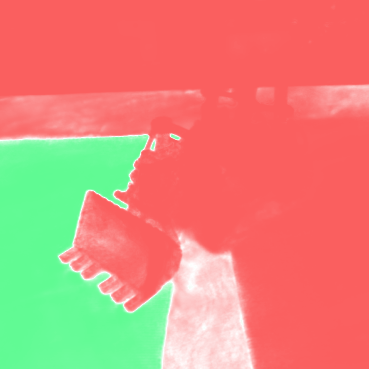}
        \end{subfigure} \hspace{-1em} &
        \begin{subfigure}{0.19\linewidth}
          \centering
          \caption*{$O_5$}
          \includegraphics[width=\linewidth]{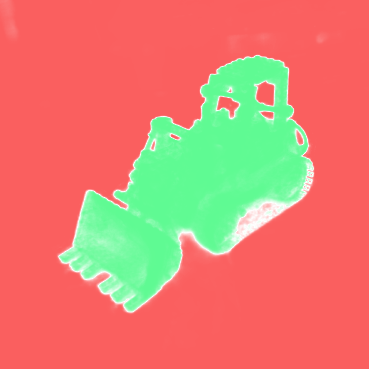}
        \end{subfigure} \vspace{0.5em} \\
        \begin{subfigure}{0.19\linewidth}
          \centering
          \includegraphics[width=\linewidth]{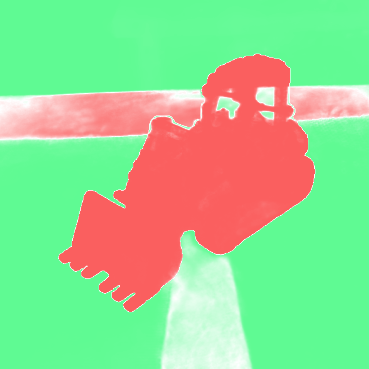}
          \caption*{$\alpha=1.8\%$}
        \end{subfigure} \hspace{-1em} &
        \begin{subfigure}{0.19\linewidth}
          \centering
          \includegraphics[width=\linewidth]{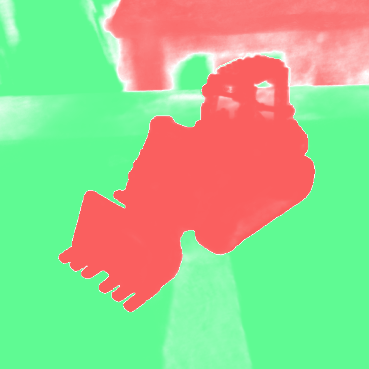}
          \caption*{$\alpha=0.8\%$}
        \end{subfigure} \hspace{-1em} &
        \begin{subfigure}{0.19\linewidth}
          \centering
          \includegraphics[width=\linewidth]{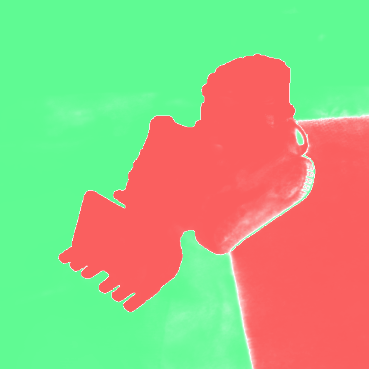}
          \caption*{$\alpha=1.3\%$}
        \end{subfigure} \hspace{-1em} &
        \begin{subfigure}{0.19\linewidth}
          \centering
          \includegraphics[width=\linewidth]{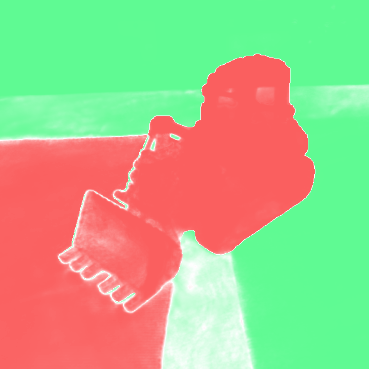}
          \caption*{$\alpha=1.8\%$}
        \end{subfigure} \hspace{-1em} &
        \begin{subfigure}{0.19\linewidth}
          \centering
          \includegraphics[width=\linewidth]{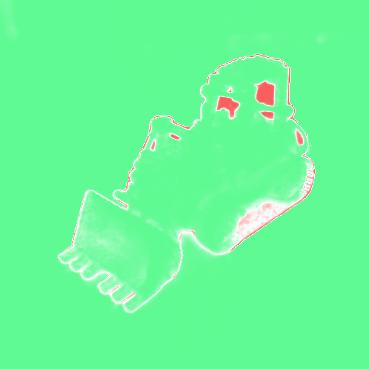}
          \caption*{$\alpha=87.1\%$}
        \end{subfigure} \\
      \end{tabular}
      \caption{}
      \label{subfig:b}
    \end{subfigure}%
    \hfill
    \begin{subfigure}{0.055\textwidth}
      \centering
      \includegraphics[width=\linewidth]{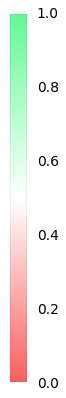}
      \label{subfig:c}
    \end{subfigure}
    \hfill
    \caption{Our affinity function $\alpha$ considers (a) a generated mask $M_m$ and (b) all object field channels $(O_n)_{1\leq n \leq N}$ independently. Bottom row: visualization of $1 - |M_m - O_n|$.}
    \label{fig:assignment}
  \end{figure*}

%% file: sec/4_experiments.tex

\section{Experiments}

Our implementation is based on \textit{torch-ngp}~\cite{torch-ngp}, which incorporates the speedup of Instant-NGP~\cite{Mueller22}. To generate the segmentation masks of $M_m$ in \cref{sec:method}, we use SAM~\cite{Kirillov23}. For object-centric scenes, we generate a background mask from the pixels that lie outside of the 3D unit sphere.

We use the same loss weights of $\lambda_{FP}=0.01$ and $ \lambda_{TV}=0.01$ for all scenes and datasets.
Finally, in all experiments, we automatically set $N = 2 \times K_{max}$, where $K_{max}$ is the number of masks in the view with the most masks produced. In practice, we found this to be high enough to capture each object in all tested scenes.

\input{fig/segmentations}

\subsection{Comparative Evaluation}

We evaluate our method quantitatively on the Mip-NeRF 360 dataset~\cite{Barron22}, which features both indoor and outdoor scenes containing a wide variety of common and less common objects. To this end, we render our object field to an image and select the $\argmax$ per pixel to produce discrete segmentation masks. We then compare them against hand-generated ground-truth masks for each scene except the \textit{stump} scene, which has no object and was therefore excluded from our experiments.

\input{fig/ablation}

\input{fig/comparison}

\parag{Metrics.}

We report the 2D IoU between ground-truth masks and masks produced by each method. 
We match each ground truth mask with one of the produced masks using the IoU, and compute a weighted average based on the area in order to prevent small masks from skewing the metric.
We additionally report the \textit{Best Dice} (BD) score between the ground truth and the masks produced by our method. For two sets of masks $\mathcal{P}$ and $\mathcal{G}$, it is defined as
\begin{equation}
BD(\mathcal{P},\mathcal{G}) = \frac{1}{N} \sum\limits_{i=1}^N \max\limits_{j=1:M} \frac{2 |p_i \cap g_j|}{|p_i| + |g_j|} \; ,    
\end{equation}
along with its symmetric counterpart $SBD(\mathcal{P},\mathcal{G}) = \min\{BD(\mathcal{P},\mathcal{G}), BD(\mathcal{G},\mathcal{P})\}$, as detailed in~\cite{Chen23c}.

\parag{Baselines.}

For comparison purposes, we implement a modernized version of Distilled Feature Fields (DFF)~\cite{Kobayashi22}, which jointly learns a NeRF model and a 3D field replicating the features produced by a foundation model in a process called \textit{feature distillation}. Specifically, we implement DFF with distillation of the DINOv2 features~\cite{Oquab23}, as opposed to DINO features in the original method~\cite{Caron21}. We will refer to this method as {\it DFFv2}. To make the distillation tractable, we first reduce the dimensionality of DINOv2 features to 64 with PCA, then produce 2D segmentations by rendering the distilled features and clustering them. 

We also compare our method against Panoptic Lifting~\cite{Siddiqui23} and Instance-NeRF~\cite{Liu23d}, two methods for semantic segmentation of NeRF models that rely on 2D and 3D segmentation models pretrained on specific object classes. 
Panoptic Lifting~\cite{Siddiqui23} jointly trains a NeRF and a multi-layer perceptron to match 2D semantic and instance maps. \cd{Additionally, we modified Panoptic Lifting to use masks produced by SAM. We refer to this method as PL+SAM. Note that the 2D prediction confidence and the semantic losses $\mathcal{L}_{sem}$ and $\mathcal{L}_{con}$ of the original paper cannot be used in this context due to the absence of ground-truth semantic classes.}
Instance-NeRF~\cite{Liu23d} fuses inconsistent 2D segmentations by querying a NeRF on a 3D grid and applying a point-cloud segmentation model. 
Combining Instance-NeRF with SAM does not improve its performance, as it does not only rely on semantic 2D masks but also on a semantic 3D point-cloud segmentation network.
Note that we do not compare with contrastive methods, as they only learn a feature field and require fine-tuning additional clustering hyperparameters to generate a usable segmentation field.

\begin{figure*}[t]
    \centering
    \begin{subfigure}[t]{0.325\textwidth}
        \centering
        \includegraphics[width=0.49\textwidth]{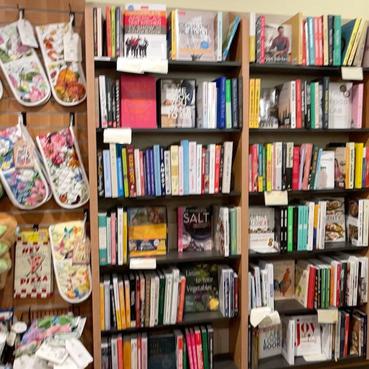}
        \includegraphics[width=0.49\textwidth]{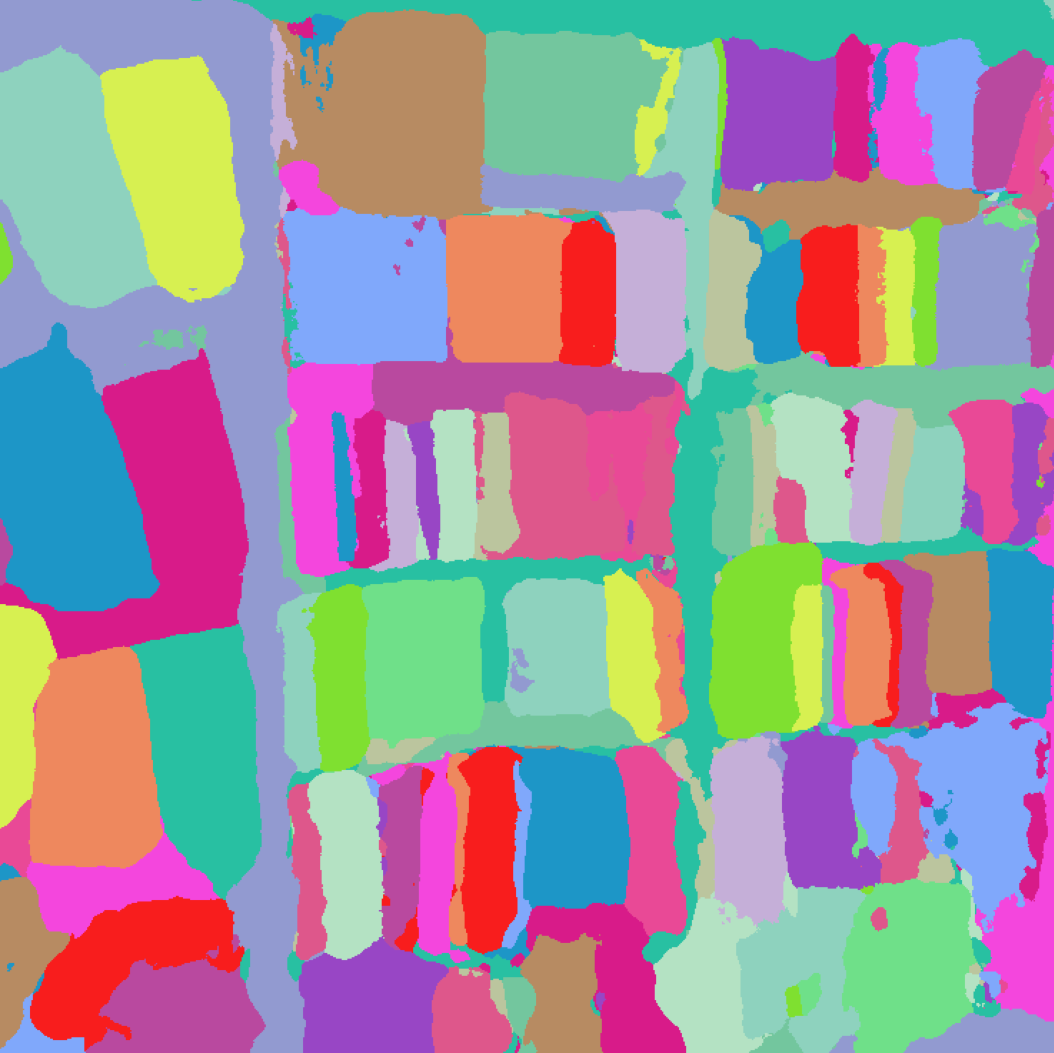}
        \caption{\textit{Book store}}
    \end{subfigure}\hfill
    \begin{subfigure}[t]{0.325\textwidth}
        \centering
        \includegraphics[width=0.49\textwidth]{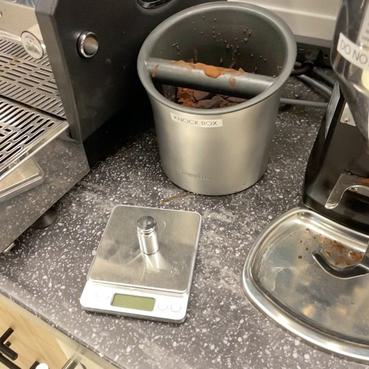} 
        \includegraphics[width=0.49\textwidth]{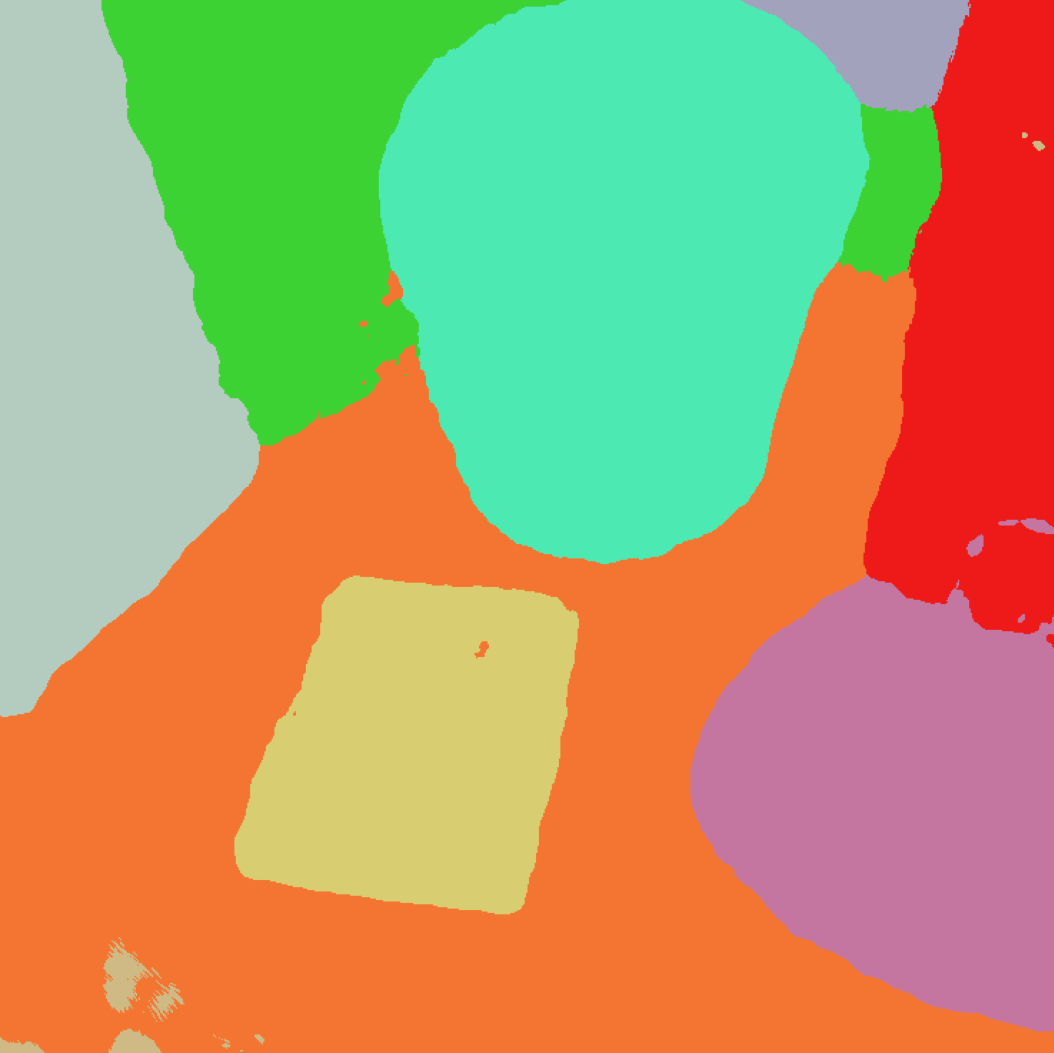}
        \caption{\textit{Espresso}}
    \end{subfigure}\hfill
    \begin{subfigure}[t]{0.325\textwidth}
        \centering
        \includegraphics[width=0.49\textwidth]{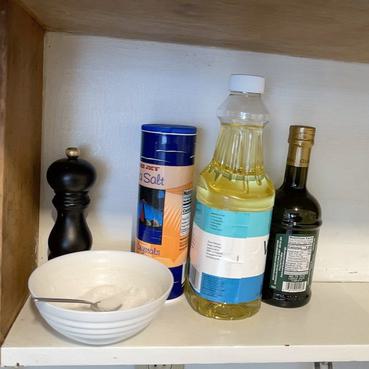}
        \includegraphics[width=0.49\textwidth]{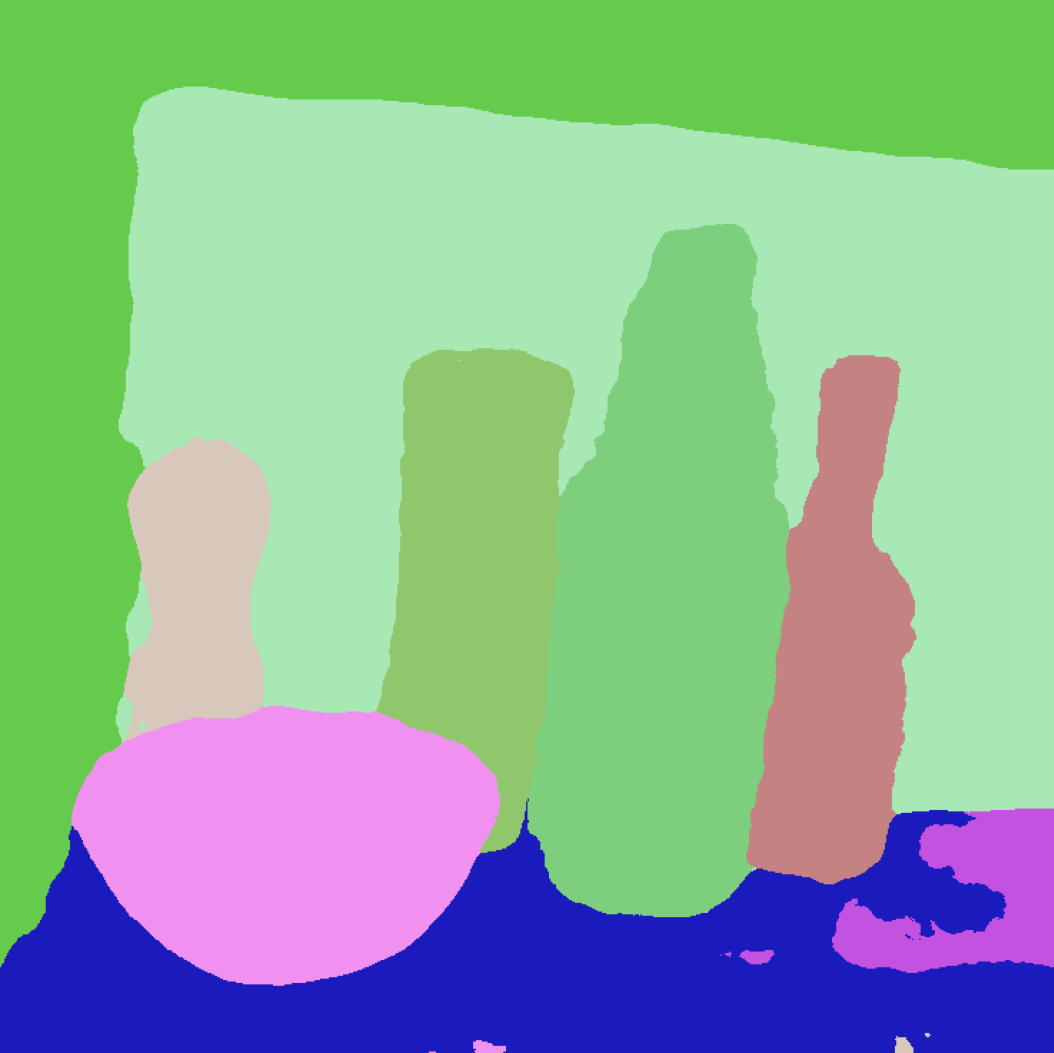}
        \caption{\textit{Waldo kitchen}}
    \end{subfigure}
    \caption{Additional results on the LERF~\cite{Kerr23} dataset. In some cases, high ambiguity in the mask generation can still lead to artifacts or fuzzy segmentations, as can be seen at the bottom right of the \textit{Waldo kitchen} scene.}
    \label{fig:lerfdata}
\end{figure*}

\begin{figure}[t]
    \centering
    \begin{subfigure}[t]{0.45\textwidth}
        \hfill
        \includegraphics[width=0.48\textwidth]{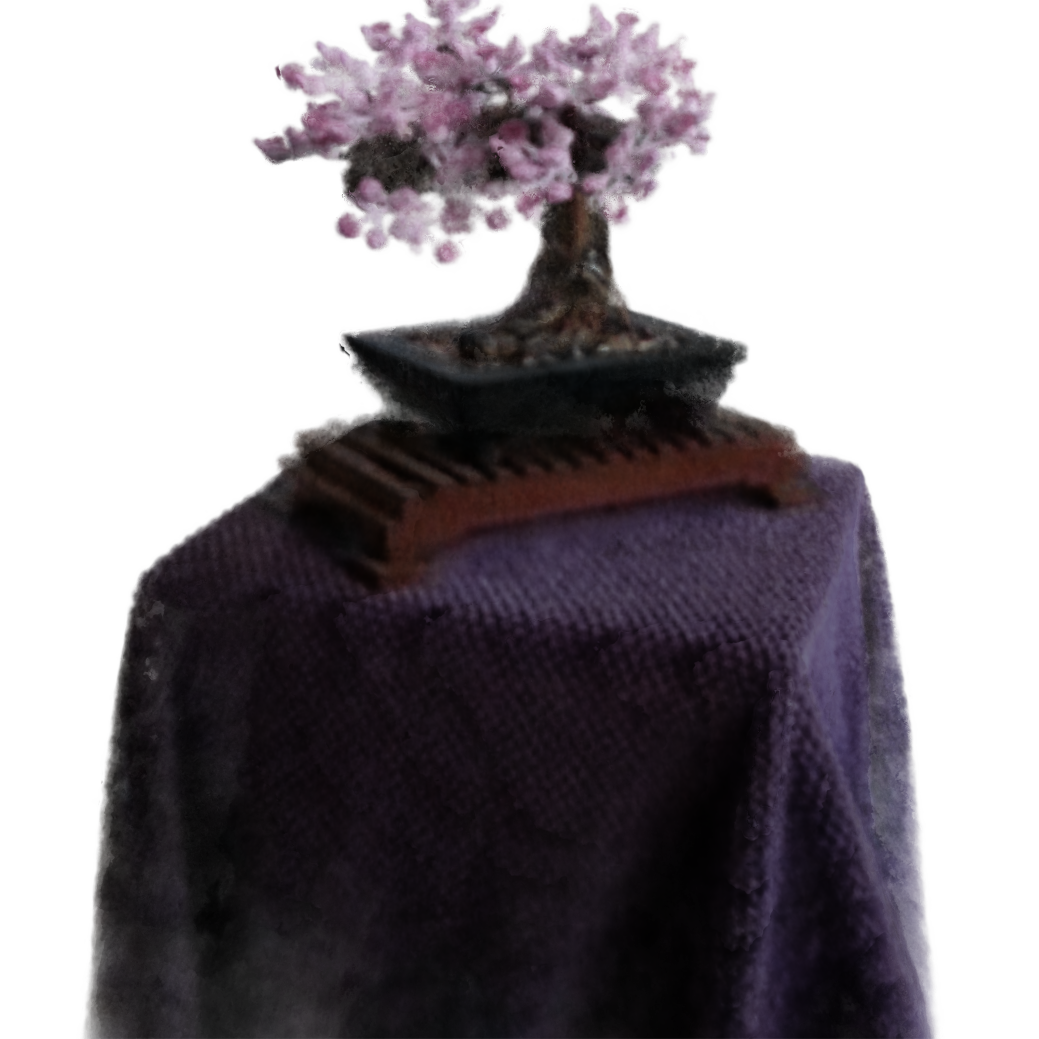}
        \hfill
        \includegraphics[width=0.48\textwidth]{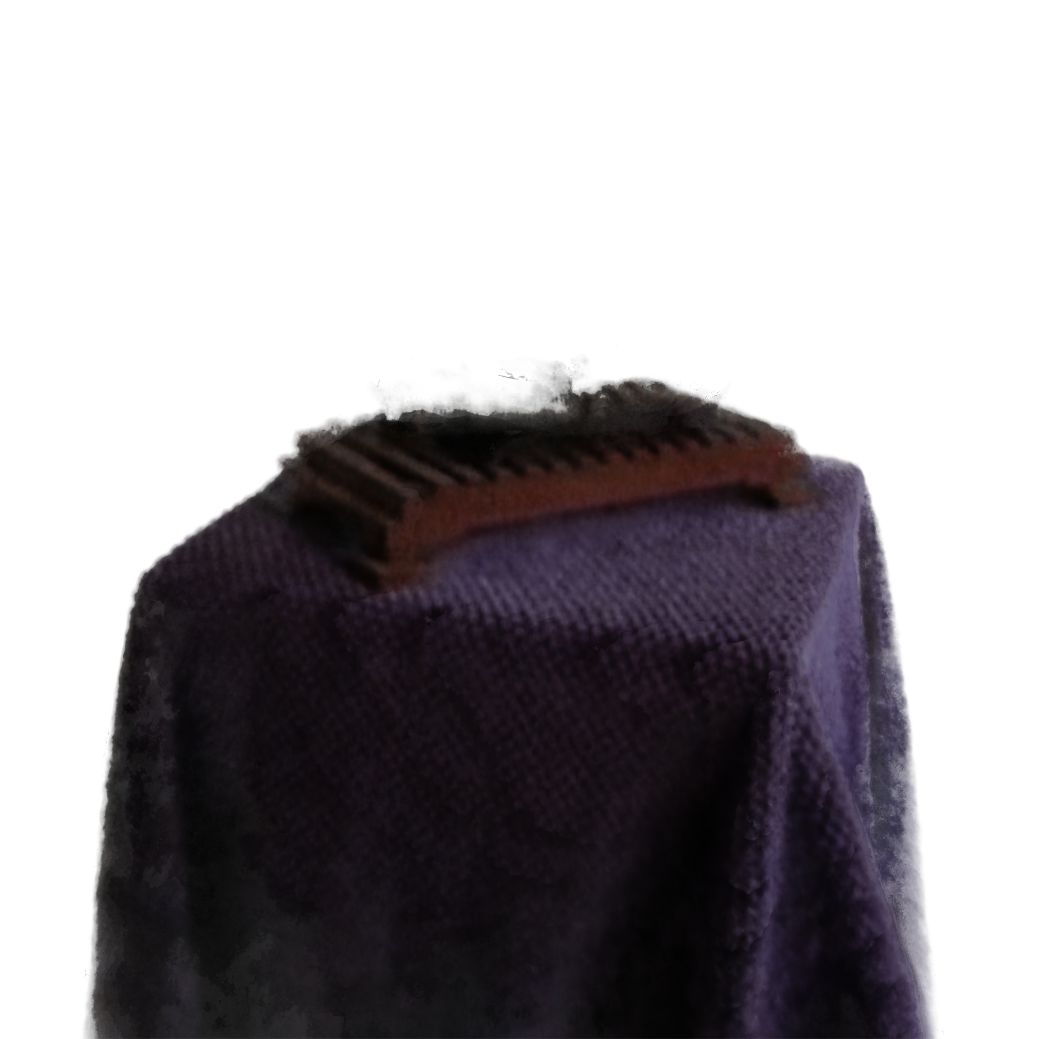}
        \hfill
        \caption{Object removal}
    \end{subfigure} \vspace{0.0em}\\
    \begin{subfigure}[t]{0.45\textwidth}
        \hfill
        \includegraphics[width=0.48\textwidth]{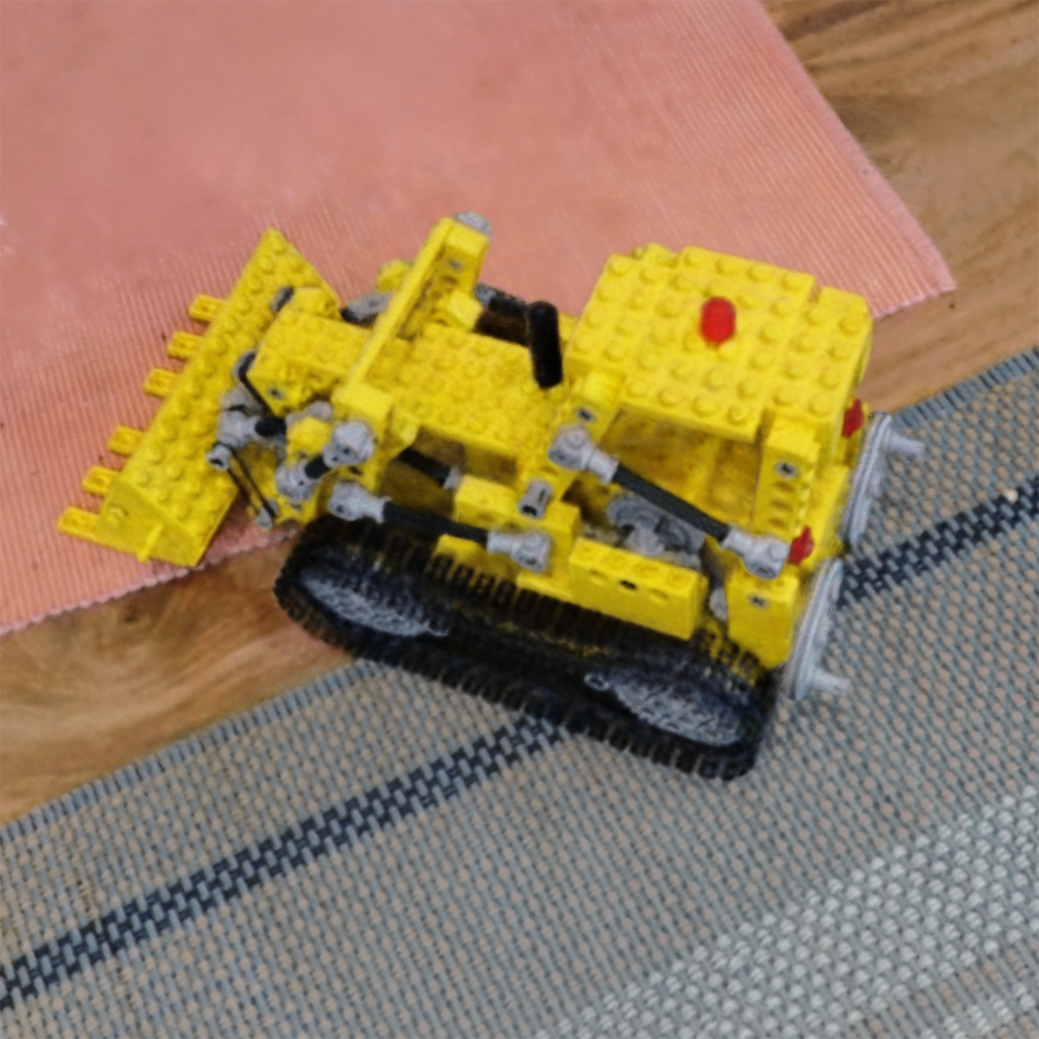} \hfill
        \includegraphics[width=0.48\textwidth]{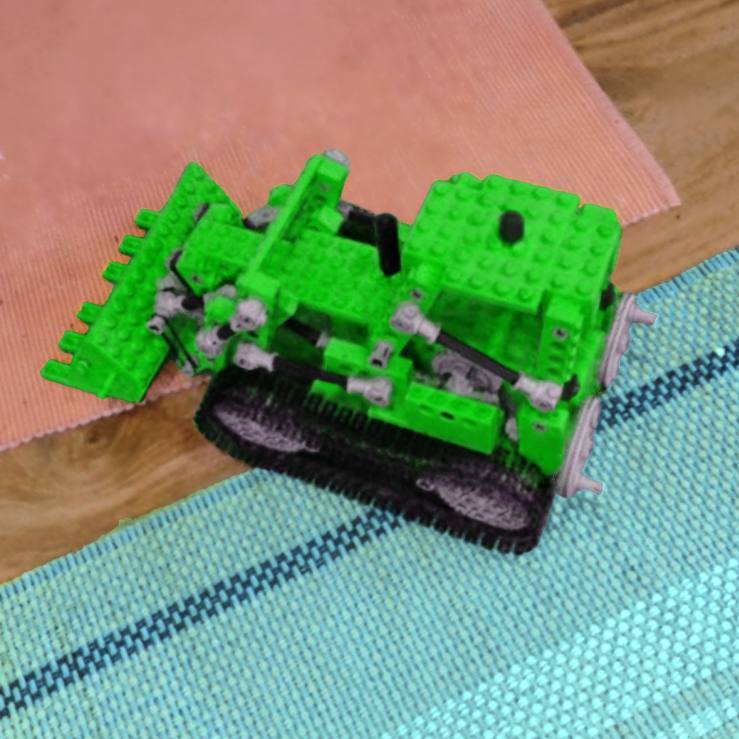}
        \hfill
        \caption{Color editing}
    \end{subfigure} \vspace{0.0em}\\
    \begin{subfigure}[t]{0.45\textwidth}
        \hfill
        \includegraphics[width=0.48\textwidth]{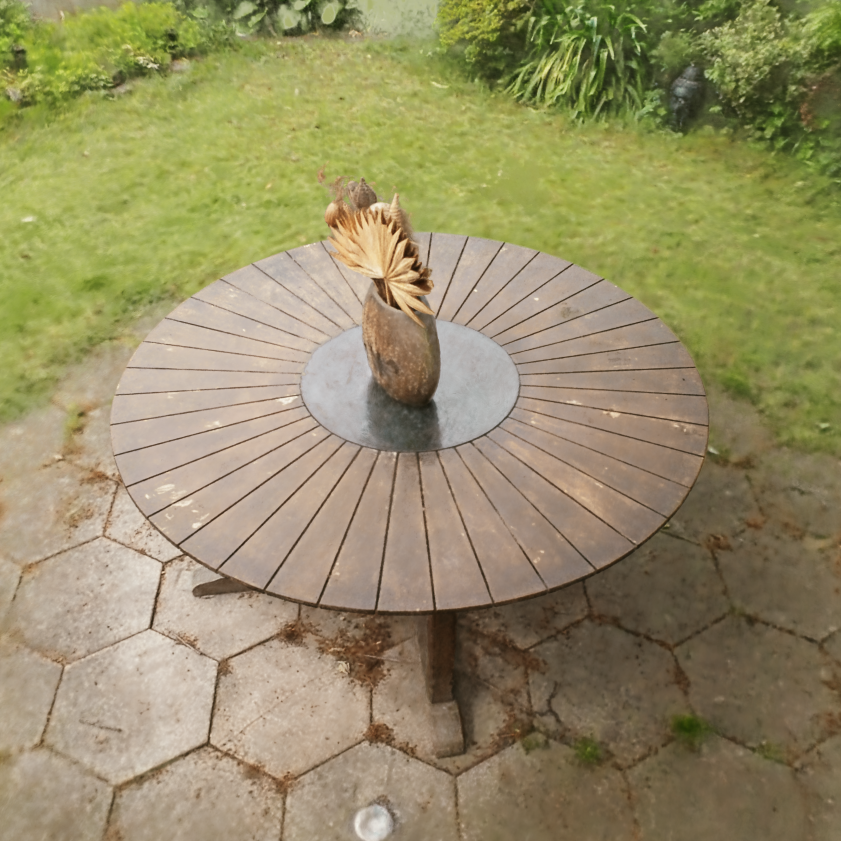}
        \hfill
        \includegraphics[width=0.48\textwidth]{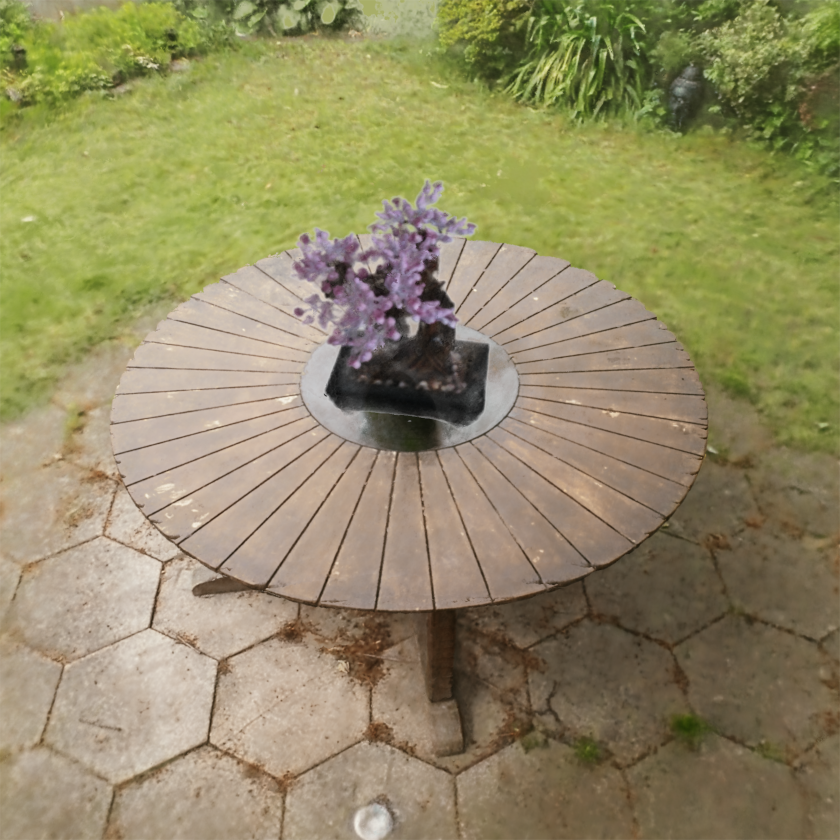}
        \hfill
        \caption{Composition}
    \end{subfigure}\hfill
    \caption{\small {\bf Applications.} Given some images (\textit{left}), we perform simple editing operations to synthesize new images (\textit{right}) by conditioning radiance field rendering with the object field value.}\vspace{-1em}
    \label{fig:applications}
\end{figure}

\parag{Results.}

We provide a qualitative comparison in \cref{fig:comparison0} and a quantitative one in \cref{tab:comparison}. Our approach outperforms the baselines by a large margin in all the metrics. In particular, Instance-NeRF fails when evaluated on in-the-wild scenes, which are more visually complex than its training data and filled with previously unseen objects. The 3D point cloud segmentation network used by Instance-NeRF often misses or wrongly merges different objects, leading to simplistic segmentations.

Similarly, Panoptic Lifting relies on a semantic panoptic segmentation of the scene~\cite{Cheng22} and fails to properly segment novel objects for in-the-wild scenes. The results of our implementation of PL+SAM, which uses the same masks as our method as supervision, show that it is unable to derive a meaningful 3D segmentation from class-agnostic masks. In complex scenes, several objects are wrongfully clustered together or associated with the background.

\cd{Additionally, we evaluate the masks produced by SAM from a single 2D image in \cref{tab:comparison}. As shown in \cref{fig:statement}(a), these masks are often noisy and fail to capture all objects. We observe that our method successfully learns from multi-view masks and outperforms the single-view supervision.}

\subsection{Ablation study}

To validate the impact of the different components of our method, we perform an ablation study. 
We remove the matching $\gamma$ and instead match masks and objects by choosing the slot that maximizes the affinity $\alpha$ independently from each other, such that the same object slot can be assigned to several masks. This leads to a sharp drop in our evaluation metrics from $79.24\%$ to $57.04\%$ IoU, as the same object slot often captures several objects in \cref{fig:ablation0}(b).

We also ablate the regularization loss $\mathcal{L}_{TV}$ and compare the produced segmentations in \cref{fig:ablation0}(c). We observe a drop in IoU down to $75.53\%$ that can be attributed to the noisy object field learned without our additional regularization. 
Interestingly, we also observed in our experiments that the addition of $\mathcal{L}_{TV}$ helped prevent different objects from using the same object slot. 
We view this as a consequence of our regularization, which prevents the optimization from getting trapped in a local minimum.
We further observe that without $\mathcal{L}_{TV}$, confidence remains low and noisy, leading to fuzzy segmentations in \cref{fig:ablation0}(c)
and making 3D object extraction less robust. In contrast, our method successfully identifies objects in the scene in separate object slots.

\subsection{Applications}

Since our method can segment 3D scenes into individual objects, the 3D reconstructions can be used in downstream applications. An object can be extracted by conditioning the rendering of the model in \cref{eq:nerf} to only render the associated slot. As shown in \cref{fig:applications}(a), we can remove the bonsai from the scene by removing all its parts. This operation reveals uncovered geometry under the removed object, which can be fuzzy as it was not supervised during NeRF training. This can easily be enhanced in post-processing by inpainting~\cite{Mirzaei23a,Wang23}. In~\cref{fig:applications}(b), we identify objects and change their color channels independently. We can also assemble objects from several scenes and render them jointly, as illustrated by~\cref{fig:applications}(c).

%% file: fig/segmentations.tex

\begin{figure*}[t]
    \centering\hfill
    \begin{subfigure}[t]{0.16\textwidth}
        \includegraphics[width=\textwidth]  
        {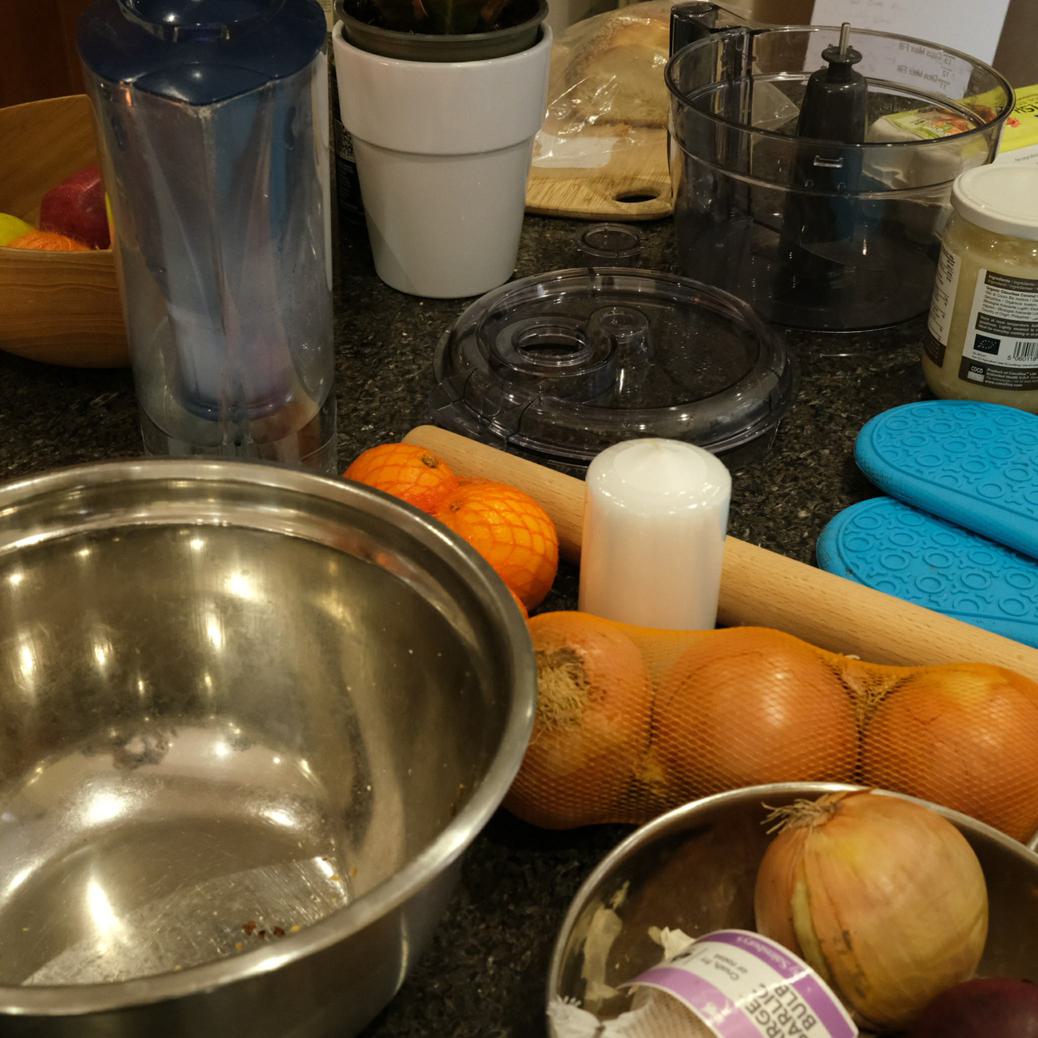}\vspace{0.13em}
        \includegraphics[width=\textwidth]  
        {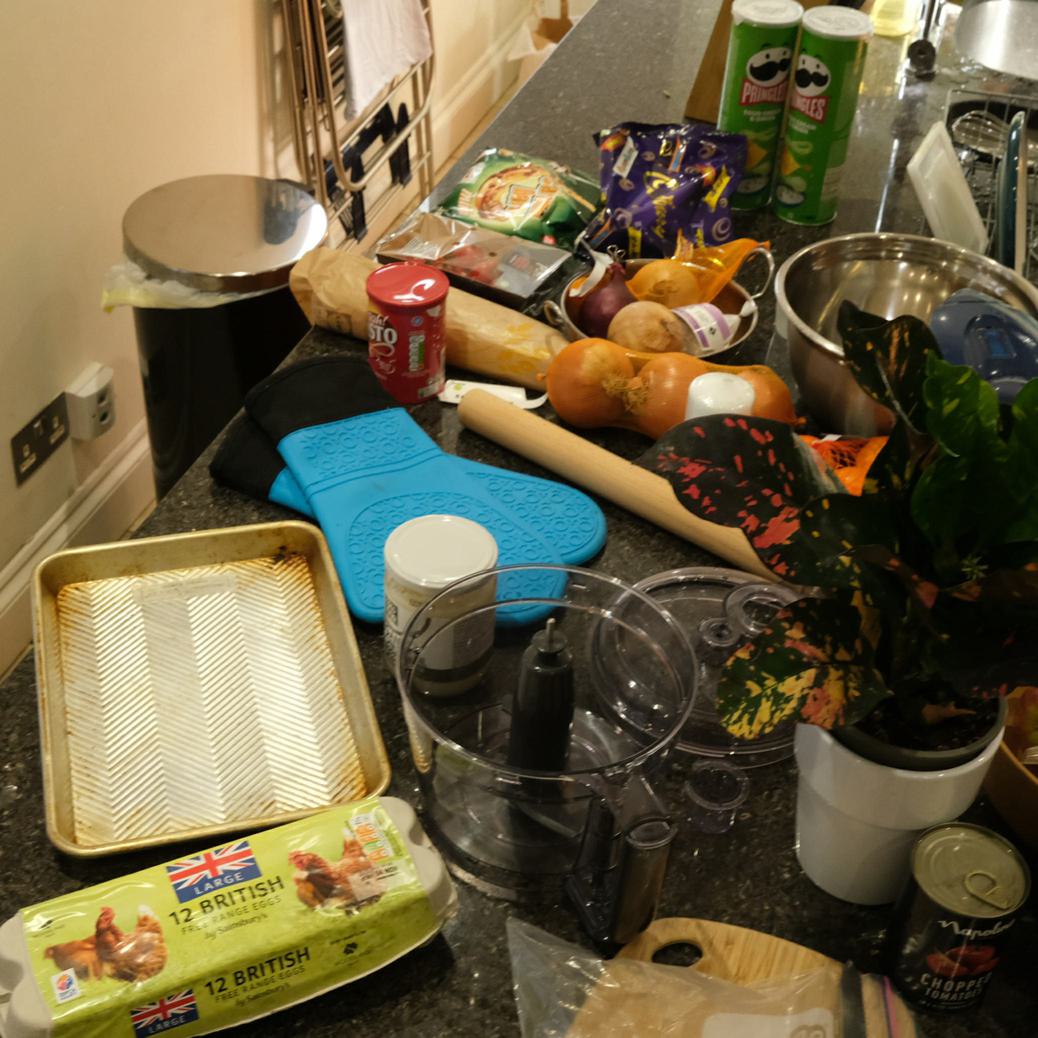}\vspace{0.13em}
        \includegraphics[width=\textwidth]
        {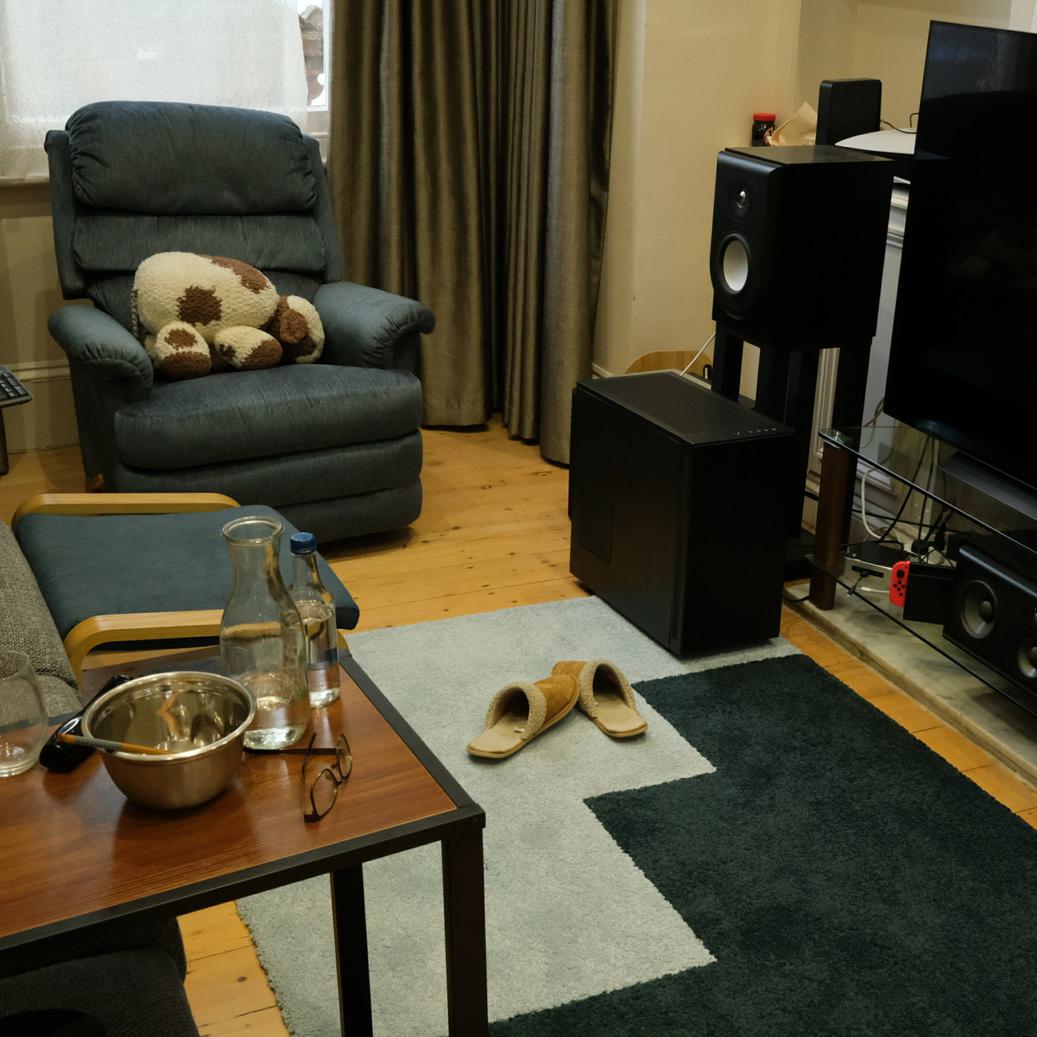}\vspace{0.13em}
        \includegraphics[width=\textwidth]
        {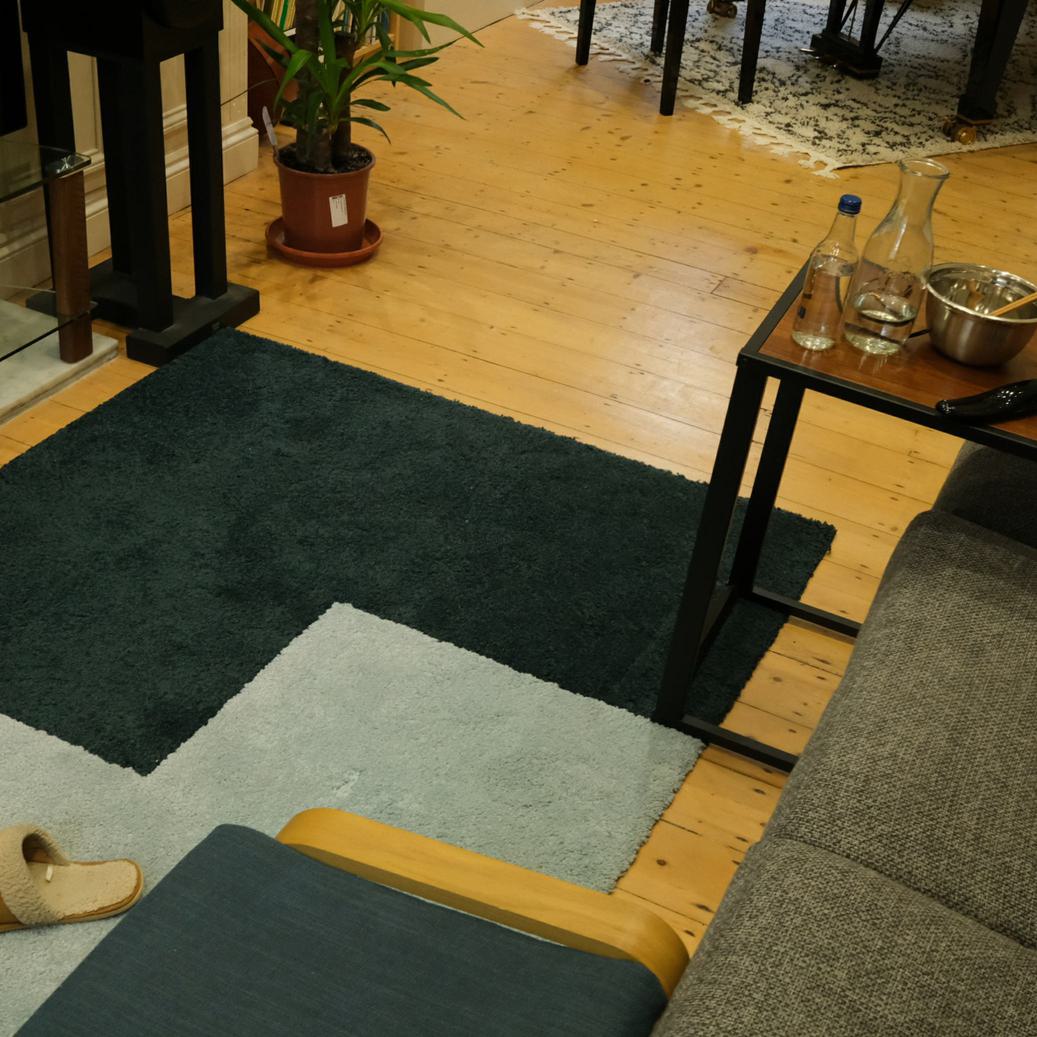}\vspace{0.13em}
        \caption*{Image}
    \end{subfigure}\hfill
    \begin{subfigure}[t]{0.16\textwidth}
        \includegraphics[width=\textwidth]  
        {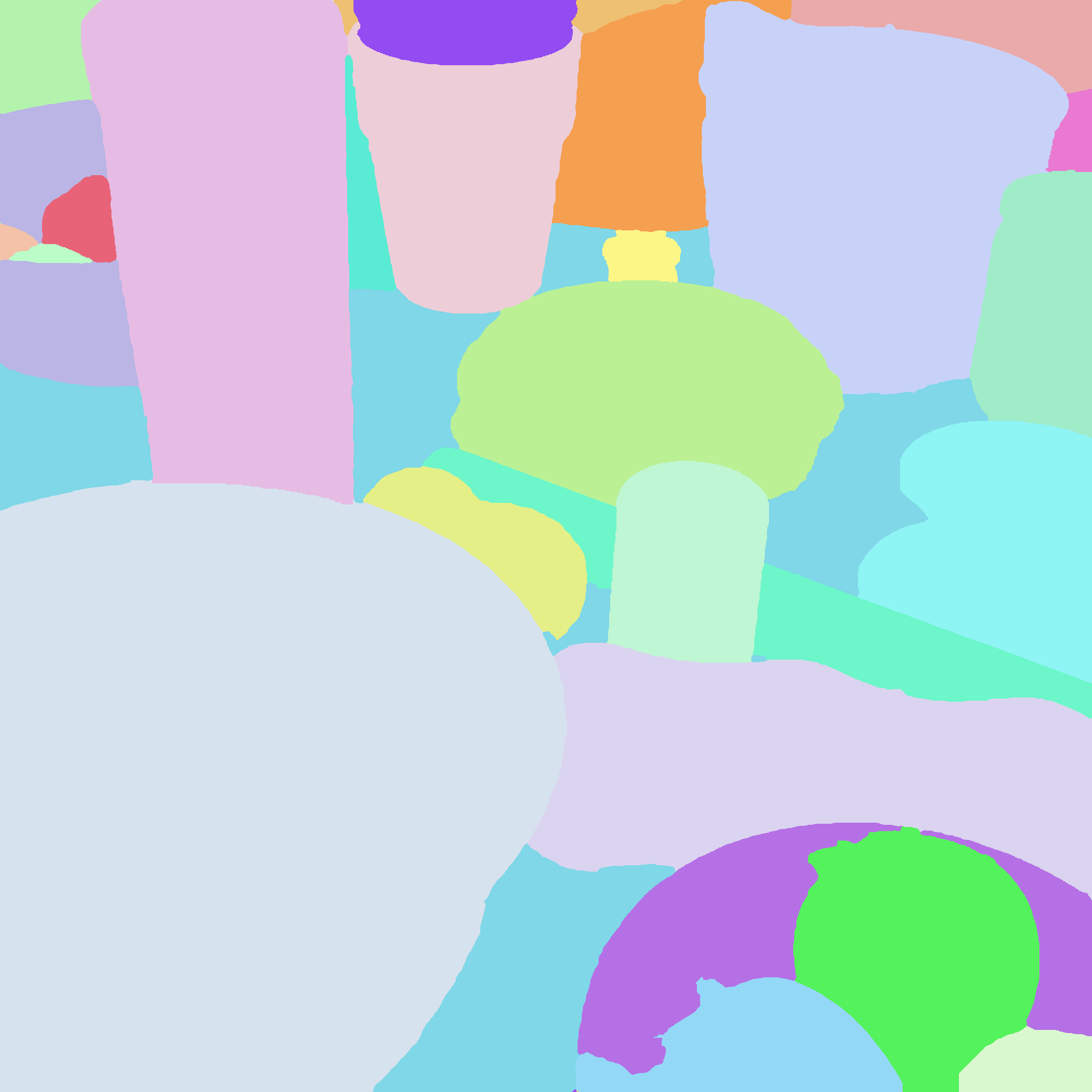}\vspace{0.13em}
        \includegraphics[width=\textwidth]  
        {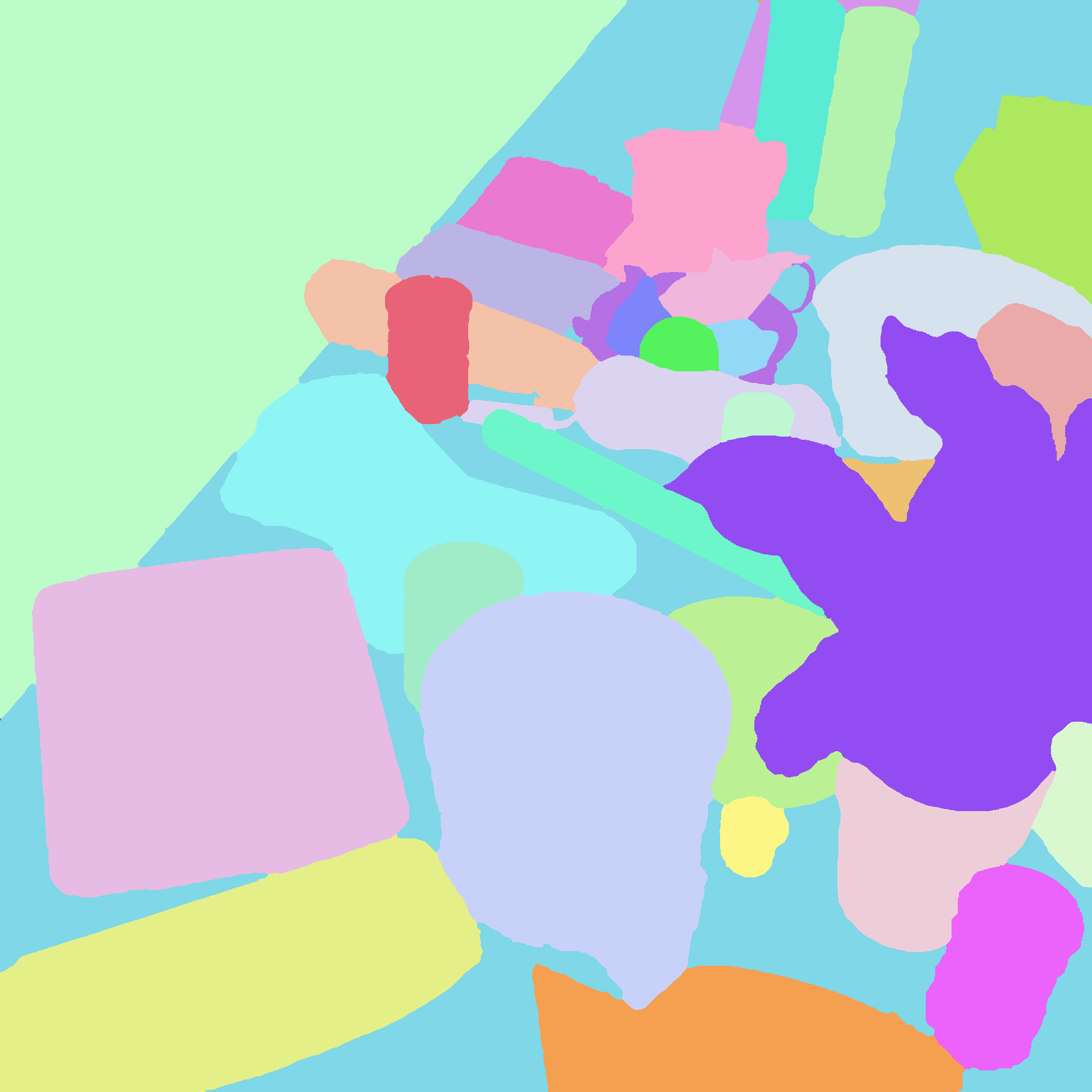}\vspace{0.13em}
        \includegraphics[width=\textwidth]
        {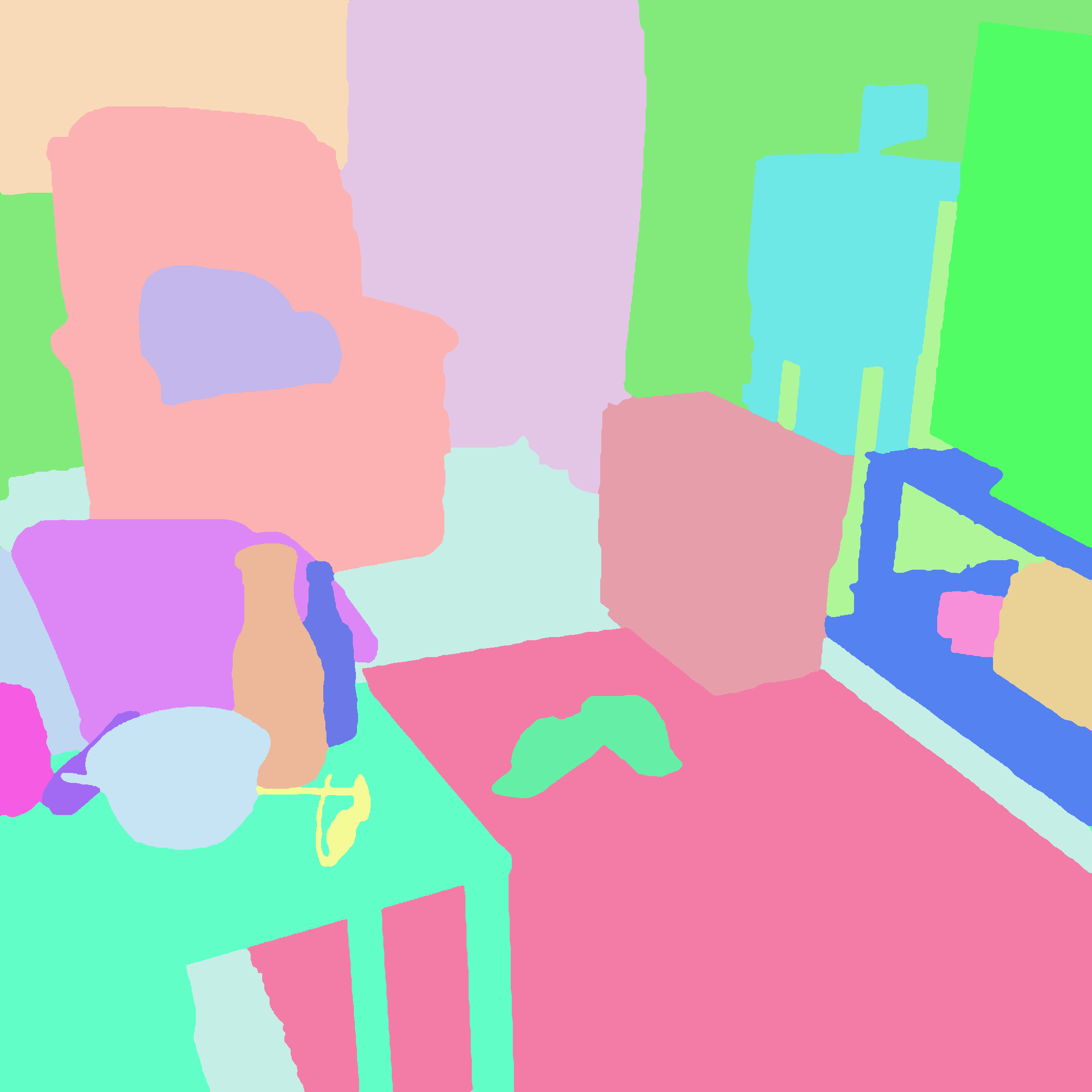}\vspace{0.13em}
        \includegraphics[width=\textwidth]
        {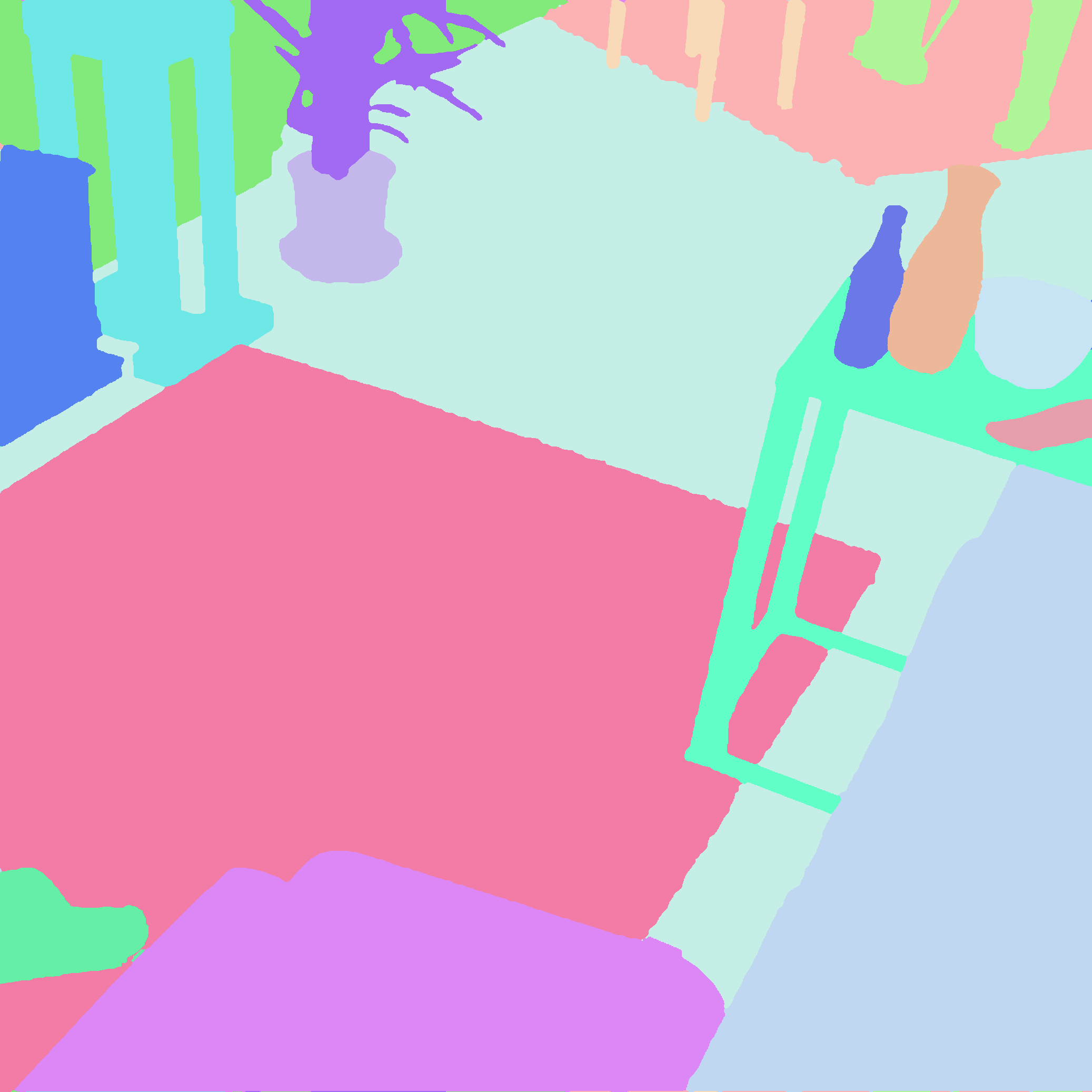}\vspace{0.13em}
        \caption*{Ground truth}
    \end{subfigure}\hfill
    \begin{subfigure}[t]{0.16\textwidth}
        \includegraphics[width=\textwidth]  
        {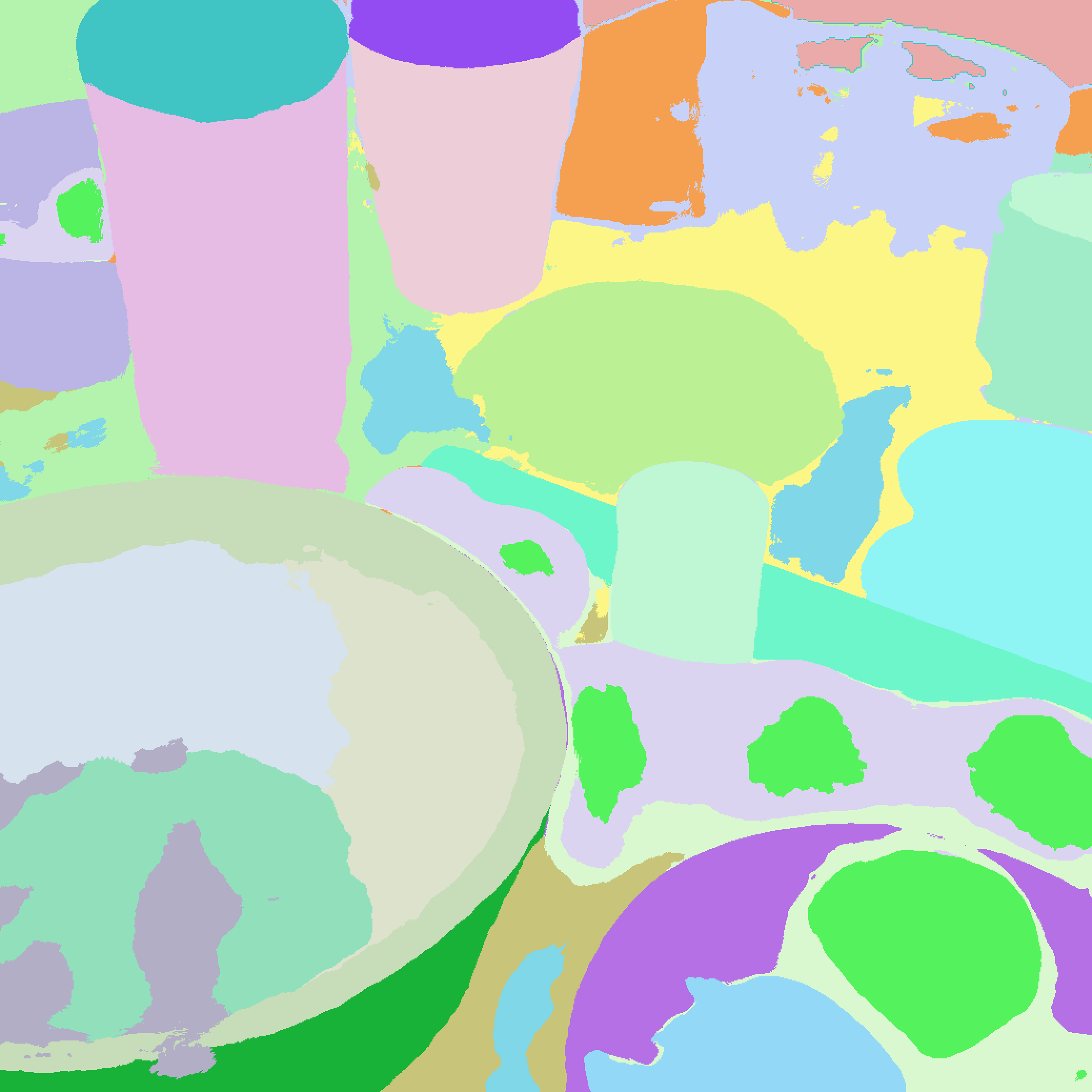}\vspace{0.13em}
        \includegraphics[width=\textwidth]  
        {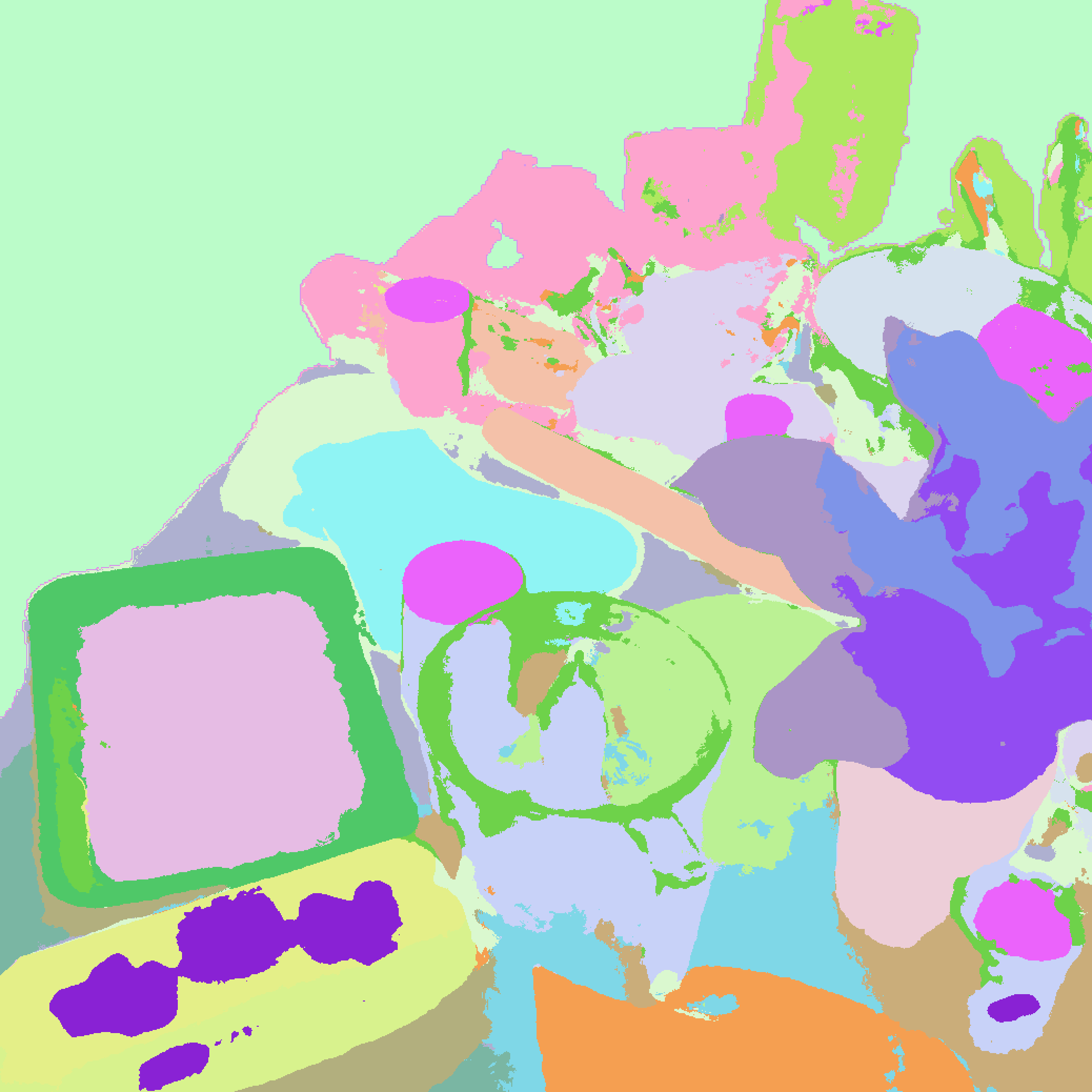}\vspace{0.13em}
        \includegraphics[width=\textwidth]
        {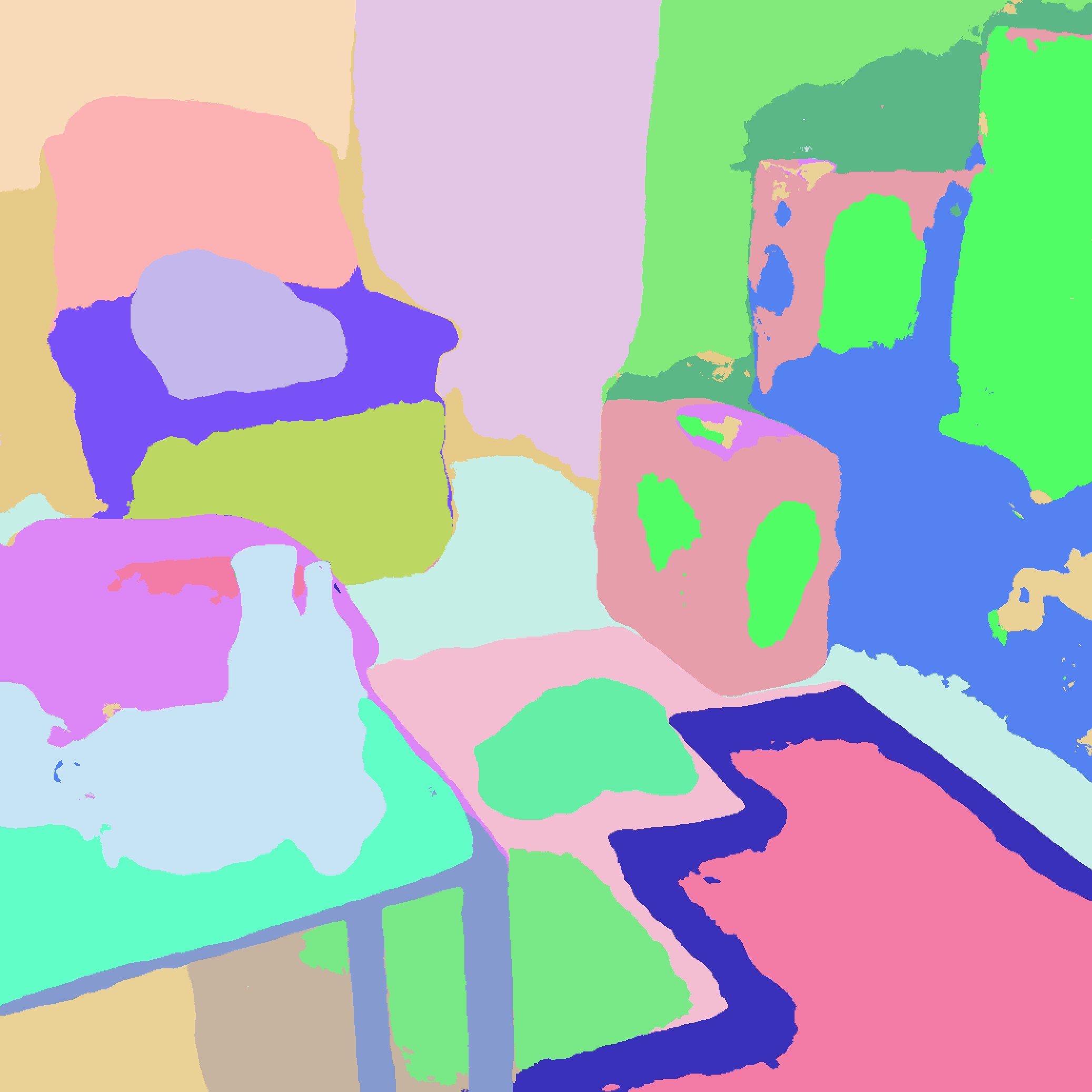}\vspace{0.13em}
        \includegraphics[width=\textwidth]
        {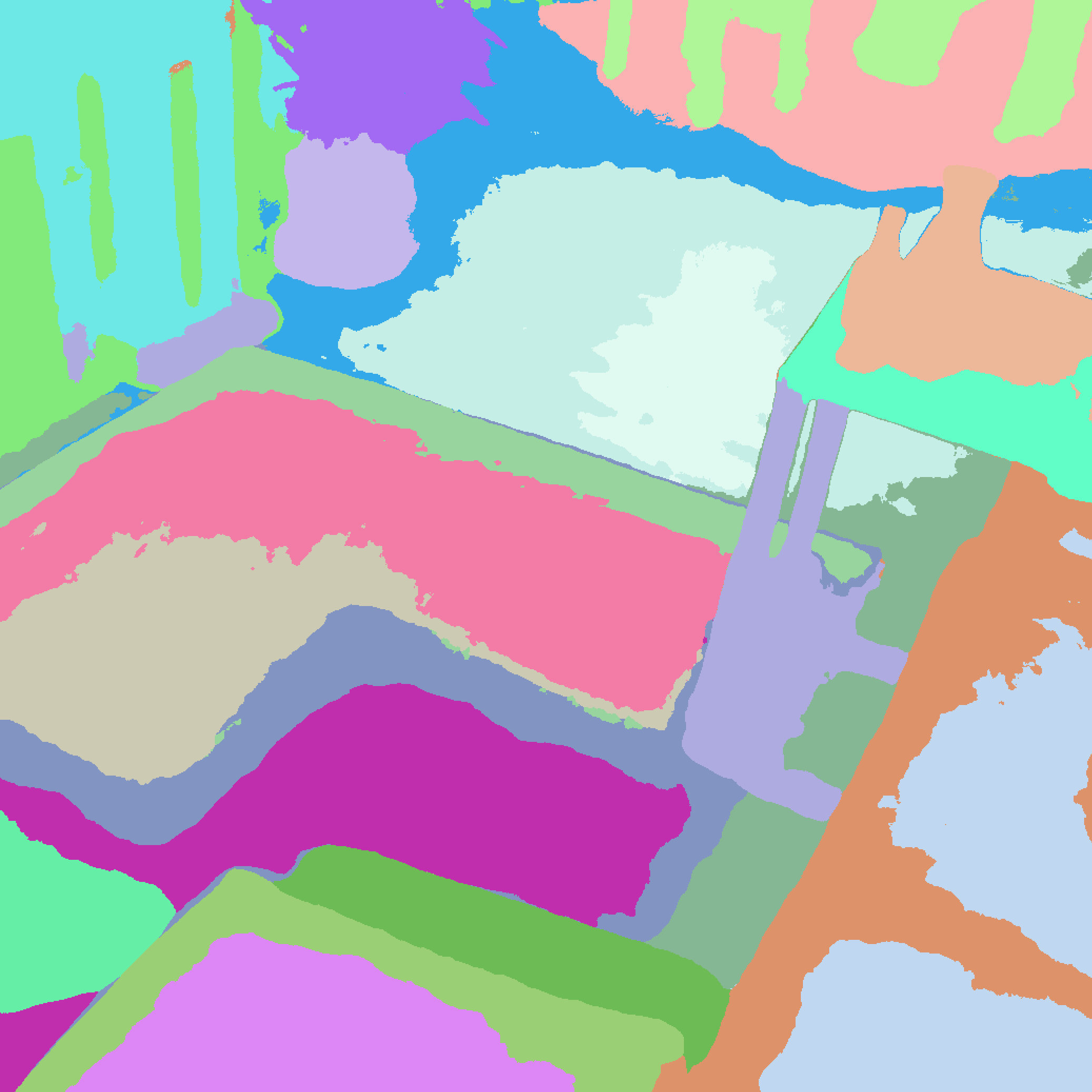}\vspace{0.13em}
        \caption*{DFFv2}
    \end{subfigure}\hfill
    \begin{subfigure}[t]{0.16\textwidth}
        \includegraphics[width=\textwidth]  
        {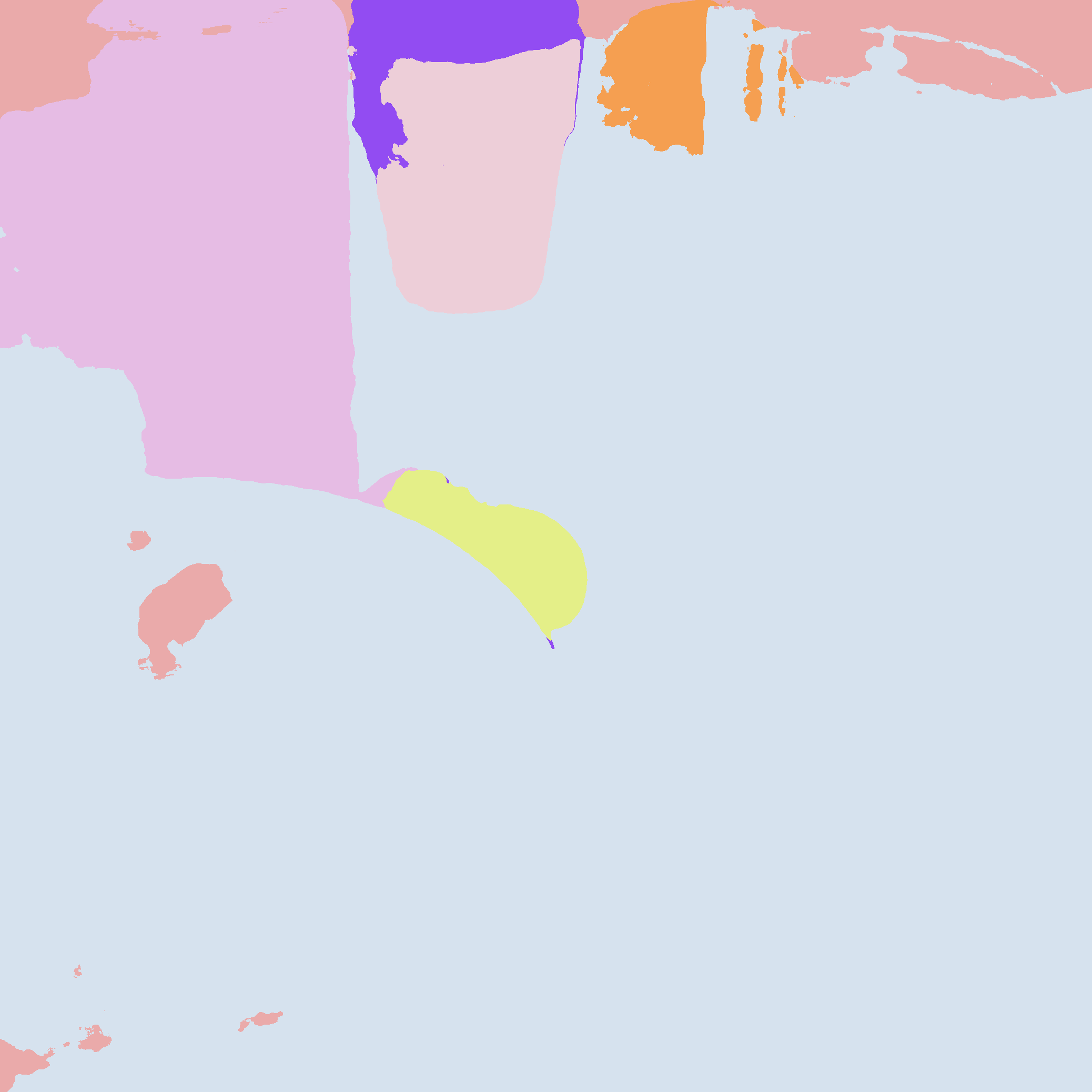}\vspace{0.13em}
        \includegraphics[width=\textwidth]  
        {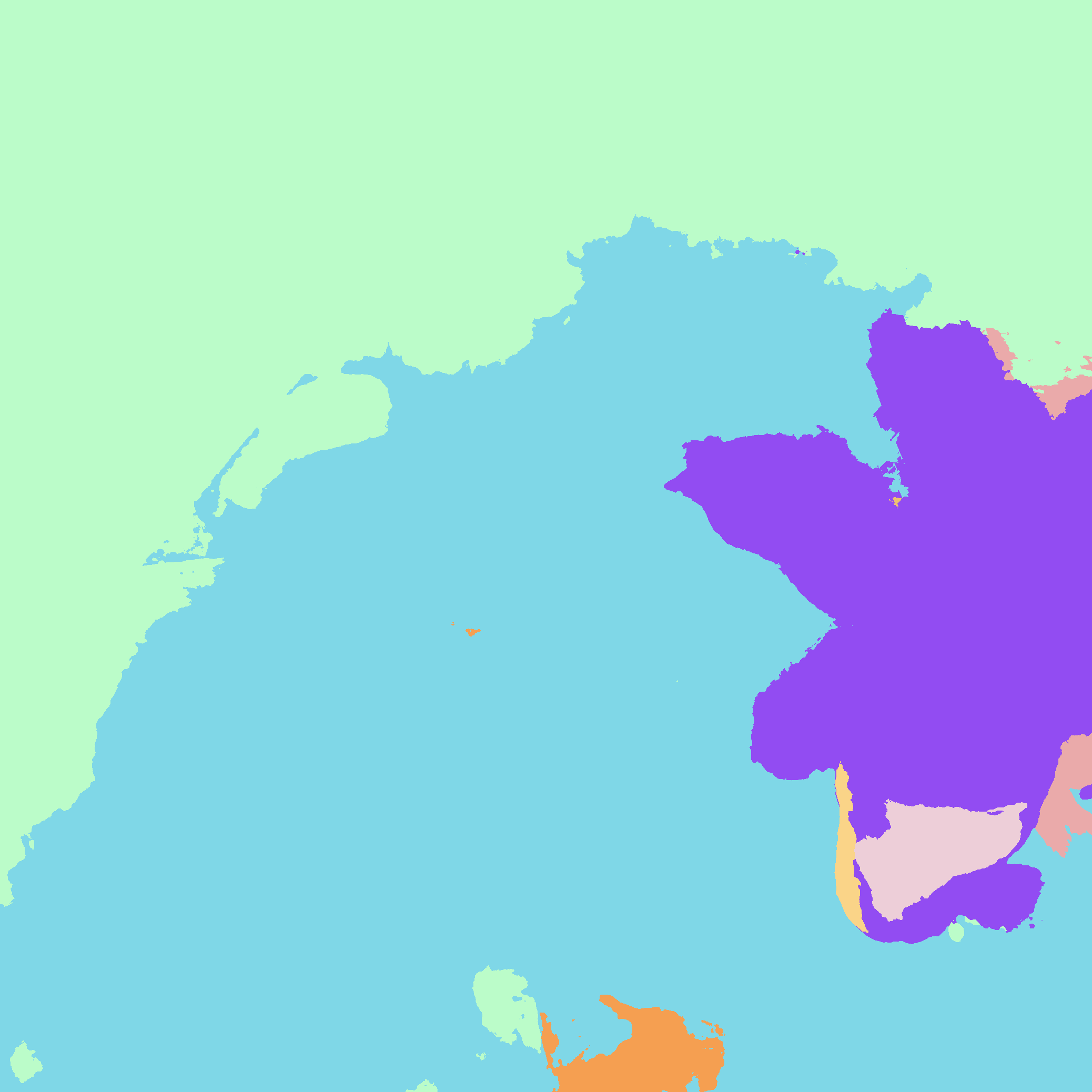}\vspace{0.13em}
        \includegraphics[width=\textwidth]
        {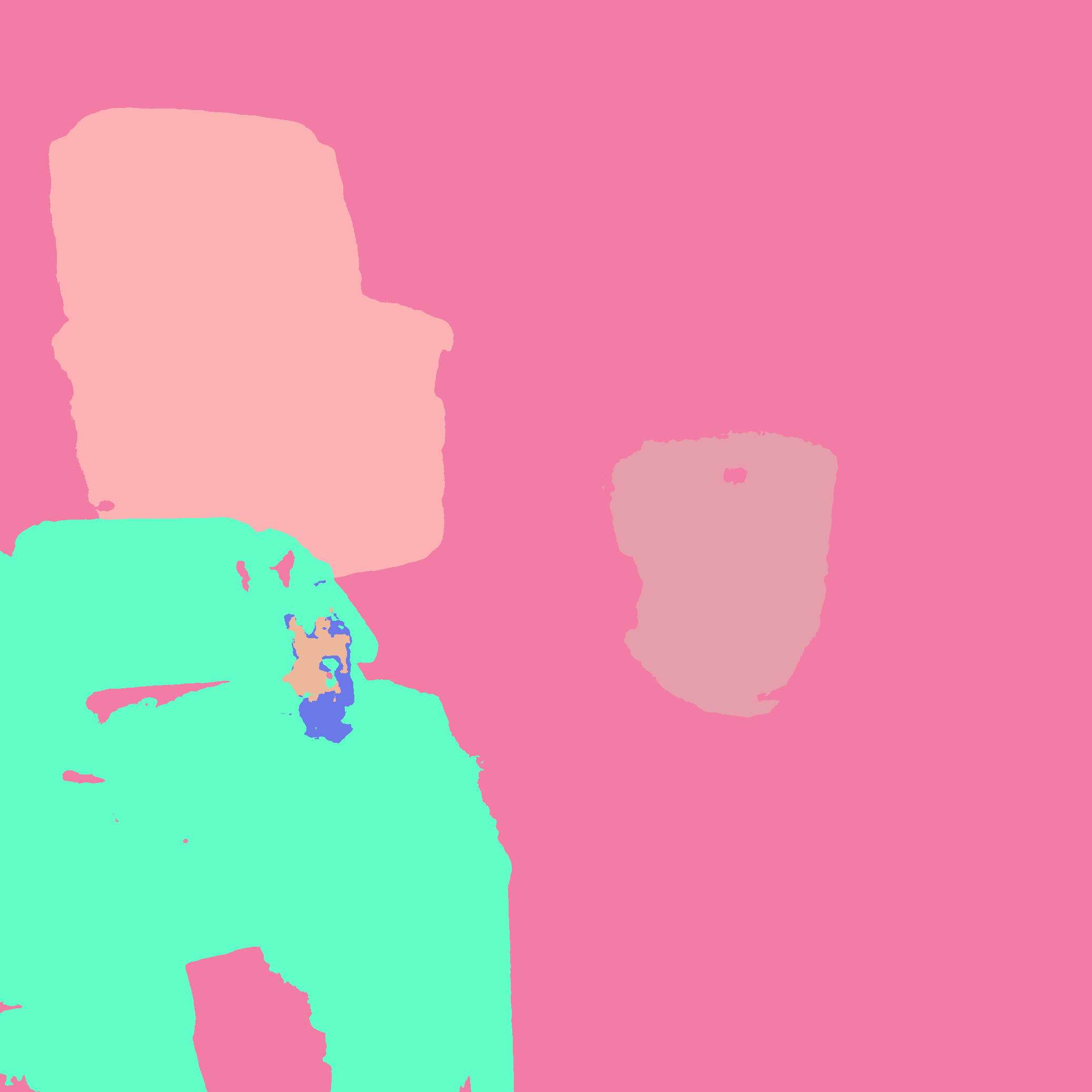}\vspace{0.13em}
        \includegraphics[width=\textwidth]
        {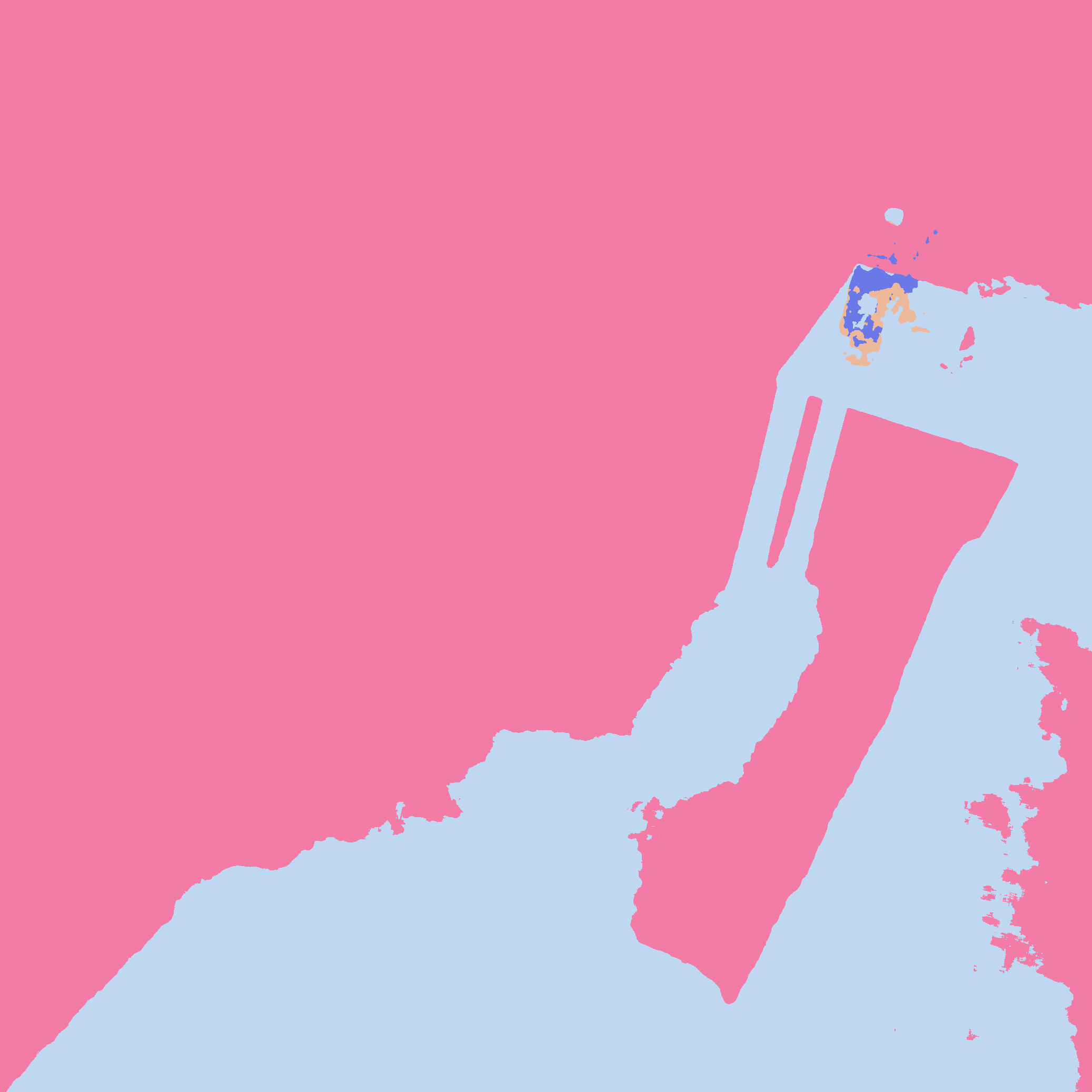}\vspace{0.13em}
        \caption*{\small{Instance-NeRF}}
    \end{subfigure}\hfill
    \begin{subfigure}[t]{0.16\textwidth}
        \includegraphics[width=\textwidth]  
        {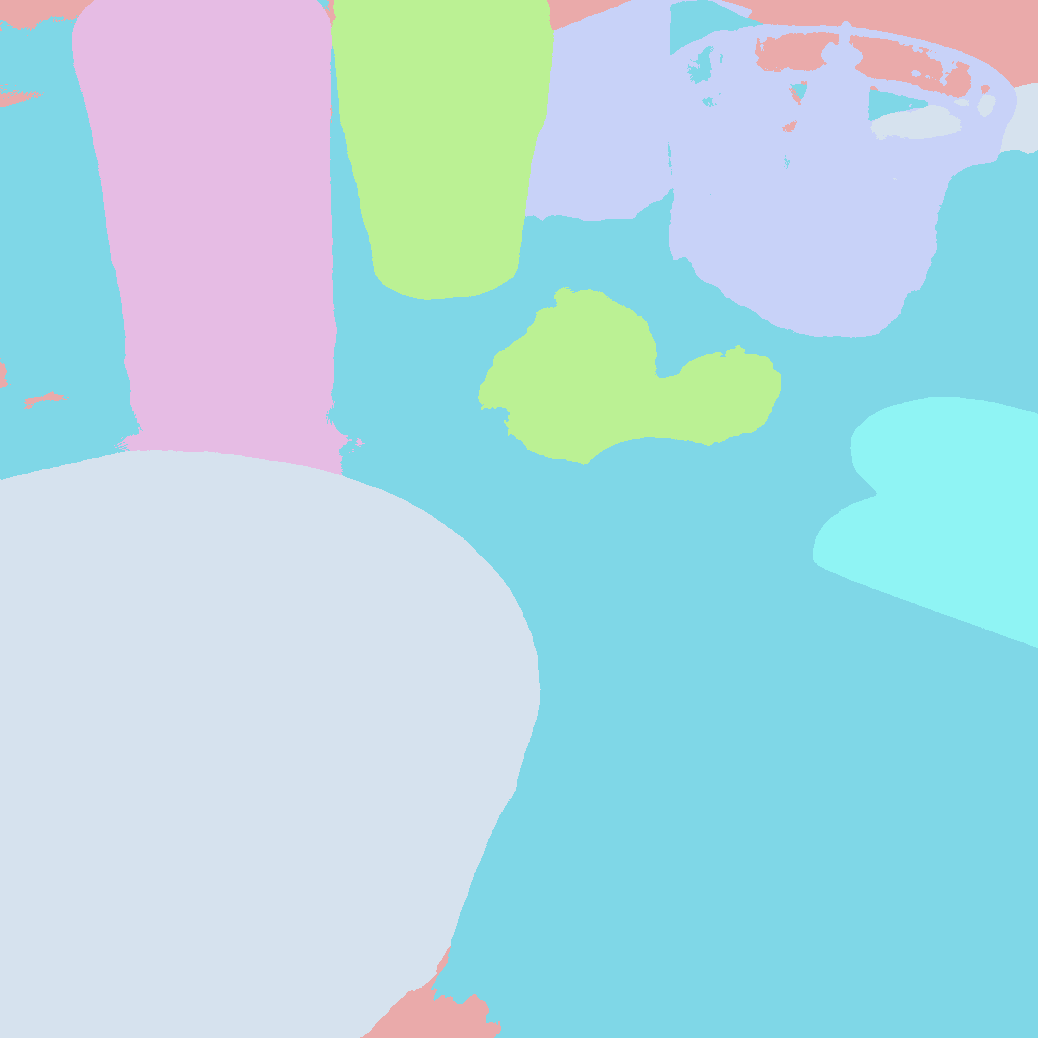}\vspace{0.13em}
        \includegraphics[width=\textwidth]  
        {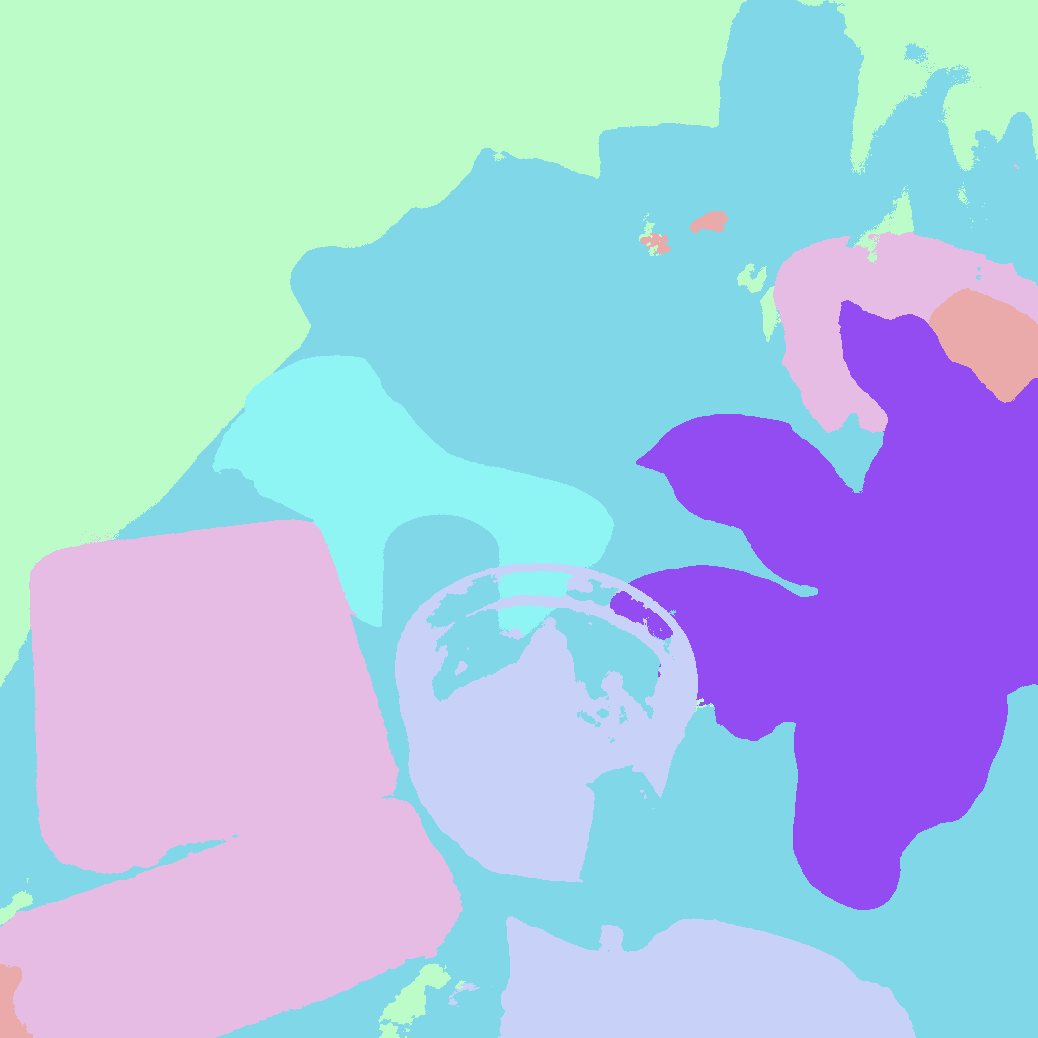}\vspace{0.13em}
        \includegraphics[width=\textwidth]
        {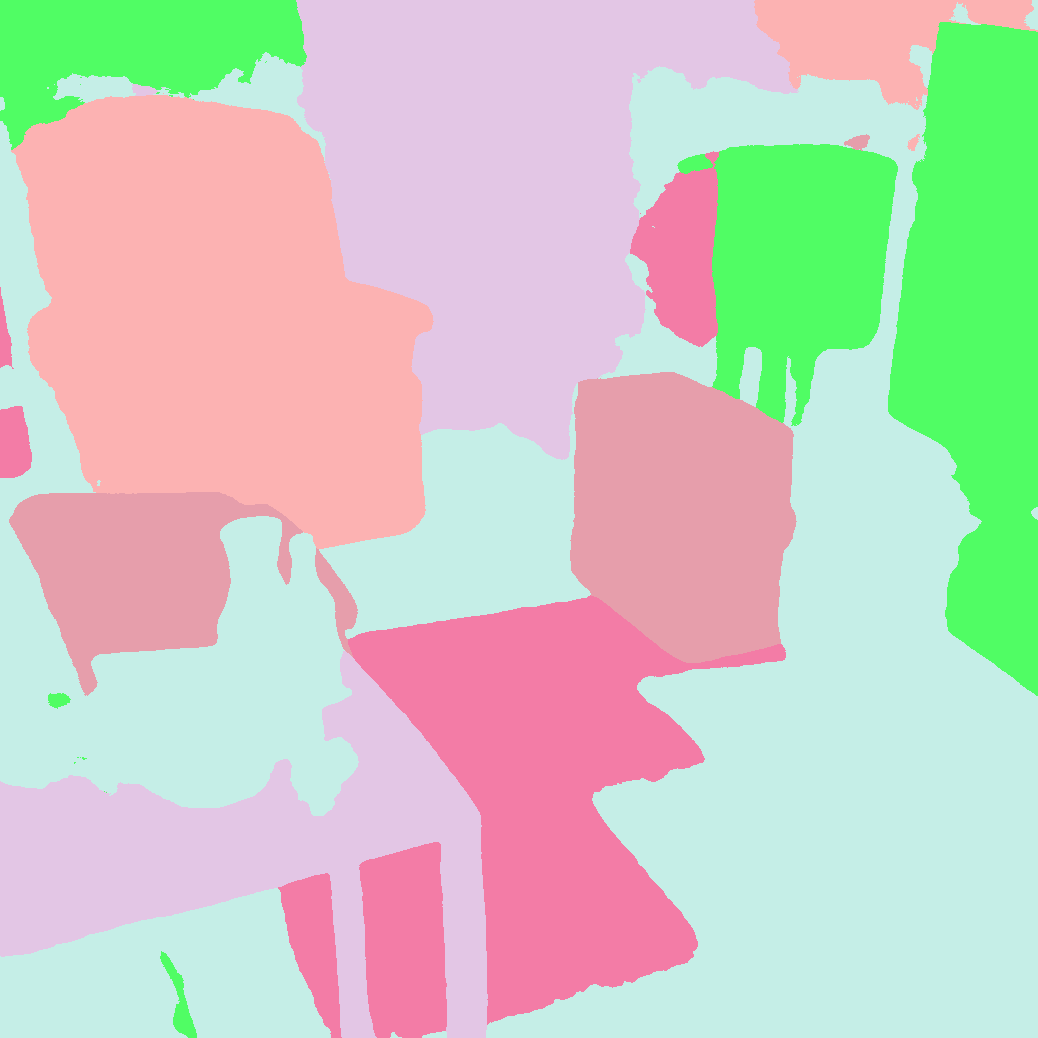}\vspace{0.13em}
        \includegraphics[width=\textwidth]
        {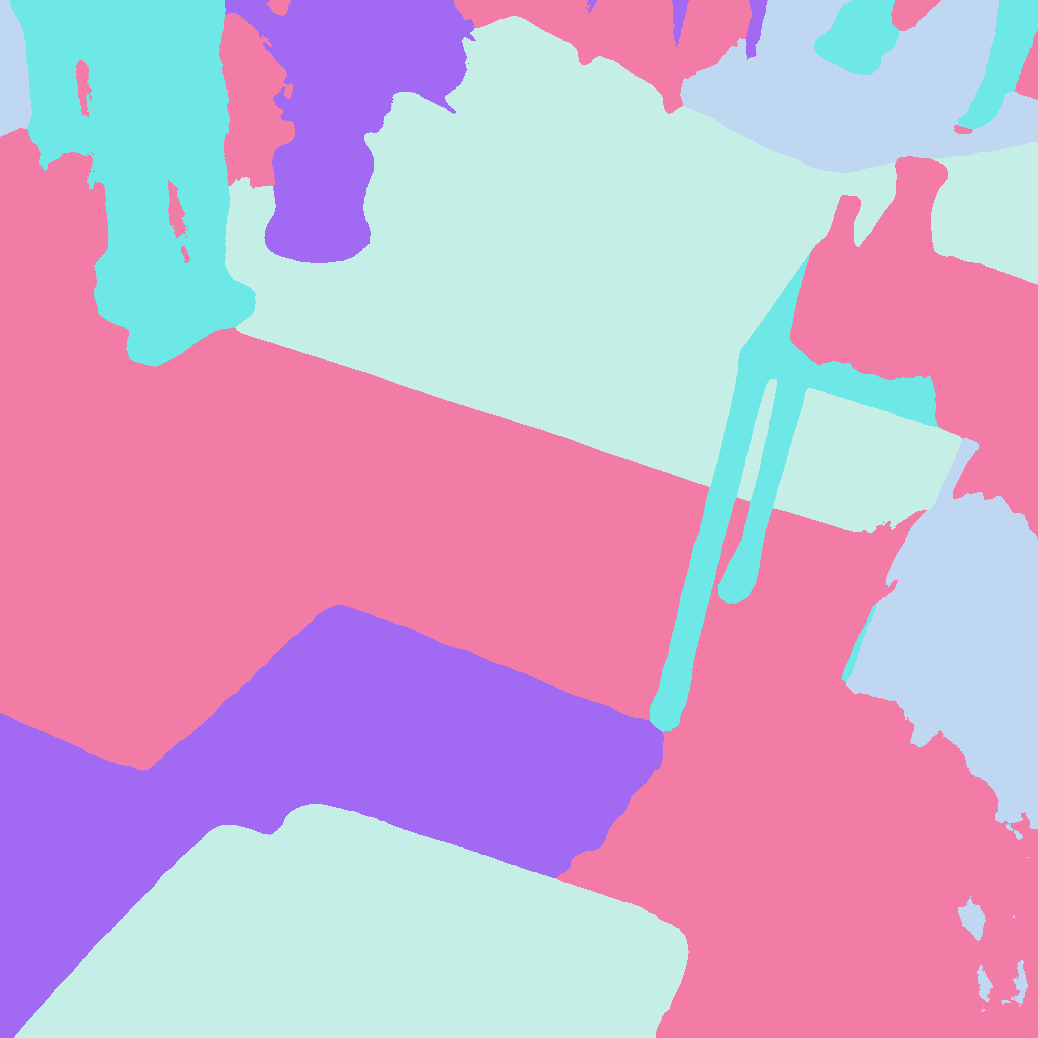}\vspace{0.13em}
        \caption*{\small{PL+SAM}}
    \end{subfigure}\hfill
    \begin{subfigure}[t]{0.16\textwidth}
        \includegraphics[width=\textwidth]  
        {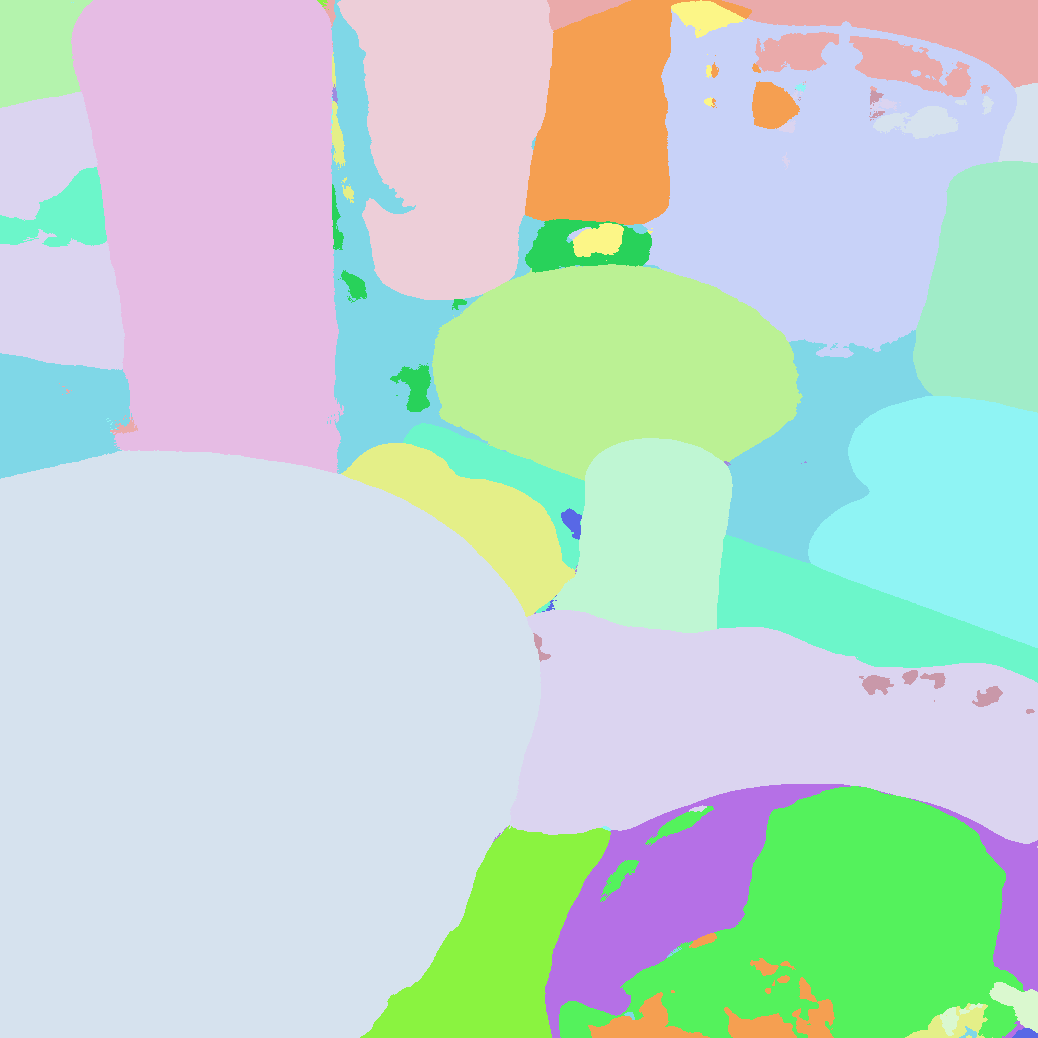}\vspace{0.13em}
        \includegraphics[width=\textwidth]  
        {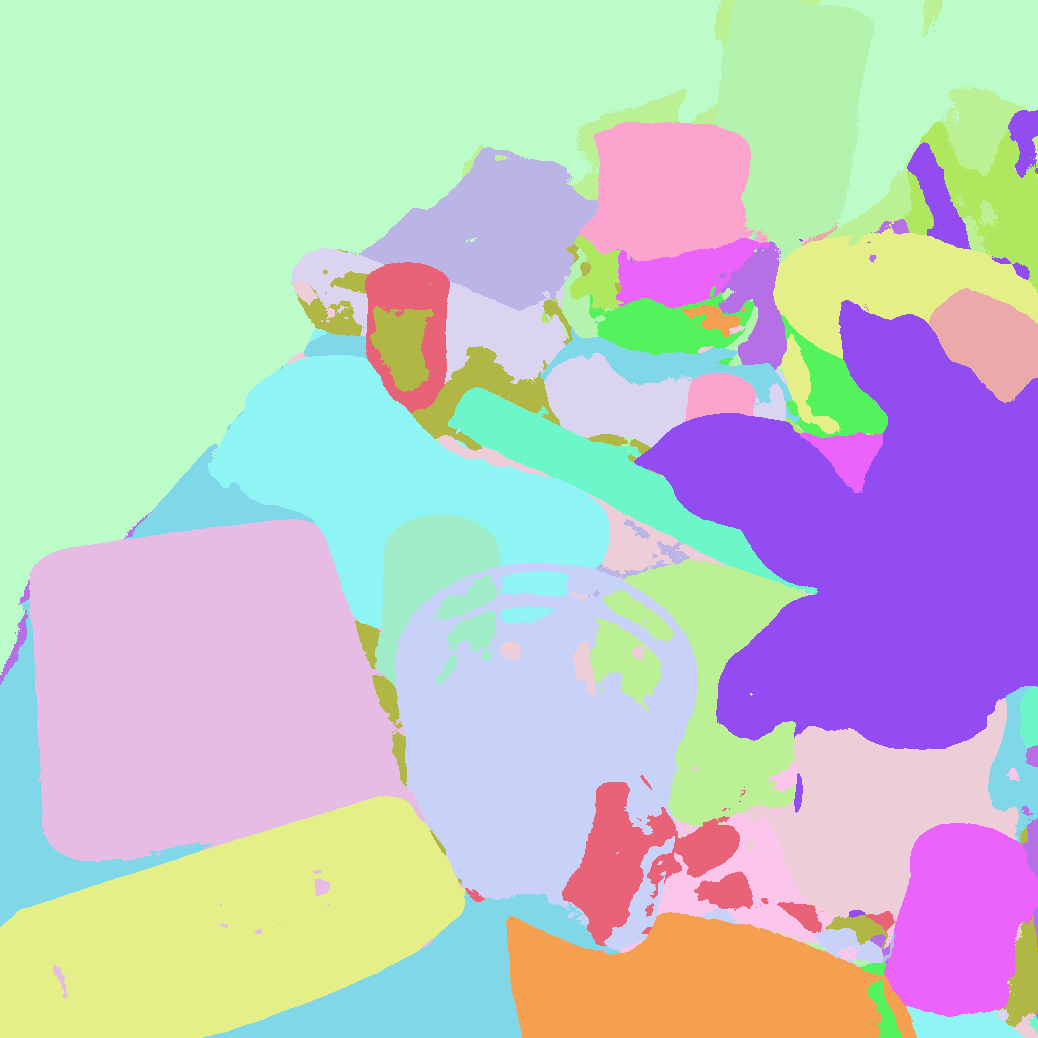}\vspace{0.13em}
        \includegraphics[width=\textwidth]
        {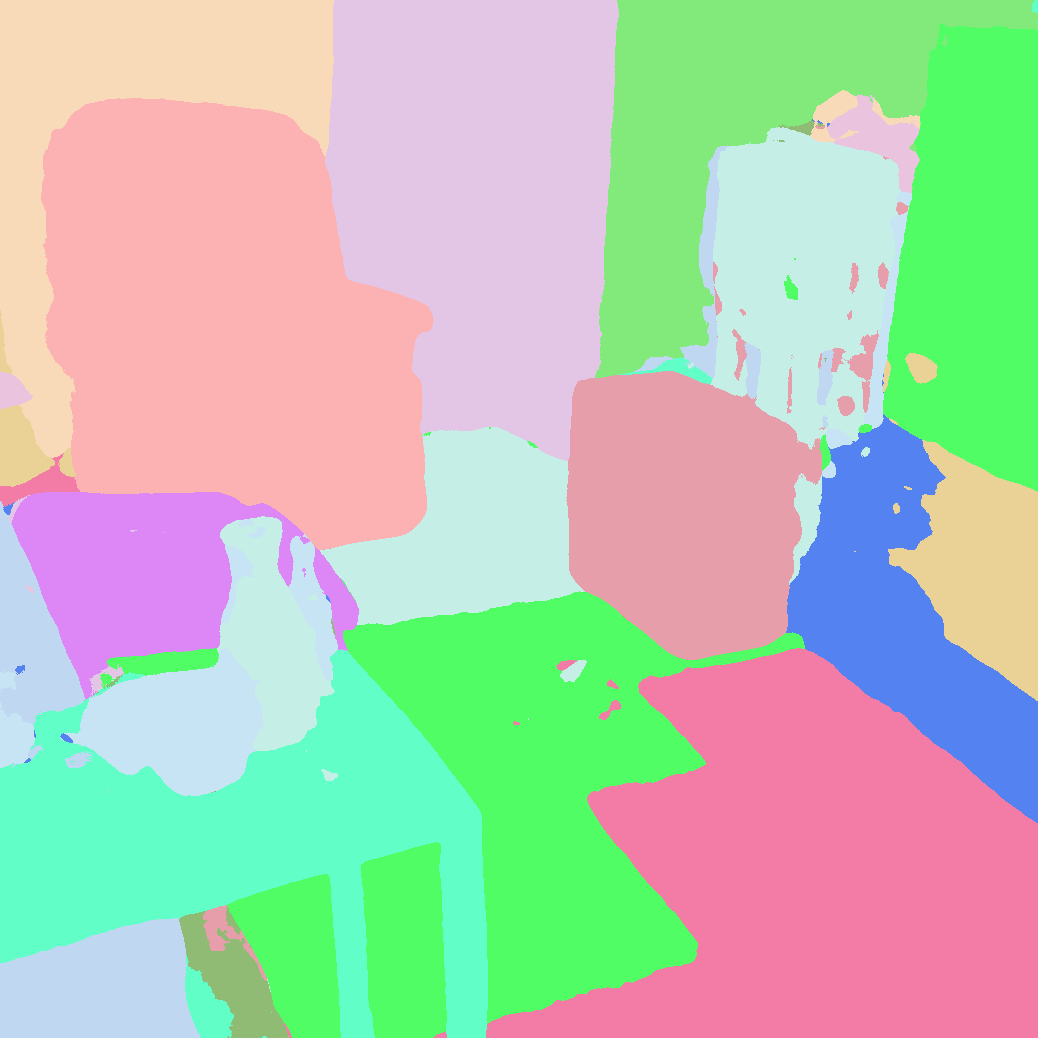}\vspace{0.13em}
        \includegraphics[width=\textwidth]
        {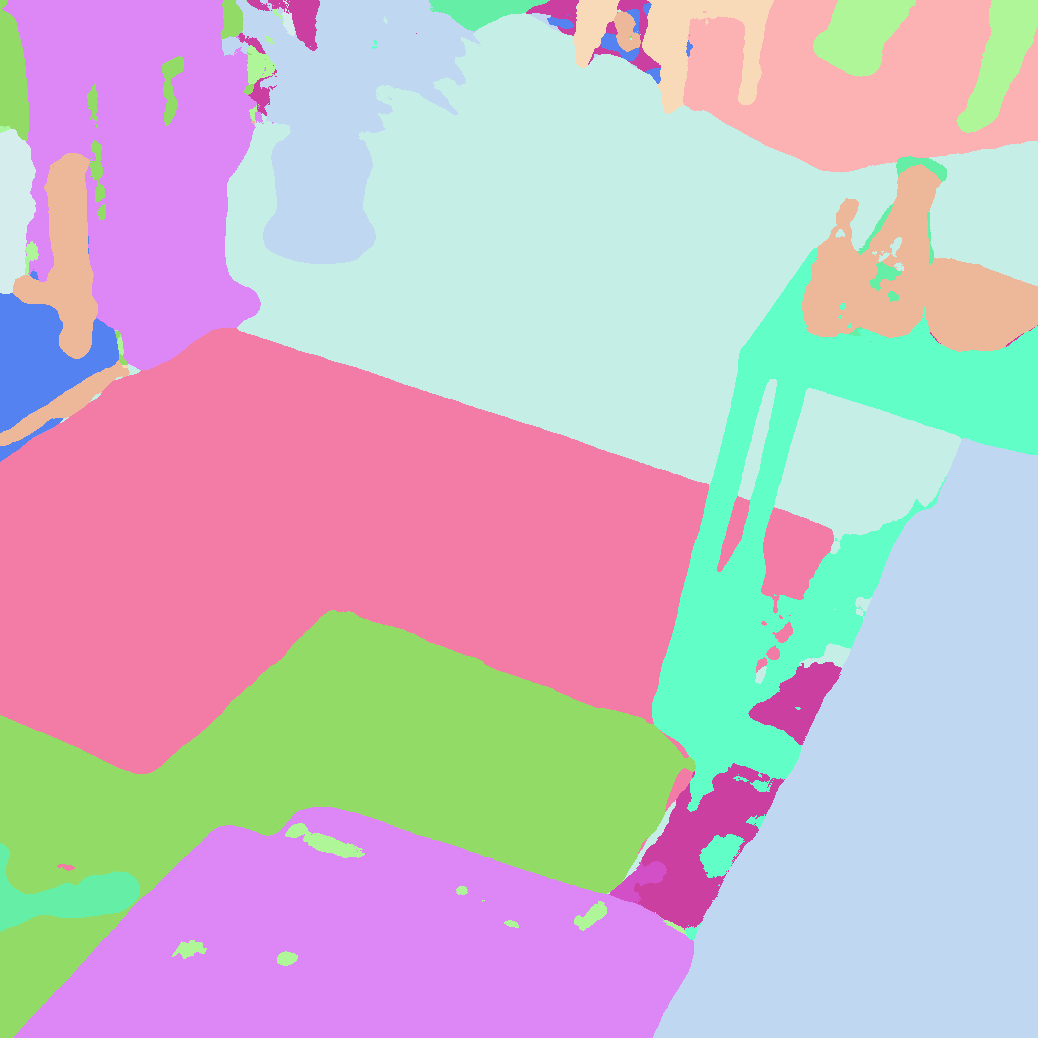}\vspace{0.13em}
        \caption*{Ours}
    \end{subfigure}\hfill
    \caption{Comparing 3D radiance field decomposition models for the scenes of \textit{counter} and \textit{room}.}
    \label{fig:comparison0}
\end{figure*}

%% file: fig/ablation.tex

\begin{figure*}[t]
    \centering\hfill
    \begin{subfigure}[t]{0.23\textwidth}
        \includegraphics[width=\textwidth]  
        {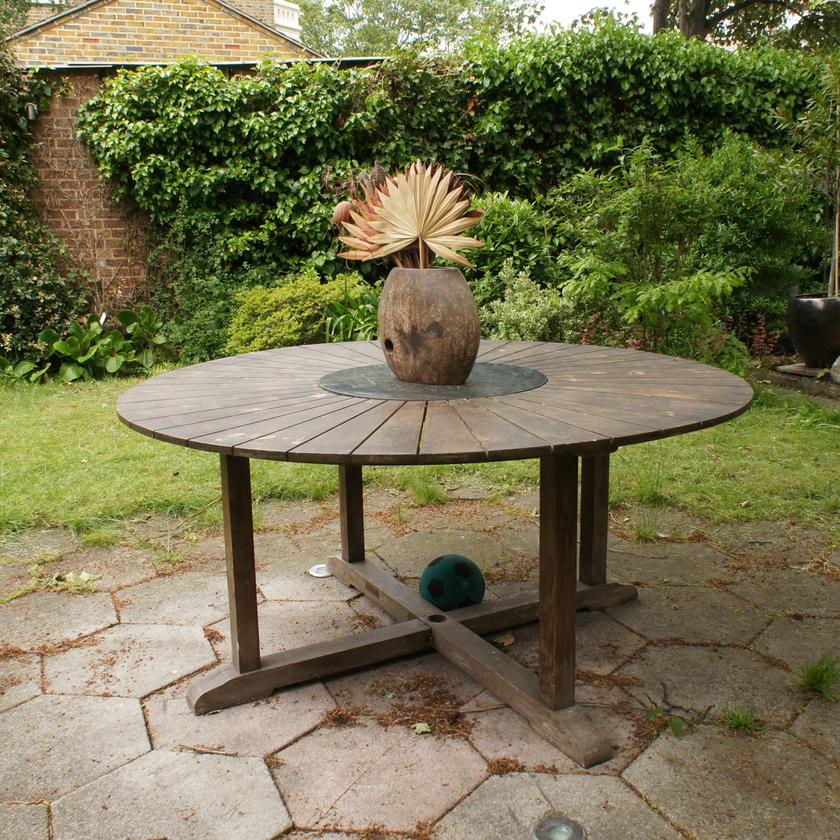}\vspace{0.25em}
        \includegraphics[width=\textwidth]
        {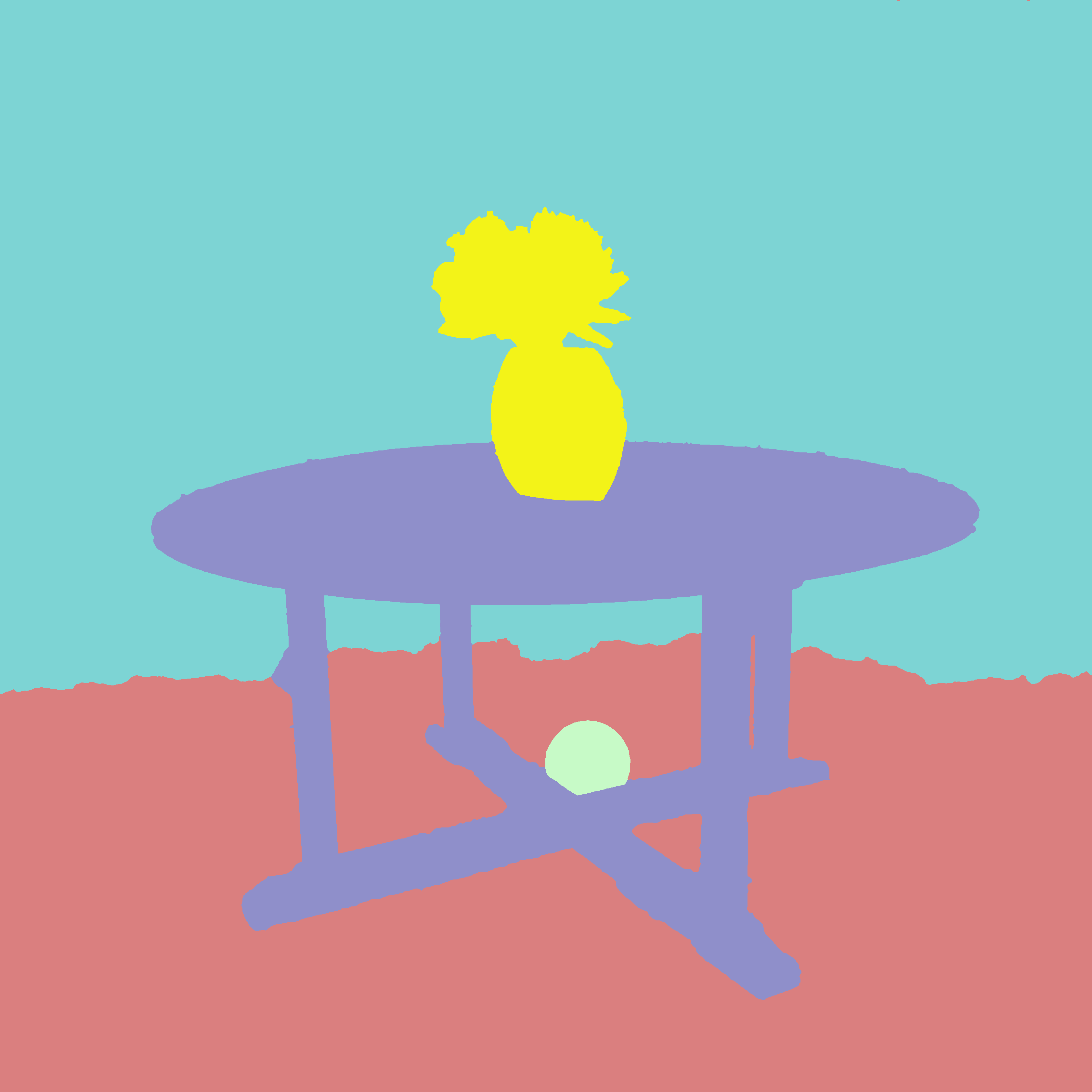}
        \caption{Ground truth}
    \end{subfigure}\hfill
    \begin{subfigure}[t]{0.23\textwidth}
        \includegraphics[width=\textwidth]  
        {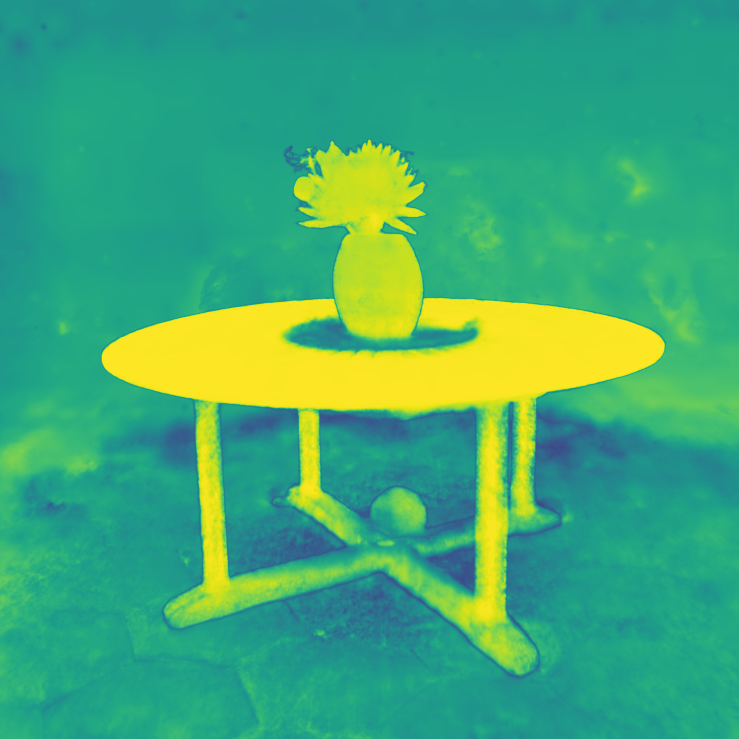}\vspace{0.25em}
        \includegraphics[width=\textwidth]
        {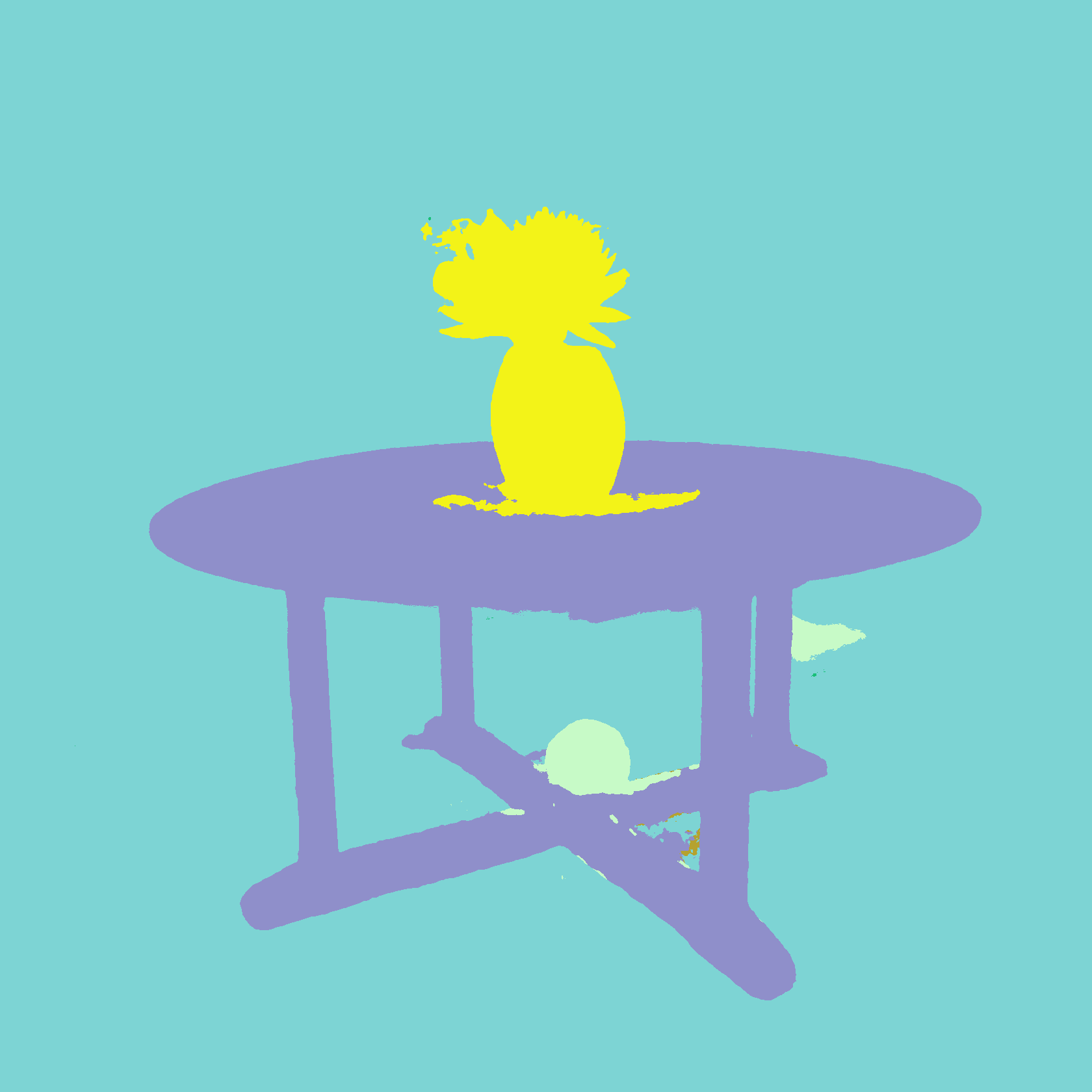}
        \caption{w/o matching}
    \end{subfigure}\hfill
    \begin{subfigure}[t]{0.23\textwidth}
        \includegraphics[width=\textwidth]  
        {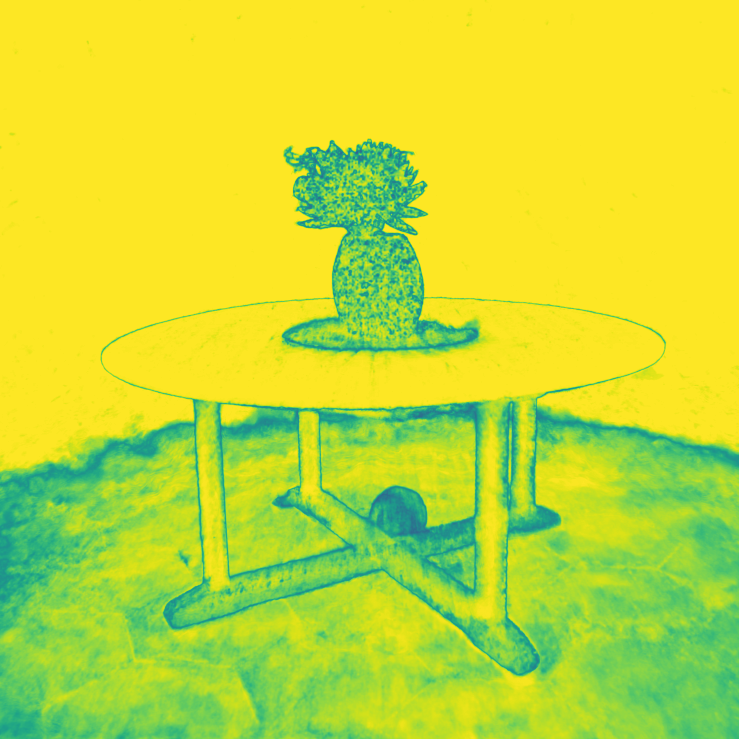}\vspace{0.25em}
        \includegraphics[width=\textwidth]
        {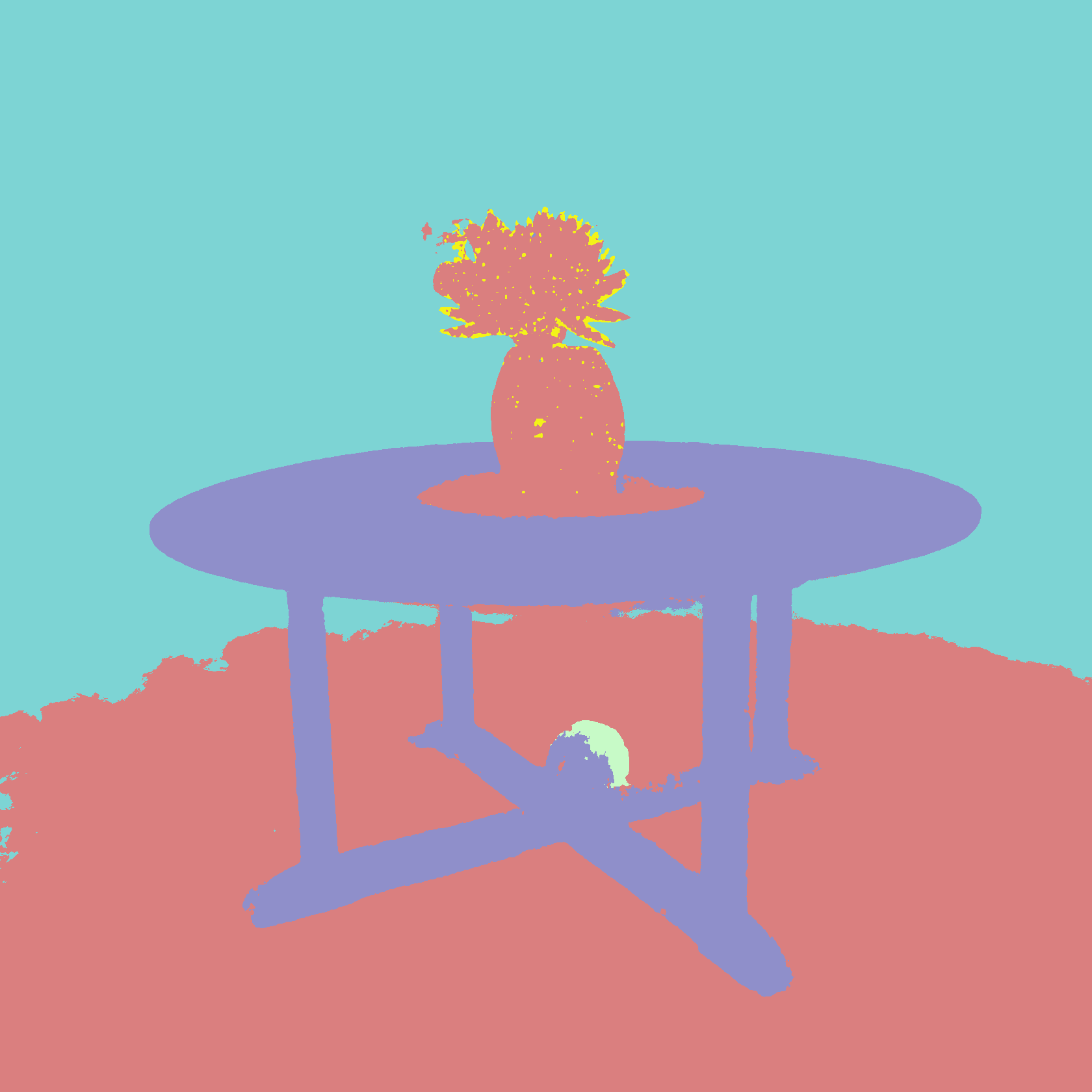}
        \caption{w/o $\mathcal{L}_{TV}$}
    \end{subfigure}\hfill
    \begin{subfigure}[t]{0.23\textwidth}
        \includegraphics[width=\textwidth]  
        {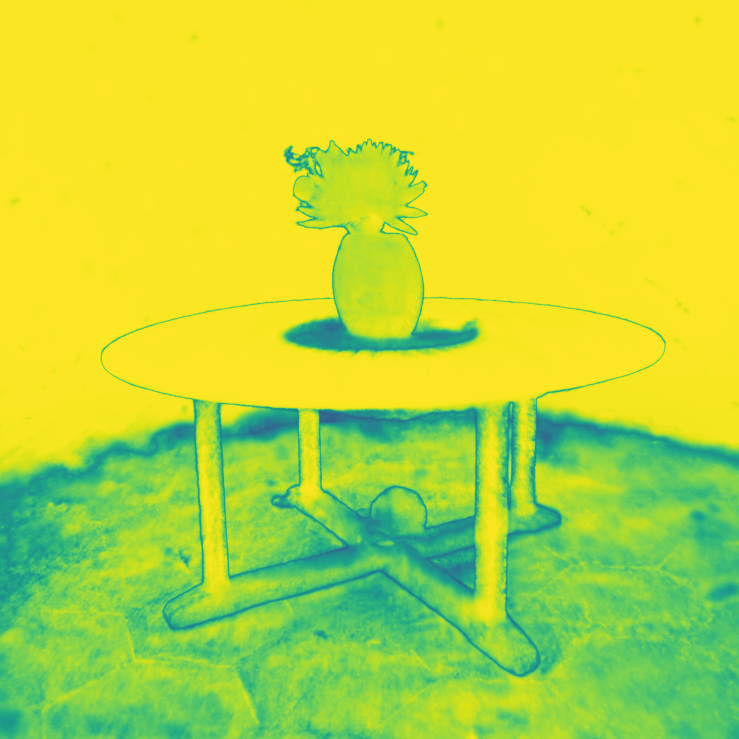}\vspace{0.25em}
        \includegraphics[width=\textwidth]
        {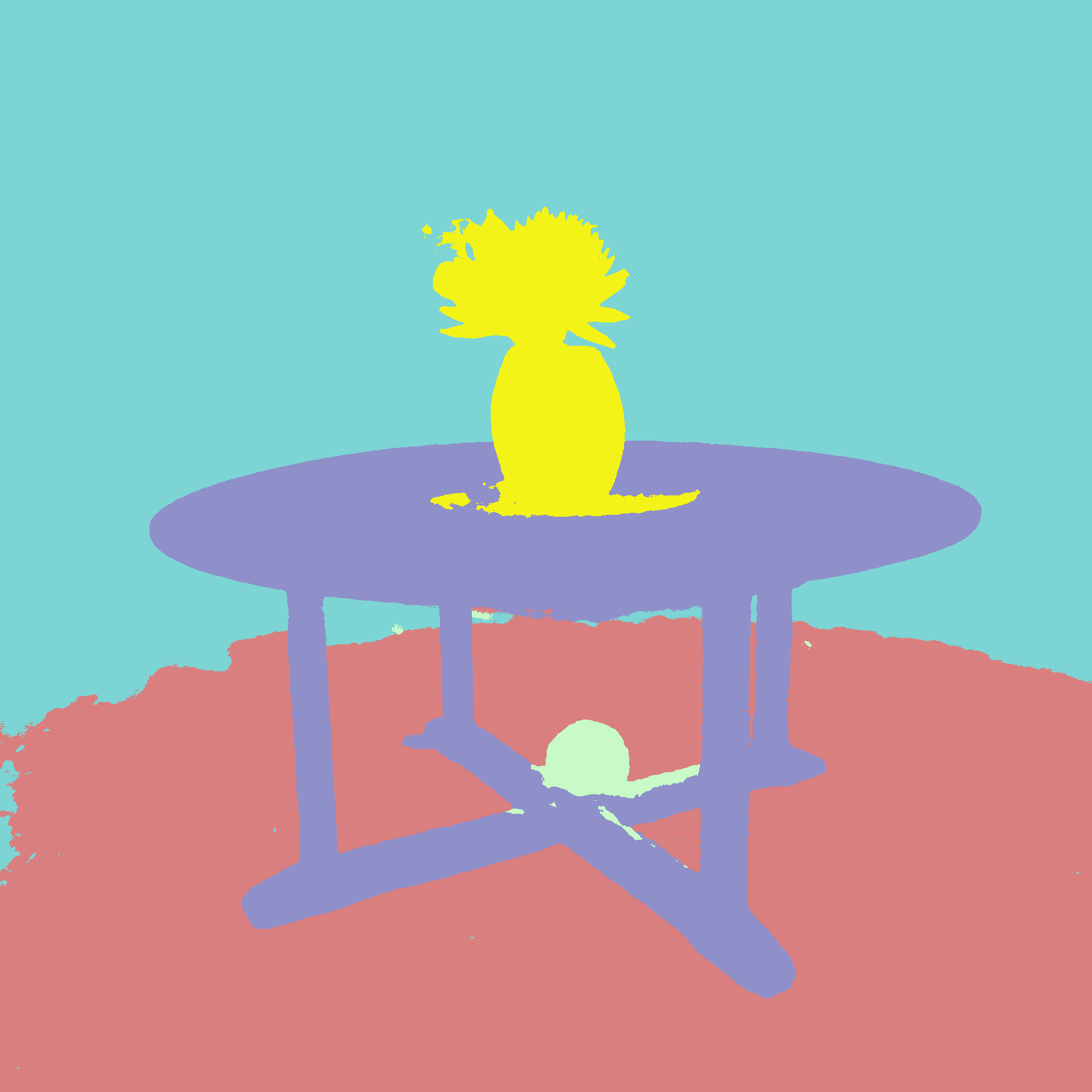}
        \caption{Ours}
    \end{subfigure}\hfill
    \begin{subfigure}[t]{0.033\textwidth}
        \includegraphics[width=\textwidth]  
        {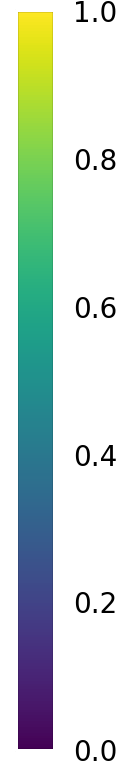}
    \end{subfigure}\hfill \vspace{-0.5em}
    \caption{\small {\bf Ablation.} Top row: the \textit{prediction confidence} is defined as the maximum value across object slots. Without matching, objects are often missed or captured within the same slots. Without $\mathcal{L}_{TV}$, the segmentation is not spatially consistent and its confidence is noisy.}
    \label{fig:ablation0}
\end{figure*}

%% file: fig/comparison.tex

\begin{table*}[ht]
    \centering
    \begin{tabularx}{0.85\textwidth}{>{\centering\arraybackslash}X*{6}{>{\centering\arraybackslash}X}}
    \toprule
    & DFFv2 & Instance-NeRF & Panoptic Lifting & PL+SAM & SAM (2D) & Ours \\
    \midrule
    IoU (\%) & 65.60 & 31.71 & 29.33  & 64.12 & 76.14  & \textbf{79.24}\\
    BD (\%)  & 74.57 & 53.75 & 55.41  & 78.32 & 87.12  & \textbf{87.51}\\
    SBD (\%) & 73.82 & 44.19 & 41.73  & 72.90 & 83.39  & \textbf{85.34}\\
    \bottomrule
    \end{tabularx}\vspace{-2mm}
    \caption{Segmentation results on the Mip-NeRF 360 \cite{Barron22} dataset.}
    \label{tab:comparison}
\end{table*}

%% file: sec/5_conclusion.tex

\section{Conclusion}

We have shown that we can successfully learn a meaningful 3D object field from inconsistent 2D masks. We match the channels of our object field with the supervision masks, and regularize the 3D field to produce a spatially consistent 3D segmentation. This is done without requiring semantic labels, which would limit the generality of our approach. 
One current limitation is that non-contiguous objects may, in rare occasions, be assigned the same label. This is mitigated by our regularization and in practice, such objects can trivially be separated.
Although our method is class agnostic, we observe a drop in quality for transparent objects, highly-reflective surfaces, or particularly small objects.
Future work will focus on dynamic radiance fields~\cite{Pumarola20} and enforcing consistency of the segmentations over time. Finally, while our current approach relies on NeRFs, it would be interesting to extend it to other representations such as 3DGS~\cite{Kerbl23} or PBDR~\cite{Jakob22}.


%% file: sec/X_suppl.tex
\clearpage
\setcounter{page}{1}
\maketitlesupplementary

\begin{figure*}[htb]
    \centering\hfill
    \begin{subfigure}[t]{0.32\textwidth}
        \includegraphics[width=\textwidth]  
        {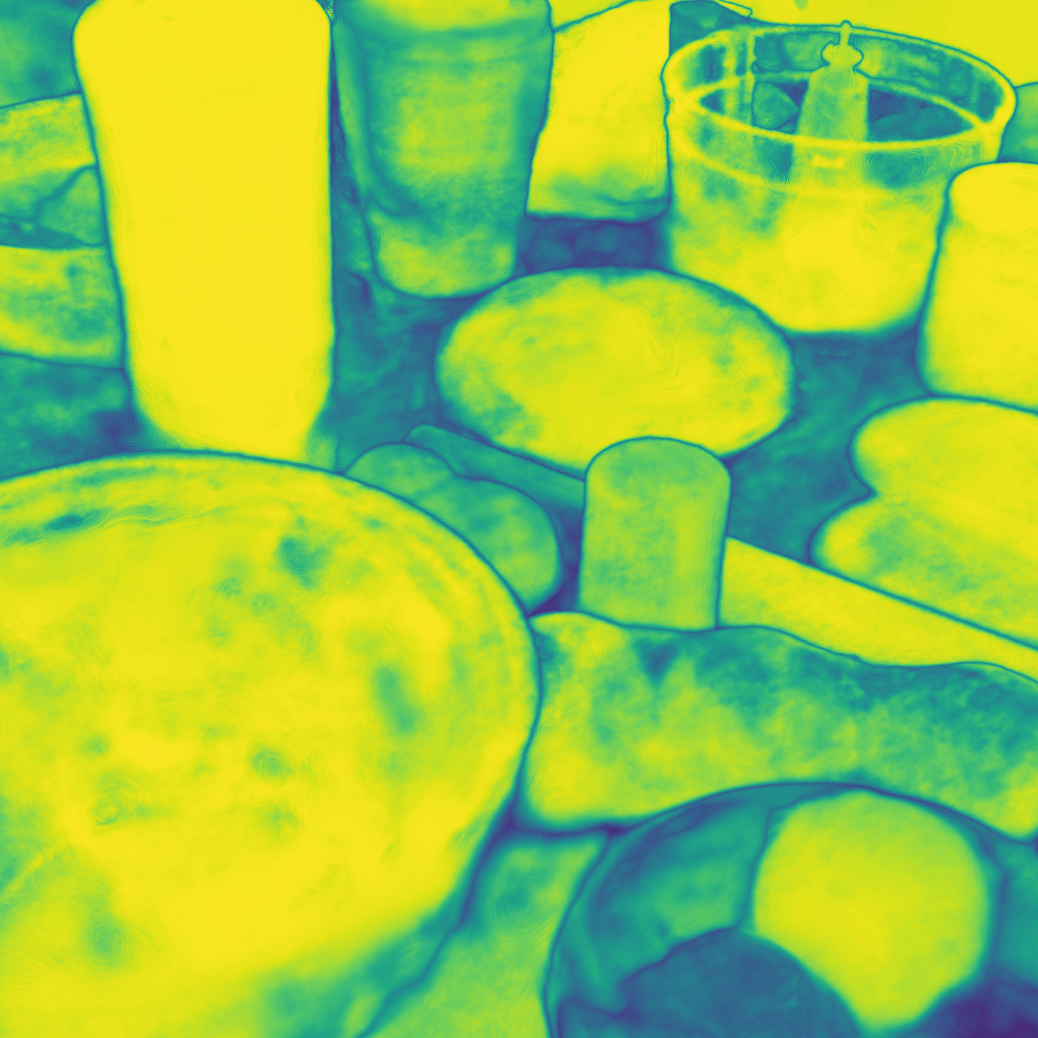}\vspace{0.35em}
        \includegraphics[width=\textwidth]
        {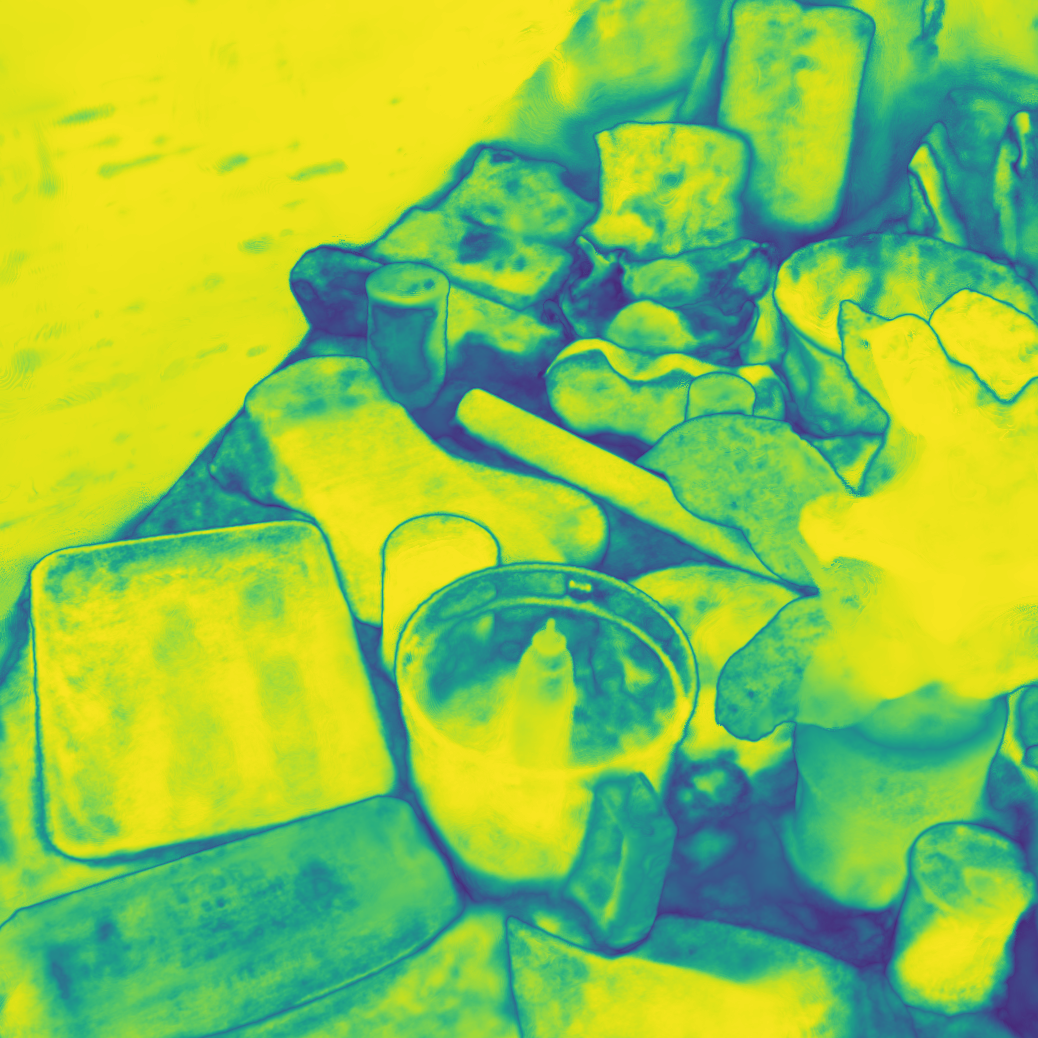}\vspace{0.35em}
        \caption{\textit{Counter}}
    \end{subfigure}\hfill
    \begin{subfigure}[t]{0.32\textwidth}
        \includegraphics[width=\textwidth]  
        {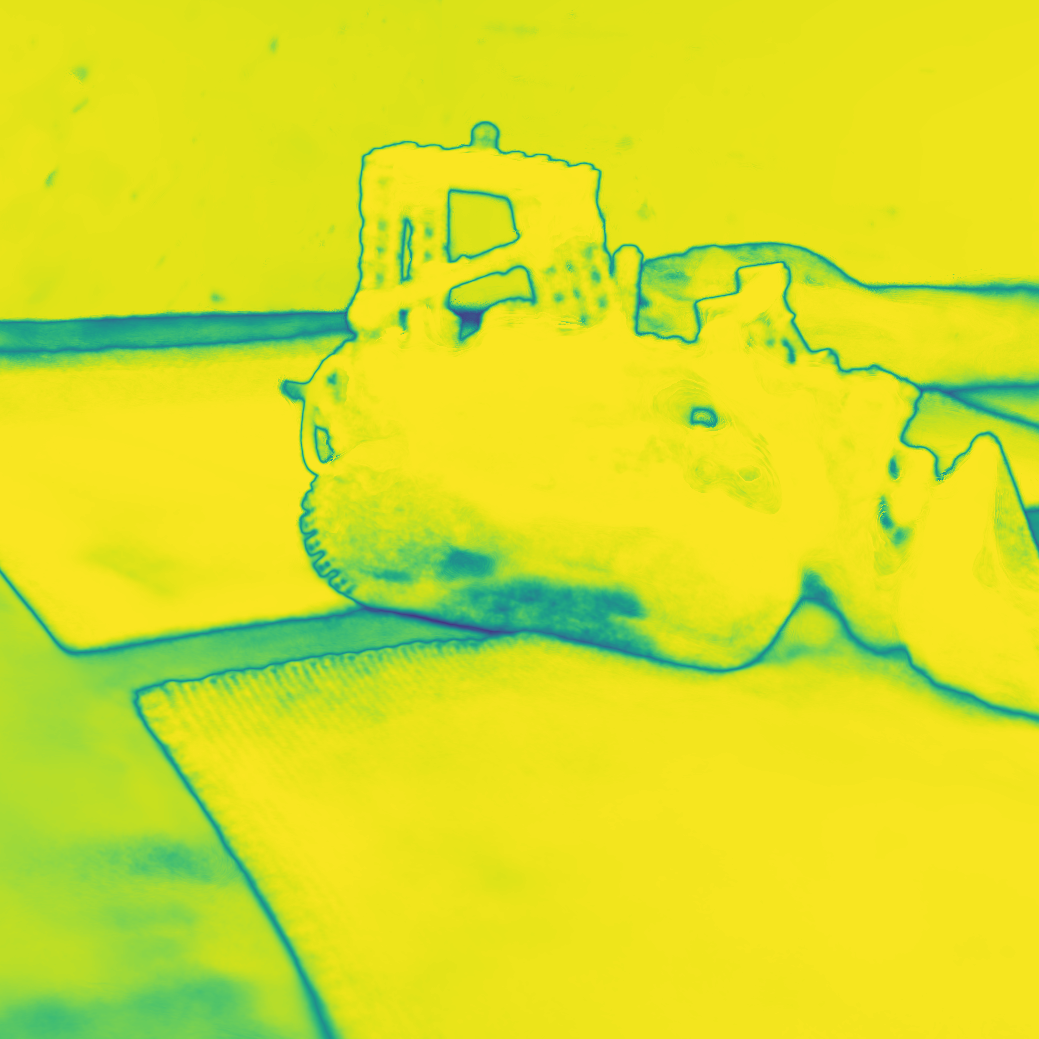}\vspace{0.35em}
        \includegraphics[width=\textwidth]
        {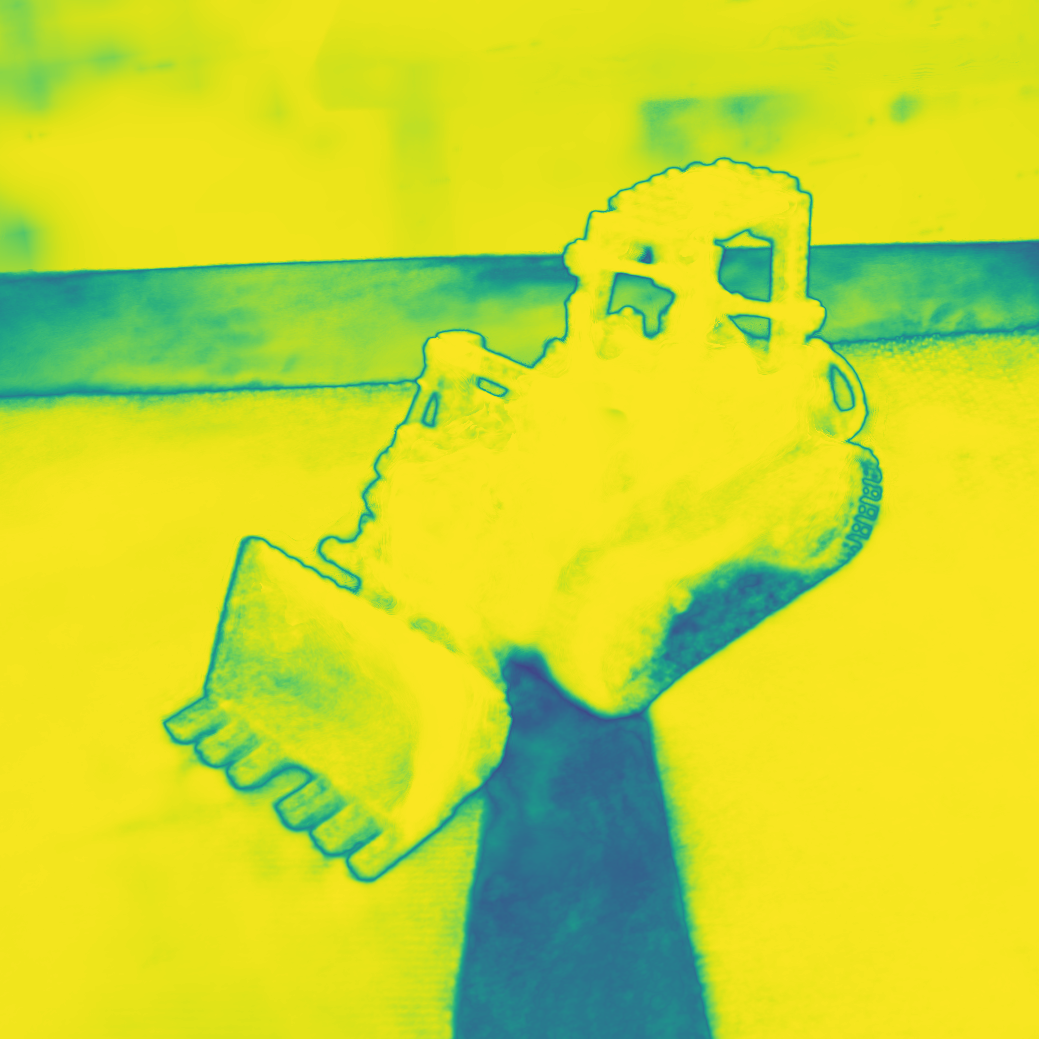}\vspace{0.35em}
        \caption{\textit{Kitchen}}
    \end{subfigure}\hfill
    \begin{subfigure}[t]{0.32\textwidth}
        \includegraphics[width=\textwidth]  
        {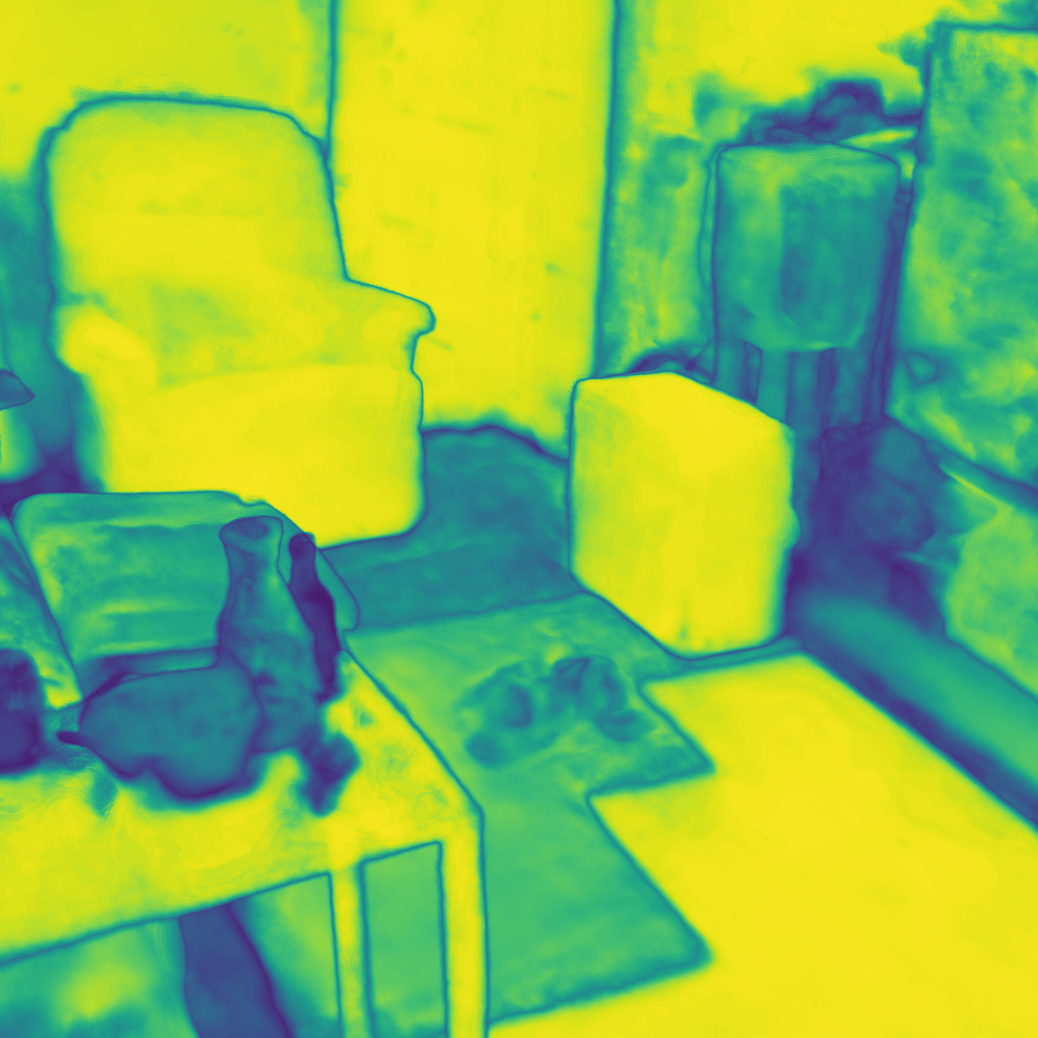}\vspace{0.35em}
        \includegraphics[width=\textwidth]
        {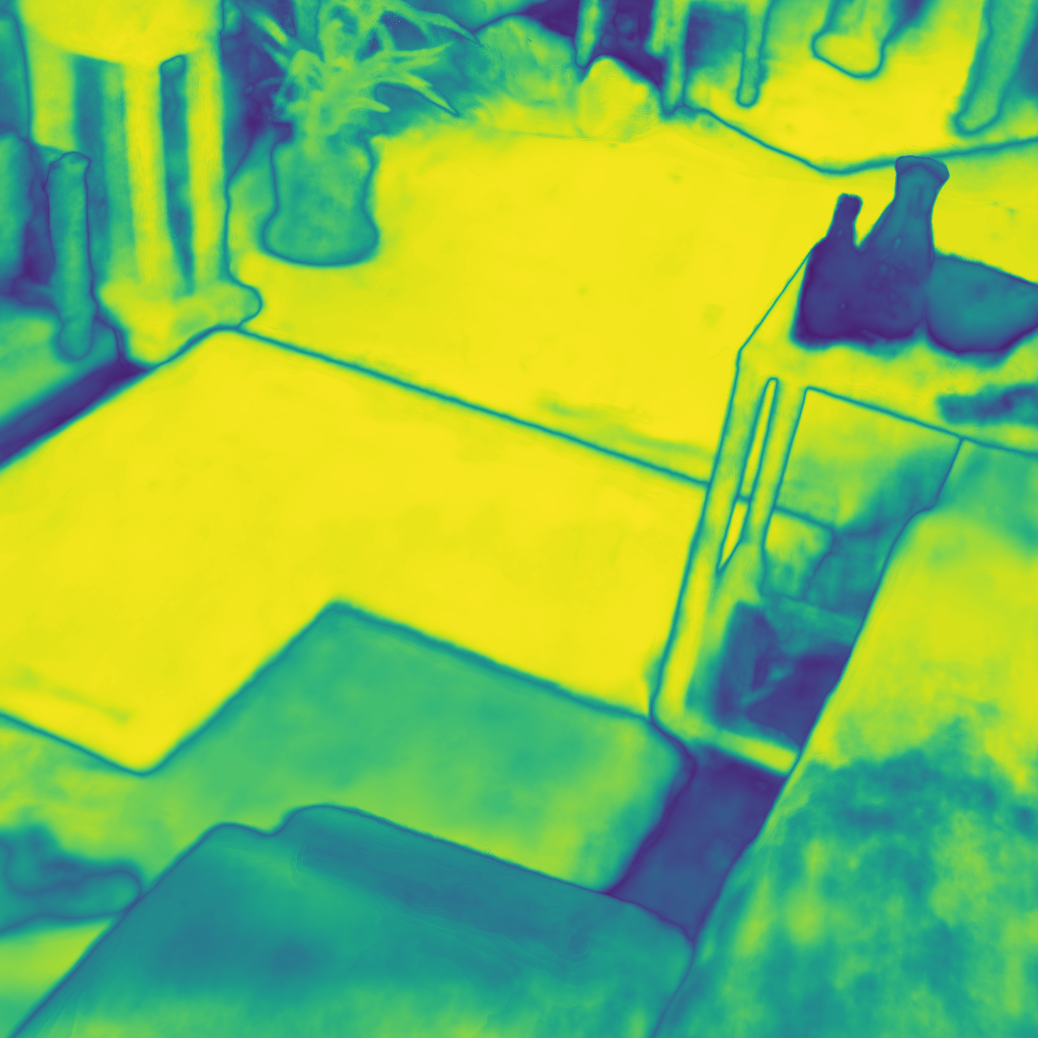}\vspace{0.35em}
        \caption{\textit{Room}}
    \end{subfigure}\hfill
    \caption{Visualization of object prediction confidence, defined as the maximal object value per-pixel. Bright regions indicate high confidence and a well-defined object.}
    \label{fig:confidence}
\end{figure*}

\section{Additional discussion}

\subsection{Analysis of confidence}

In \cref{fig:confidence}, we visualize the confidence estimation of our object network. We define confidence as the maximum value per-pixel across all object slots.
Low-confidence regions often correspond to objects that were not correctly identified, such as the bottles in the \textit{room} scene or the isolated patch of table in the \textit{kitchen}. In future work, an interesting direction would be to adaptively generate 2D mask supervision for regions with low object field confidence.

\subsection{Detailed quantitative results}
In \cref{tab:comparison_full}, we provide the complete results of our method and baselines on the Mip-NeRF 360 dataset. The ground truth masks are segmented by hand and cover all pixels in their respective images. The first and the one-hundredth images of each scene are selected for evaluation, and the number of object per-image ranges between 5 and 38. Similarly to the aggregated results reported, our method outperforms baselines for all tested scenes and for all metrics.

\begin{table*}[htb]
    \centering
    \begin{tabularx}{0.9\textwidth}{>{\centering\arraybackslash}X*{6}{>{\centering\arraybackslash}X}}
    \toprule
    Input & Metric & DFFv2 & Instance-NeRF & Panoptic Lifting & PL+SAM & Ours \\
    \midrule
    \multirow{3}{*}{\textit{bicycle}} 
    & IoU & 70.29 & 26.04 & 26.04  & 75.06 & \textbf{79.24} \\
    & BD & 82.47 & 51.23 & 51.23  & 84.56 & \textbf{87.51} \\
    & SBD & 79.56 & 40.24 & 40.24  & 83.82 & \textbf{85.34} \\
    \midrule
    \multirow{3}{*}{\textit{bonsai}} 
    & IoU & 76.57 & 34.68 & 39.15 & 85.26 & \textbf{88.63} \\
    & BD & 83.51 & 53.16 & 70.59 & \textbf{94.85} & 91.12 \\
    & SBD & 83.14 & 49.55 & 53.19 & 89.73 & \textbf{90.33} \\
    \midrule
    \multirow{3}{*}{\textit{counter}} 
    & IoU & 49.08 & 25.66 & 19.77 & 46.00 & \textbf{69.86} \\
    & BD & 60.84 & 52.76 & 44.83 & 65.45 & \textbf{83.33} \\
    & SBD & 59.62 & 35.33 & 27.53 & 55.24 & \textbf{78.52 }\\
    \midrule
    \multirow{3}{*}{\textit{garden}} 
    & IoU & 85.64 & 57.57 & 34.45 & 58.85 & \textbf{93.67} \\
    & BD & 88.66 & 72.04 & 62.86 & 72.58 & \textbf{97.03} \\
    & SBD & 88.66 & 69.00 & 49.87 & 71.16 & \textbf{96.33} \\
    \midrule
    \multirow{3}{*}{\textit{kitchen}} 
    & IoU & 68.28 & 19.98 & 20.81 & 86.05 & \textbf{88.87} \\
    & BD & 77.56 & 40.95 & 41.75 & 93.84 & \textbf{95.36} \\
    & SBD & 77.56 & 32.82 & 33.93 & 90.06 & \textbf{92.80} \\
    \midrule
    \multirow{3}{*}{\textit{room}} 
    & IoU & 43.76 & 26.36 & 35.75 & 33.49 & \textbf{59.95} \\
    & BD & 54.39 & 52.34 & 61.21 & 58.60 & \textbf{73.64} \\
    & SBD & 54.39 & 38.22 & 45.64 & 47.38 & \textbf{71.70} \\
    \midrule
    \multirow{3}{*}{Average} 
    & IoU & 65.60 & 31.71 & 29.33 & 64.12 & \textbf{79.24} \\
    & BD & 74.57 & 53.75 & 55.41 & 78.32 & \textbf{87.51} \\
    & SBD & 73.82 & 44.19 & 41.73 & 72.90 & \textbf{85.34} \\
    \bottomrule
    \end{tabularx}\vspace{2mm}
    \caption{Segmentation results on the Mip-NeRF 360 \cite{Barron22} dataset.}
    \label{tab:comparison_full}
\end{table*}

\subsection{Limitations}

\begin{figure*}[t]
    \centering\hfill
    \begin{subfigure}[t]{0.32\textwidth}
        \includegraphics[width=\textwidth]  
        {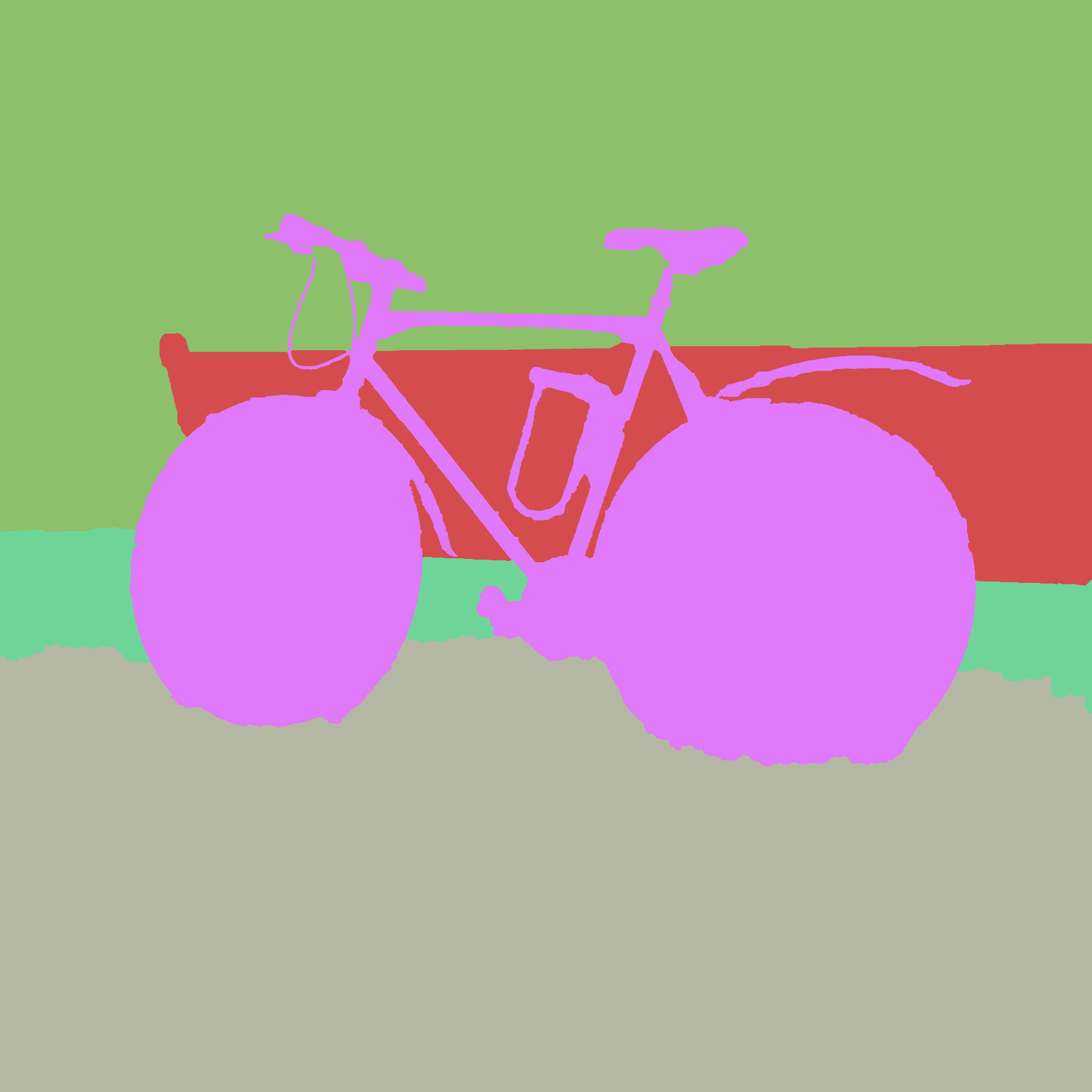}\vspace{0.35em}
        \caption{Ground-truth}
    \end{subfigure}\hfill
    \begin{subfigure}[t]{0.32\textwidth}
        \includegraphics[width=\textwidth]  
        {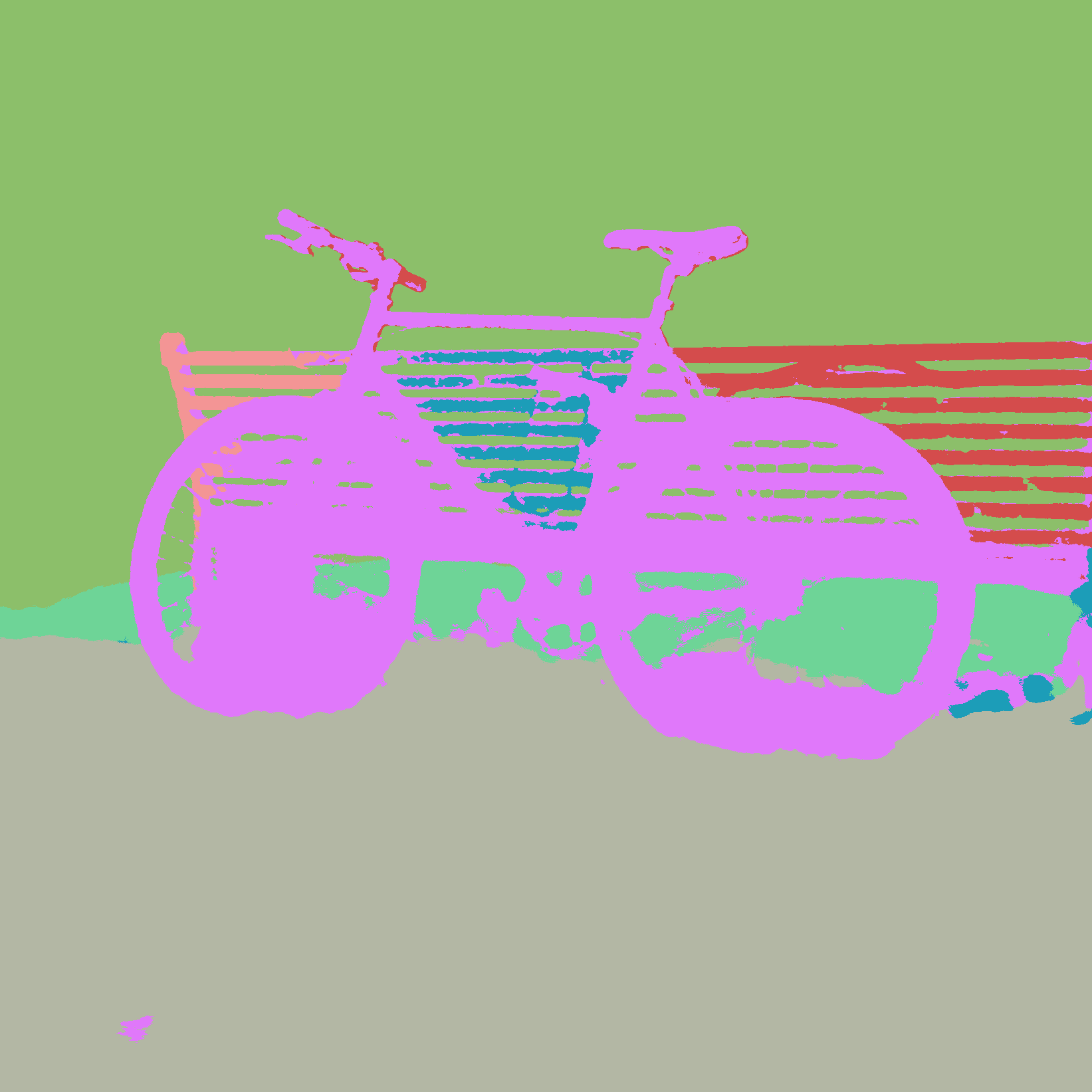}\vspace{0.35em}
        \caption{Segmentation}
    \end{subfigure}\hfill
    \begin{subfigure}[t]{0.32\textwidth}
        \includegraphics[width=\textwidth]  
        {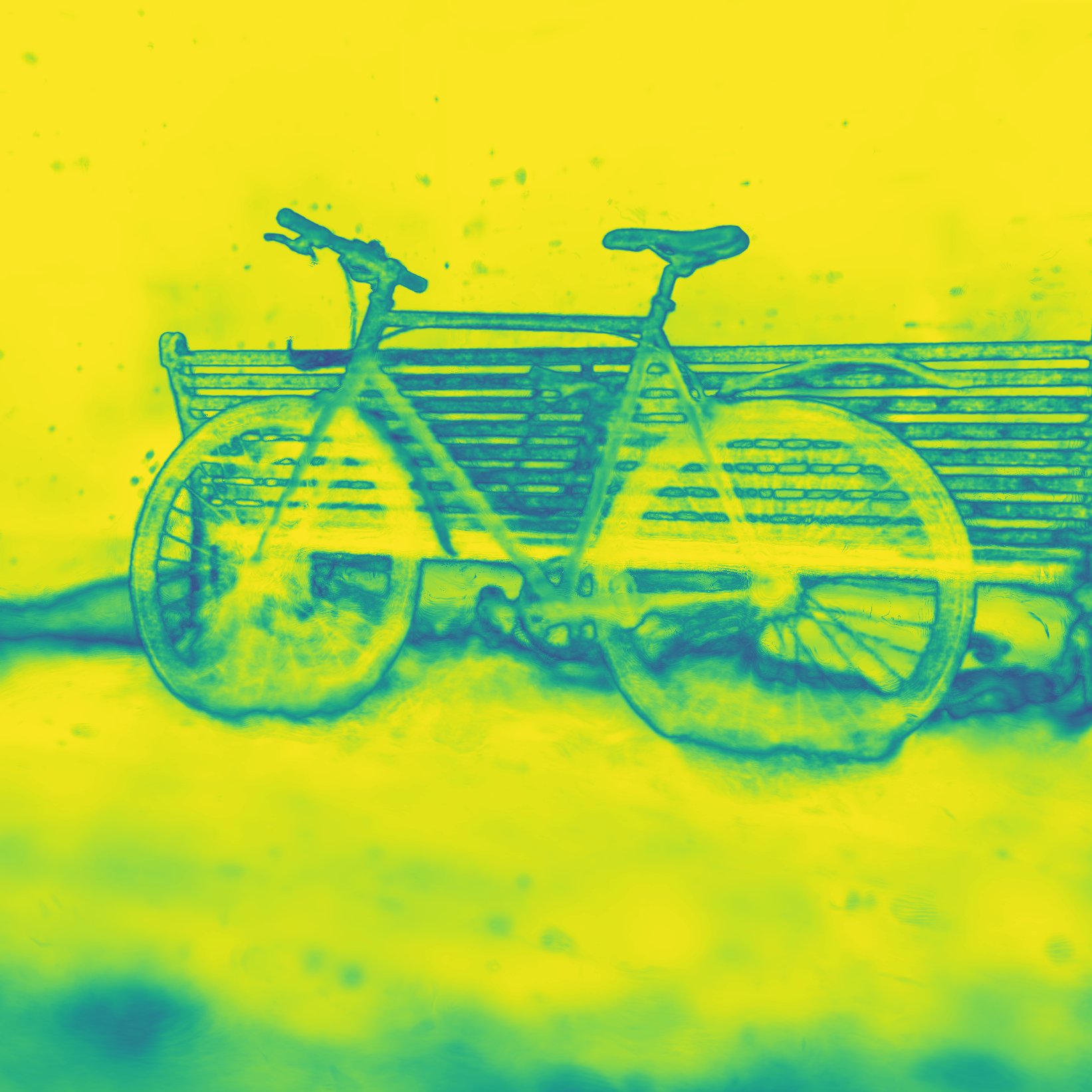}\vspace{0.35em}
        \caption{Confidence}
    \end{subfigure}\hfill
    \caption{\small {\bf Limitations.} Accurately segmenting thin structures remains a challenge.}
    \label{fig:limit}
\end{figure*}

Despite our proposed matching and regularization to reduce the effect of 2D mask inconsistency, we observed that thin structures remain a challenge. 
As illustrated in \cref{fig:limit} with the \textit{bicycle} scene, our method is unable to assign a single object slot to the bench, which is partially occluded by the bicycle. Thin structures are challenging to reconstruct accurately for NeRFs, and SAM often fails to capture them and tends to produce masks that encapsulate more than the object itself.

\begin{figure*}[htb]
    \centering\hfill
    \begin{subfigure}[t]{0.25\textwidth}
        \includegraphics[width=\textwidth]  
        {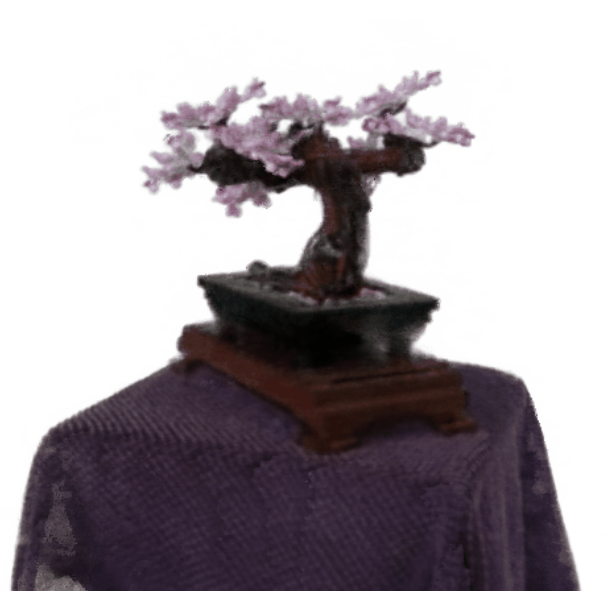}\vspace{0.35em}
        \caption{Render}
    \end{subfigure}\hfill
    \begin{subfigure}[t]{0.25\textwidth}
        \includegraphics[width=\textwidth]  
        {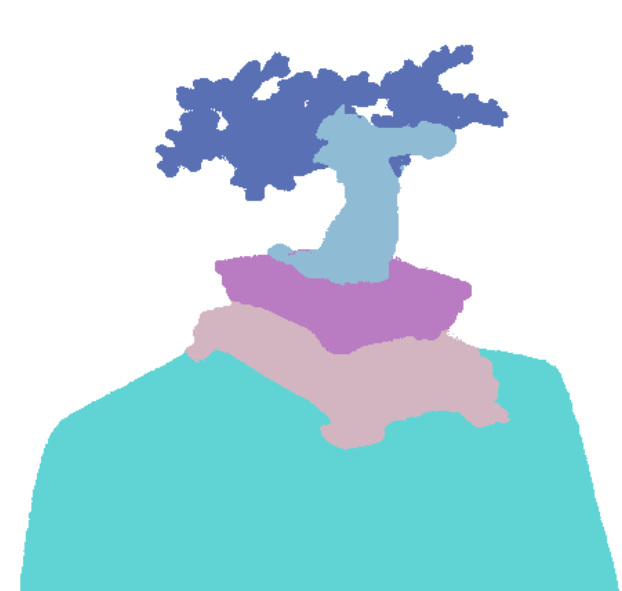}\vspace{0.35em}
        \caption{Object field}
    \end{subfigure}\hfill
    \begin{subfigure}[t]{0.32\textwidth}
        \includegraphics[width=\textwidth]  
        {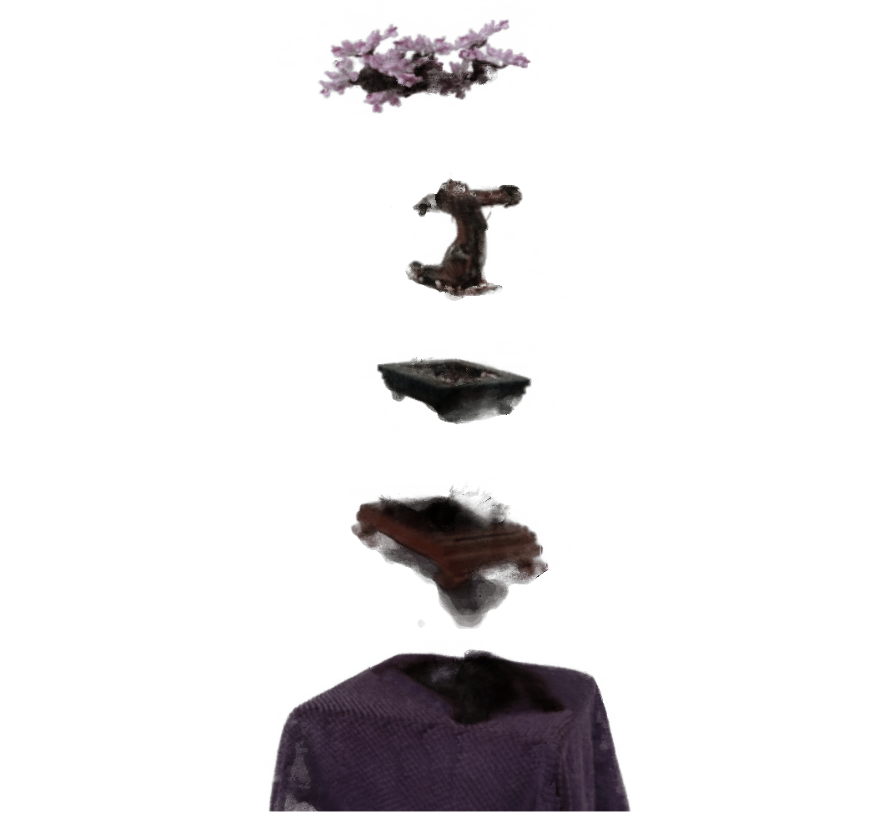}\vspace{0.35em}
        \caption{Decomposition}
    \end{subfigure}\hfill
    \caption{Object decomposition of the \textit{bonsai} model.}
    \label{fig:decomposition}
\end{figure*}

Furthermore, partially occluded objects may not be satisfyingly reconstructed by the initial NeRF. As a result, object extraction will sometimes reveal NeRF \textit{floaters} in regions of space which are not supervised by the ground-truth images. This effect is visible in \cref{fig:decomposition}, where the table underneath the pedestal includes some reconstruction artifact.

\section{Implementation details}

\subsection{Mask generation}

We generate the 2D supervision masks with SAM \cite{Kirillov23}. Specifically, we used the pretrained ViT-H model weights and the \textit{Automatic Mask Generator} class, which takes as input an RGB image and \textit{no prompt}, and outputs a set of masks separating objects in the image. Since SAM pads images to be square, we first crop the input images to squares in order to produce higher quality masks. In post-processing, we remove masks which are included in another mask.
We also discard masks that take up less than $3\%$ of the total image in pixels.
For each mask generated, SAM estimates a predicted IoU score and a stability score. We additionally discard masks for which any of these scores is below $70\%$.
Note that the union of these masks may not cover each pixel. This is taken into consideration in our method, and some object slots may not be matched with a mask at every iteration.
 
For our experiments on the Mip-NeRF 360 dataset, we select a SAM granularity parameter $g \in [1,3]$ such that at least 5 different masks are produced for each scene. Precisely, we use granularity levels of 1 for \textit{bicycle}, 0 for \textit{bonsai}, and 2 for the remaining scenes. Since the images in this dataset are very high resolution, we downscale images by a factor of 4 prior to mask generation, and observe that this tends to lead to smoother masks. For this same reason, we dilate masks by 5 pixels to reduce the effect of possible artifacts produced by SAM.
For object-centric scenes, we additionally generate a background mask by masking the region outside of the unit-sphere in the 3D NeRF. The foreground is then given to SAM as input to generate masks for the remaining objects.

\subsection{NeRF reconstruction}

In this section, we describe the NeRF implementation used in our approach. 
Note that this NeRF reconstruction is not central to our method and could easily be replaced with another implementation. In fact, our work could be extended to other differentiable rendering techniques as the only requirement of our method is a differentiable method to produce 2D object images from a learned 3D field.

We first encode the position $\bx$ using learned hash grid features. We start NeRF optimization with the 4 lowest resolution hash grids initially, and increase the number of activated hash grids every 50 iterations, up to a maximum of 16. The interpolated features are then decoded by a multi-layer perceptron with 3 hidden layers and a hidden layer dimension of 64. Afterwards, features of dimension 15 are concatenated to the view direction and fed to the color decoder, which has 3 hidden layers of 32 dimensions.

We assume the region of interest is located within the $[-1,1]^3$ cube, and contract the background region $[-128,-1]\times[1,128]$ such that we only sample points in $[-2, 2]$, as described in \cite{Barron22}. Additionally, we optimize a proposal network with 2 hidden layers of 16 dimensions. In our experiments, we observed that a satisfying 3D segmentation was learned within the first minute of training. We further train for a total of 10000 iterations to attain convergence, which takes between 10 and 30 minutes depending on the complexity of the scene and the number of images. All networks are trained on a single A100 GPU with 40Gb of VRAM.

In order to prevent NeRF \textit{floaters} and simplify object extraction, we add an L1 loss on the densities of points randomly sampled in $[-2, 2]$, referred to as $\mathcal{L}_{empty}$. This loss is paired with a very small $\lambda_{empty} = 10^{-4}$ to regularize unseen regions of space without interfering with the actual geometry.

\subsection{Object Field}

We optimize the object field for 2000 iterations with a batch size of $B = 5$. This batch size defines the number of randomly sampled views each iteration, where each view is associated to up to $K$ masks. In our experiments, $K$ ranges from 5 to 60 based on the selected granularity level and the complexity of the scene. Note that the matching of object slots and masks is performed per-view, such that different views don't interfere with each other and the matching method is computed $B$ times per iteration. 

For a 3D point $\bx$, we first encode $\bx$ using a learned feature hash grid with 8 levels. Then, these features are decoded by a fully-connected network with 2 hidden layers and a hidden dimension of 32. The final layer outputs $N$ values, where $N$ is the maximum number of objects. Finally, the resulting vector is fed to a softmax layer.
We set the initial learning rate to $0.15$, and apply a learning rate decay of $0.005$ after 20 iterations of warm-up. The loss weights are set to $\lambda_{TV} = \lambda_{FP} = 0.01$.

Note that we freeze the weights of the NeRF reconstruction in order to train the object field. Previous work \cite{Siddiqui23} has shown that without a stop gradient on the semantic loss, additional floaters appeared in the 3D reconstruction. Furthermore, we optimize NeRFs by sampling random rays across all views, whereas the object network is trained by sampling an image and considering all masks available for that image. This major difference in training method and the results of previous work explain our decision to not jointly train our NeRF and object field networks.